\documentclass[10pt]{article} 
\usepackage[accepted]{rlc}

\usepackage{amssymb}            
\usepackage{mathtools}          
\usepackage{mathrsfs}           
\mathtoolsset{showonlyrefs}     
\usepackage{graphicx}           
\usepackage{subcaption}         
\usepackage[space]{grffile}     
\usepackage{url}                
\usepackage{url}
\usepackage{bm}
\usepackage{gensymb}
\makeatletter
\let\MYcaption\@makecaption
\makeatother
\usepackage[font=footnotesize]{subcaption}
\makeatletter
\let\@makecaption\MYcaption
\makeatother
\usepackage{subcaption}
\usepackage{booktabs}

\usepackage{ifthen} 
\usepackage{amsmath,amsfonts} 
\usepackage{multirow}
\usepackage{colortbl}
\usepackage{arydshln}
\usepackage{tabularx}
\usepackage{wrapfig}
\definecolor{lightgray}{gray}{0.8}
\definecolor{lightorange}{rgb}{1.0, 0.917, 0.655}
\definecolor{lightblue}{rgb}{0.9,0.9,1.0}
\newcommand{\spm}[2][normal]{%
    \ifthenelse{\equal{#1}{bold}}%
    {\ensuremath{{\scriptstyle\boldsymbol{\pm} \mathbf{#2}}}}%
    {\ensuremath{{\scriptstyle \pm #2}}}%
}

\title{A Super-human Vision-based Reinforcement Learning Agent for Autonomous Racing in Gran Turismo}

\author{Miguel Vasco\thanks{Equal contribution.} \thanks{Work done during his internship at Tokyo Laboratory, Sony AI.} \\
    KTH Royal Institute of Technology \\
    miguelsv@kth.se
    \And
    Takuma Seno$^*$ \\
    Sony AI \\
    takuma.seno@sony.com
    \And
    Kenta Kawamoto \\
    Sony AI \\
    kenta.kawamoto@sony.com
    \And
    Kaushik Subramanian \\
    Sony AI \\
    kaushik.subramanian@sony.com
    \And
    Peter R. Wurman \\
    Sony AI \\
    peter.wurman@sony.com
    \And
    Peter Stone \\
    Sony AI \\ The University of Texas at Austin \\
    peter.stone@sony.com
    }

\begin{document}

\maketitle

\begin{abstract}
Racing autonomous cars faster than the best human drivers has been a longstanding grand challenge for the fields of Artificial Intelligence and robotics. Recently, an end-to-end deep reinforcement learning agent met this challenge in a high-fidelity racing simulator, Gran Turismo. However, this agent relied on global features that require instrumentation external to the car. This paper introduces, to the best of our knowledge, the first super-human car racing agent whose sensor input is purely local to the car, namely pixels from an ego-centric camera view and quantities that can be sensed from on-board the car, such as the car's velocity.  By leveraging global features only at training time, the learned agent is able to outperform the best human drivers in time trial (one car on the track at a time) races using only local input features. The resulting agent is evaluated in Gran Turismo 7 on multiple tracks and cars. Detailed ablation experiments demonstrate the agent's strong reliance on visual inputs, making it the first vision-based super-human car racing agent.
\end{abstract}

\section{Introduction}
Autonomous car racing is a challenging task for intelligent artificial agents, where performance gaps in milliseconds can be the difference between winning and losing a race. To effectively perform this task, agents must be able to (i) process high-dimensional sensor data to estimate the state of the autonomous vehicle, (ii) continuously plan optimal driving lines while avoiding obstacles and other vehicles, and (iii) control the vehicle, while accounting for its nonlinear behavior and the conditions of the road~\citep{betz2022autonomous}. Recently, deep reinforcement learning (RL) methods have shown great promise in learning racing behavior through trial-and-error interaction with the race track environment, without the need for extensive domain knowledge~\citep{jaritz2018end,gts_vision,cai2021vision,remonda2021formula,herman2021learn}. Despite their ability to consistently drive around the track, most learned policies still perform slower than median human racers~\citep{cai2021vision,herman2021learn}.

In this work, we focus on learning RL~\citep{sutton2018reinforcement} agents that are able to achieve
\emph{super-human} performance in autonomous racing tasks, i.e., they are able to outperform (in terms of lap time) the best human drivers on a given track. Recently, two RL methods have reported super-human performance in Gran Turismo, a high-fidelity racing simulator~\citep{gts,gt_sophy}. However, during execution these methods rely on \emph{global features}, such as forward looking course shape information, that require instrumentation external to the vehicle. In contrast, human drivers rely on car-centric \emph{local features} to race, such as \emph{visual information} and \emph{propriocentric features} that can be estimated from on-board the vehicle (e.g. velocity of the car). In this work, we ask the question: \emph{can we train RL agents that are able to consistently achieve super-human performance when provided only with local features at execution time?}

Learning optimal racing behavior requires information that might not be possible to access only through local features at each time step: for example, in a tight corner the agent might be unable to see the apex and the end of the curve, which are fundamental to select an optimal driving trajectory. To overcome this challenge, and motivated by recent works on reinforcement learning with partial observability~\citep{pinto2017asymmetric,salter2021attention,sinha2023asymmetric,baisero2022asymmetric}, we leverage a distributed \emph{asymmetric} actor-critic architecture that provides global features to the critic during training. The policy (actor) is provided only with local features, i.e., image and propriocentric features, allowing the agent to race without global information at execution time.

We evaluate our agent in Gran Turismo 7 (GT7), a high-fidelity driving simulator for PlayStation$^\text{\textregistered}$. We show that our agent consistently achieves faster lap times than all human reference drivers (over 130K per scenario) across multiple time trial races, in which the goal is to complete a lap around the track in the minimum amount of time. Additionally, we conduct an extensive ablation study that shows the significant contribution of local features and of the asymmetric training scheme to the agent's overall performance. Furthermore, we perform a qualitative study on the learned policy and highlight novel driving lines, in comparison with the best human reference drivers, and demonstrate the strong reliance on image features for the agent's decision-making. To the best of our knowledge, we present the first vision-based super-human car racing agent. 

In summary, our contributions are three-fold: (i) we contribute a vision-based RL agent for autonomous racing that employs an asymmetrical actor-critic training scheme; (ii) our agent consistently outperforms all human reference drivers (over 130K) across multiple time trial races in Gran Turismo 7, while having access only to local features, and performs on par with other super-human racing agents that rely on global features during execution; (iii) we conduct an extensive evaluation study that highlights the importance of the asymmetrical training scheme, novel driving behavior in comparison with the best human reference drivers, and the strong dependence on image features for the decision-making of our agent.

\section{Related Work}

Autonomous Racing (AR) is a subfield of autonomous driving research that concerns autonomous vehicles that operate at their dynamical and power limits within racing environments (such as racing circuits)~\citep{betz2022autonomous}.  Research in autonomous racing can traditionally be categorized into perception~\citep{massa2020lidar, peng2021vehicle}, planning~\citep{herrmann2020minimum, vazquez2020optimization} and control~\citep{williams2018robust, hao2022outracing}. In our paper, instead, we employ end-to-end RL that combines perception, planning, and control into a single process, in particular with vision inputs, and is able to achieve super-human performance. For an extended version of the related work, including a discussion on asymmetrical training in RL, please refer to Appendix~\ref{sec:appendix_extended_related_work}.

\noindent \textbf{Vision-based Reinforcement Learning for AR}: RL has been shown to be a promising approach to learn competitive racing behavior~\citep{o2019f1,herman2021learn,rong2020lgsvl,dosovitskiy2017carla}. \citet{jaritz2018end} explore vision-based RL for driving agents in the context of a rally game. However, the authors find that their method does not achieve optimal racing trajectories, ``lacking anticipation'', and is unable to complete the racing tracks without colliding several times with obstacles. \citet{cai2021vision} described an approach that merges imitation learning and model-based RL to learn racing behavior. However, their method requires expert-level demonstrations to pretrain the policy. The racing performance of current vision-based methods is still sub-optimal. Some works lack a performance comparison against humans~\citep{remonda2021formula,jaritz2018end} or, when such comparison is made, the methods still under perform significantly against median human users~\citep{cai2021vision,herman2021learn}. \citet{gts_vision} also explores vision-based RL for racing agents using a pretrained image encoder on random observations of the track environment. The authors report that they are unable to outperform the best human players. To the best of our knowledge, we contribute the first vision-based agent that is able to consistently outperform all human reference drivers across multiple time trial races.

\noindent \textbf{Super-human Performance in AR}: Recently, \citet{gts} and~\citet{gt_sophy} have reported super-human performance by autonomous racing RL agents in time-trial and actual racing, respectively. \citet{gts} introduced a model-free RL approach and designed a novel proxy reward that considers the progress of the agent in the course. Their method is able to achieve super-human performance in time trial races in Gran Turismo Sport, a highly realistic racing simulation. \citet{gt_sophy} introduced Gran Turismo Sophy (GT Sophy), an RL agent that is able to achieve super-human performance both in time trial and racing scenarios with multiple opponents. To achieve super-human performance both approaches require global features (e.g., forward looking course shape information) at execution time.

\begin{figure*}[t]
    \centering
    \includegraphics[width=\linewidth]{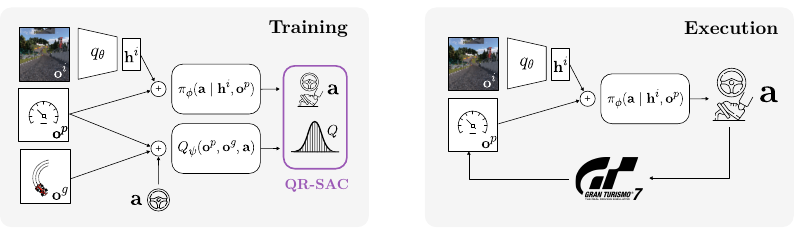}%
\caption{Our vision-based RL agent for autonomous car racing. (Left) We exploit an asymmetric actor-critic architecture to train our agent: the policy network $\pi_{\phi}$ is provided with propriocentric information $\mathbf{o}^p$ and image features $\mathbf{h}^i$, encoded with a convolutional neural network $q_\theta$, to output actions $\mathbf{a}$. The critic network $Q_\psi$ is provided with local propriocentric observations and global observations $\mathbf{o}^g$ (i.e., course shape information) to predict quantiles of future returns. (Right) During execution, our agent only receives local features from the Gran Turismo 7 simulator.}
\label{fig:method}
\end{figure*}

\section{Methodology}
To train a vision-based autonomous racing RL agent that achieves super-human performance without global features at execution time, we design a distributed \emph{asymmetric} actor-critic architecture and employ Quantile Regression Soft Actor-Critic (QR-SAC), a recently introduced distributional RL algorithm~\citep{gt_sophy}. Our method is depicted in Figure~\ref{fig:method}.

\subsection{Observation Space}
We build the multimodal observations $\mathbf{o}$ of our racing agent at time step $t$, following,
\begin{equation*}
    \mathbf{o}_t = (\mathbf{o}^{l}_t, \mathbf{o}^{g}_t),
\end{equation*}
where $\mathbf{o}^{l}_t$ corresponds to local (to the car) features and $\mathbf{o}^g_t \in \mathbb{R}^{531}$ corresponds to the global features (i.e., course shape information). As local features ${\mathbf{o}^{l}_t = (\mathbf{o}^{i}_t, \mathbf{o}^{p}_t)}$ we consider an image $\mathbf{o}^{i}_t \in \mathbb{R}^{64\times64\times3}$ and propriocentric information  $\mathbf{o}^p_t \in \mathbb{R}^{17}$.

\noindent \textbf{Image features} ($\mathbf{o}^{i}_t$): 
At each time step we extract an image directly from the game, considering a first-person view of the track ahead (a camera view denoted by \texttt{Normal view} in the game), from the front of the vehicle. The image is scaled from 1920$\times$1080 (the native resolution of the game) to 64$\times$64, with RGB information. Empirically we found this resolution to be sufficient to allow the agent to race at super-human speeds, as we show in Section~\ref{sec:results:ablation}. Given the low resolution of the image observation, we turned off all extra information on the screen like the heads-up-display (HUD). As an artifact of the simulator, the image observation also includes the car's rear-view mirror. In Appendix~\ref{appendix_additional_ablation_study_masked_rear_view_mirror} we show that our agent can still consistently achieve super-human lap times without any rear-view mirror information. We provide examples of image observations in Figure~\ref{fig:game_images} and in Appendix~\ref{sec:appendix_evaluation_scenarios}.

\noindent \textbf{Propriocentric features} ($\mathbf{o}^p_t$): 
We select features for $\mathbf{o}^p_t$ that can be easily accessible in a real-world autonomous racing scenario,
\begin{equation*}
    \mathbf{o}^p_t = [\mathbf{v}_t, \mathbf{\dot{v}}_t, \mathbf{v}^r_t, \mathbf{c}_t, \mathbf{h}^a_t, \mathbf{h}^d_t],
\end{equation*}
where $\mathbf{v}_t \in \mathbb{R}^3$ corresponds to the linear velocity of the car in its local coordinate system, $\mathbf{\dot{v}}_t \in \mathbb{R}^3$ corresponds to the linear acceleration of the car, $\mathbf{v}^r_t \in \mathbb{R}^3$ corresponds to the angular velocity of the car, $\mathbf{c}_t \in \mathbb{R}^3$ corresponds to the current steering, throttle and brake vector, $\mathbf{h}^a_t \in \mathbb{R}^3$ corresponds to a short history of the steering angles in the last three steps and $\mathbf{h}^d_t \in \mathbb{R}^2$ corresponds to the delta steering changes in the last three steps. The velocity and acceleration features can be estimated using inertial measurement units (IMU), which are often included in real autonomous vehicles~\citep{betz2022autonomous}, and the steering features can be easily extracted from the car's guidance system.

\noindent \textbf{Global features} ($\mathbf{o}^{g}_t$): Following~\citet{gt_sophy}, we explore \emph{course point information} as global features. Course points are built using the shape of the track, including track limits of the left and right, and a center line of the track. At each time step, the range of the course points is a function of the current velocity of the vehicle: we consider the 3D relative coordinates of the course points ahead of the agent from 0.1 sec up to 6 sec ahead (maintaining the current velocity), equally spaced on 0.1 sec intervals. This results in 59 course points for each course line (left, center and right). In Appendix~\ref{appendix_additional_ablation_study}, we evaluate the effect of course point range on the performance of our agent.

\subsection{Action Space}
Similarly to \citet{gts,gt_sophy}, we define the actions of our agent $\mathbf{a}_t \in \mathbb{R}^2$, consisting of a \emph{delta steering angle} and a \emph{combined throttle and brake} value. The delta steering angle at a single time step is limited within $[-3\degree, +3\degree]$ to prevent steering changes from exceeding human limitations.
The combined throttle and brake is represented by a normalized scalar in $[-1, +1]$. Values in the positive range represent throttle and ones in the negative range correspond to brake. The gear shift of the vehicle is controlled by automatic transmission. We set the control frequency to 10 Hz and the game, running at 60 Hz, linearly interpolates the steering angle between steps.

Due to technical constraints when retrieving images from the game in real-time, we utilized a synchronous communication process between the game and our agent, instead of asynchronous communication. This mode ensures alignment between the image and propriocentric features. In this mode our agent does not execute its policy in real-time during training due to the synchronicity of the simulator. However, we show in Appendix~\ref{appendix_additional_ablation_study_sync} that executing the trained policies asynchronously, i.e., in real-time, still allows our agent to achieve super-human performance. In Appendix~\ref{sec:appendix_model}, we provide more details regarding our communication configuration.

\subsection{Reward Function}
\label{sec:reward}
Following \citet{gt_sophy}, we designed the reward function of the agent as the weighted combination of multiple components,
\begin{equation*}
    r_t = r^{p}_t + \lambda^o r^{o}_t + \lambda^w r^{w}_t + \lambda^s r^{s}_t + \lambda^h r^{h}_t.
\end{equation*}

\noindent \textbf{Course progress} ($r^{p}$): 
We formulate the lap time minimization problem as a course progress maximization problem: we compute the progress of the vehicle position, projected onto the center line of the track, since the last step;

\noindent \textbf{Off-course penalty} ($r^{o}$): We define a shortcut penalty to prevent the agent from violating racing rules by cutting corners, ${r^{o}_t = -(s^{o}_t - s^{o}_{t-1})|\mathbf{v}_t|}$, where $s^{o}$ is the total time that the vehicle had (at least) three tires outside the track limits.

\noindent \textbf{Wall penalty} ($r^{w}$): We define a wall-hit penalty to prevent the agent from exploiting walls to quickly change its direction of movement, $r^{w}_t = -(s^{w}_t - s^{w}_{t-1})|\mathbf{v}_t|$, where $s^{w}$ is the total time the vehicle was in contact with a wall in the track.

\noindent \textbf{Steering change penalty} ($r^{s}$):  To discourage large changes of steering angles in a single step, we define a steering change penalty, ${r^{s}_t = -|\theta^{s}_t - \theta^{s}_{t-1}|}$ where $\theta^{s}_t$ is the steering angle in radian at time step $t$;

\noindent \textbf{Steering history penalty} ($r^{h}$):  We additionally define a steering history penalty to discourage the agent to make inconsistent decisions in a short period of time,
\begin{equation*}
    r^{h}_t = -m_t \cdot 1 / (1 + \exp(-c^{s} \cdot (\Delta_t - c^{o}))),
\end{equation*}
where $\Delta_t = |\delta_t| + |\delta_{t-1}|$, $\delta_t = \theta^{s}_t - \theta^{s}_{t-1}$, $m_t = \mathbb{I}_{\delta_t > c^{d}} \cdot \mathbb{I}_{\delta_{t-1} > c^{d}} \cdot \mathbb{I}_{\text{sgn}(\delta_t) \neq \text{sgn}(\delta_{t-1})}$, $c^d$ is a threshold angle, $c^{s}$ is a sensitivity factor and $c^{o}$ is an offset value. In Appendix~\ref{sec:appendix_training} we provide the reward function parameter values used in our evaluation.

\subsection{Training Architecture}

We train our agent using QR-SAC, a distributional RL extension to Soft Actor-Critic~\citep{sac} with multi-step TD error. In QR-SAC, critic functions are represented with a quantile distribution function~\citep{qr_dqn} that estimates quantiles of returns.

To achieve super-human performance at execution time, inspired by recent works in RL under partial observability~\citep{pinto2017asymmetric, salter2021attention,sinha2023asymmetric,baisero2022asymmetric}, we consider an \emph{asymmetric} actor-critic architecture for QR-SAC training, as shown in Figure~\ref{fig:method}. During training, the critic functions are provided with global features $\mathbf{o}^g$, instead of image observations $\mathbf{o}^{i}$, allowing them to learn accurate returns. The policy is only provided with image and propriocentric features $\mathbf{o}^p$. Since the policy does not depend on the course points to predict actions, the agent is able to race at execution time only with local observations. We detail our model architecture in Appendix~\ref{sec:appendix_model} and our training hyperparameters in Appendix~\ref{sec:appendix_training}.

\section{Evaluation} \begin{wrapfigure}{r}{0.5\linewidth}
\vspace{-9.7ex}
\centering
    \begin{minipage}[b]{0.3\linewidth}
        \centering
        \includegraphics[width=\linewidth]{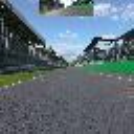}%
    \end{minipage}%
    \hfil
    \begin{minipage}[b]{0.3\linewidth}
        \centering
        \includegraphics[width=\linewidth]{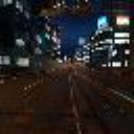}
    \end{minipage}%
    \hfil
    \begin{minipage}[b]{0.3\linewidth}
        \centering
        \includegraphics[width=\linewidth]{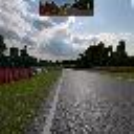}
    \end{minipage}%
\caption{Examples of 64$\times$64 image observations in (left) \texttt{Monza}, (middle) \texttt{Tokyo}, and (right) \texttt{Spa}.}
\vspace{-4ex}
\label{fig:game_images}
\end{wrapfigure} 
We evaluate our agents in time trial tasks, where the goal is to complete a lap across the track in the minimum time possible. In this section we cover the track and car scenarios used for testing, the racing baselines, and the human player data used for comparisons.

\begin{table*}[t]
\renewcommand{\arraystretch}{1.2} 
\setlength\arrayrulewidth{0.1pt} 
\centering
\caption{Time trial scenarios for the evaluation of our vision-based racing agent. We evaluate our approach across different tracks, cars and track conditions. We also consider different tire settings: racing soft (RS), sport soft (SS) and sport medium (SM). We compare our agent against over 130K human players in each scenario.}
\label{table:scenario}
\resizebox{\textwidth}{!}{%
\begin{tabular}{llllcc}
\toprule
 Scenario &  Track & Condition & Car & Tire & Number of participants \\
\midrule
 {\ttfamily Monza} & Autodromo Nazionale Monza, Italy & Day, Clear & Ferrari 330 P4 '67 & RS & 138,306 \\
 \arrayrulecolor{gray!20}\hdashline\arrayrulecolor{black}
 {\ttfamily Tokyo} & Tokyo Expressway - Central Clockwise, Japan & Night, Clear &  NISMO 400R '95 & SS & 131,598 \\
 \arrayrulecolor{gray!20}\hdashline\arrayrulecolor{black}
 {\ttfamily Spa} & Circuit de Spa-Francorchamps, Belgium & Day, Cloudy & Alfa Romeo 4C Launch Edition '14 & SM & 144,308 \\
\end{tabular}
}
\end{table*}

\subsection{Scenarios}
We evaluate our agent in three scenarios in GT7 with different combinations of cars, tracks, and conditions (track time and weather): \texttt{Monza}, \texttt{Tokyo}, and \texttt{Spa}, modeled after real-world circuits and roads. These scenarios were selected based on past GT7 online race events, where human players joined time trial races using the exact same car setup as our approach\footnote{ For more details regarding the online race events, refer to \url{https://www.gran-turismo.com/us/gt7/sportmode/}.}.We provide a more detailed description of the evaluation scenarios in Table~\ref{table:scenario} and image observations of the different scenarios in Figure~\ref{fig:game_images} and in Appendix~\ref{sec:appendix_evaluation_scenarios}.

\subsection{Baselines}

\noindent\textbf{GT Sophy}~\citep{gt_sophy}: We use GT Sophy, a recently introduced super-human racing agent for Gran Turismo, as a baseline in our experiments. As this baseline was shown to be able to outperform the best human drivers and exploits global features to act, we consider its performance as an upper-bound to the performance of our method. We modify the action space of GT Sophy, which originally outputs absolute steering angles rather than delta angles, to match the action space of our agent. Moreover, we use the same training hyperparameters and reward function as our method. We train this baseline for GT7 using the same training method described in~\citet{gt_sophy}.

\noindent\textbf{Human Players}:
Human player data was provided by \textit{Polyphony Digital Inc.}, the development studio of Gran Turismo. For each scenario we collected over 130K lap times and trajectories. We consider our agent to have super-human performance if it is able to achieve a faster lap time than the one achieved by the fastest human player in each scenario.

\noindent \textbf{Built-in AI}:
The built-in AI of GT7 follows a pre-defined human expert trajectory using a rule-based tracking approach, similar to MPC methods, and serves as a traditional control-based baseline. We report the minimum lap time of the built-in AI after executing 4 laps in each scenario.

\begin{figure}[t]
\centering
\bgroup
\renewcommand{\arraystretch}{1.2} 
\setlength{\arrayrulewidth}{0.1pt} 

\begin{tabular}{lccc}
        \toprule
        Method & \multicolumn{1}{c}{{\ttfamily Monza}} & \multicolumn{1}{c}{{\ttfamily Tokyo}}  & \multicolumn{1}{c}{{{\ttfamily Spa}} } \\
        \midrule
         Built-in AI & 107.828 \spm{\text{\,\textemdash}}\phantom{0.} & 87.905 \spm{\text{\,\textemdash}}\phantom{0.} & 168.280 \spm{\text{\,\textemdash}}\phantom{0.} \\

        \arrayrulecolor{gray!20}\hdashline\arrayrulecolor{black} 
        Fastest Human & 104.378 \spm{\text{\,\textemdash}}\phantom{0.} & 80.782 \spm{\text{\,\textemdash}}\phantom{0.} & 157.796 \spm{\text{\,\textemdash}}\phantom{0.} \\

        \arrayrulecolor{gray!20}\hdashline\arrayrulecolor{black}
        GT Sophy~\citep{gt_sophy} & \textbf{104.281} \spm[bold]{0.061} & \textbf{80.227} \spm[bold]{0.047} & \textbf{157.424} \spm[bold]{0.038}\\ 
         
        \arrayrulecolor{gray!20}\hdashline\arrayrulecolor{black}
         \rowcolor{lightblue} Our Agent &  \textbf{104.300} \spm[bold]{0.050} & \bfseries 80.401 \bfseries \spm[bold]{0.091} & \bfseries 157.554 \bfseries \spm[bold]{0.055} \\ 
    \end{tabular}
\egroup
\vspace{2ex}
\begin{minipage}{0.32\textwidth}
  \centering
  \includegraphics[width=0.99\linewidth]{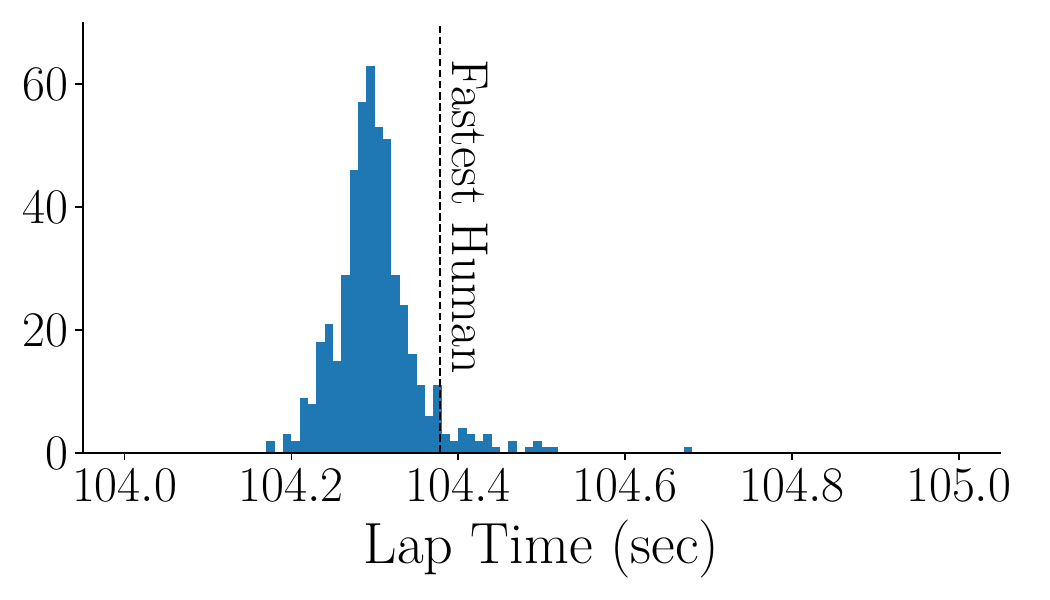}
\end{minipage}
\hfill
\begin{minipage}{0.32\textwidth}
  \centering
  \includegraphics[width=0.99\linewidth]{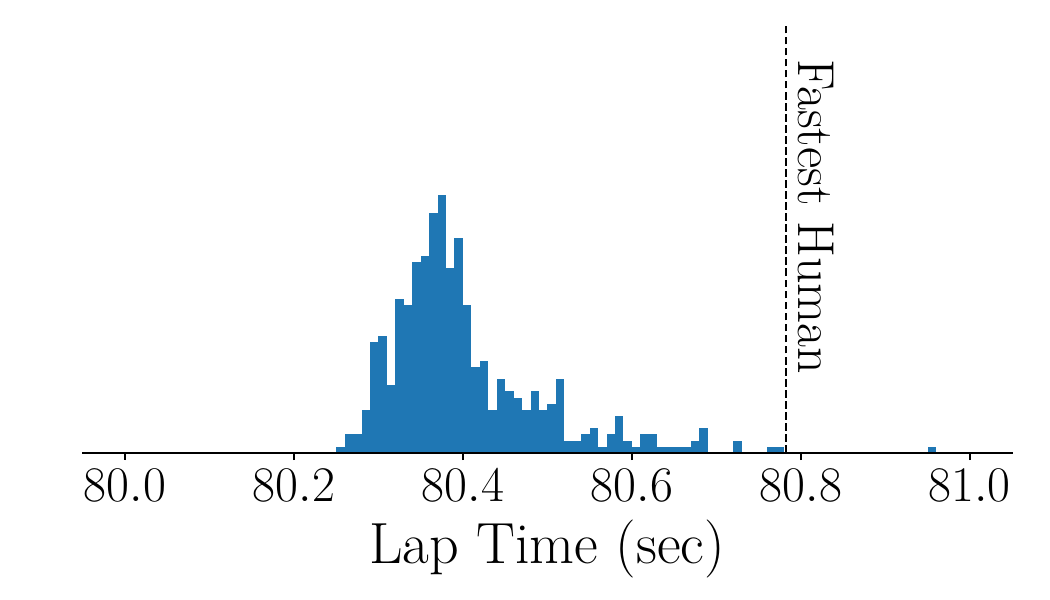}
\end{minipage}
\hfill
\begin{minipage}{0.32\textwidth}
  \centering
  \includegraphics[width=0.99\linewidth]{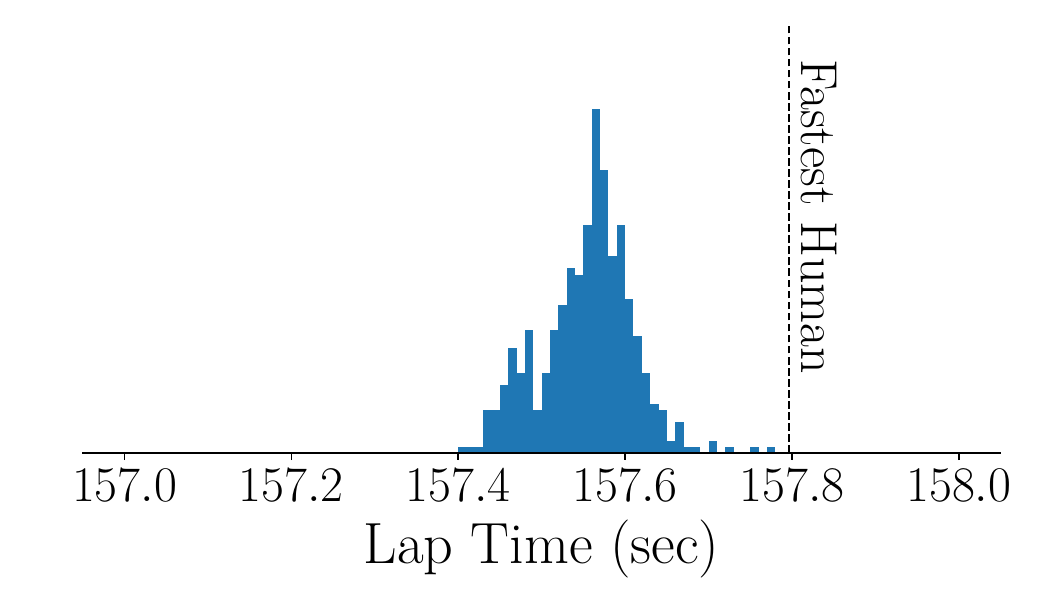}
\end{minipage}
\caption{(Top) Lap time across all scenarios. We consider five randomly-seeded training runs and average the results over 500 evaluation laps, with 100 laps executed by the fastest policy in each training run.  We highlight results that are significantly faster than the fastest human time (using a Wilcoxon signed-rank test, with $p<0.001$); (Bottom) Distribution of lap times in \texttt{Monza} (left), \texttt{Tokyo} (middle) and \texttt{Spa} (right).}
\label{table:laptime}
\end{figure}

\section{Results}
\label{sec:results}

We show the minimum lap times achieved by the different agents in Figure~\ref{table:laptime}. In all three scenarios, our agent achieves super-human performance, with lap times that significantly surpass the performance of the best human player. Our agent also achieves comparable performance to GT Sophy, despite not having any global features at execution time. Our agent achieves this level of performance \emph{consistently}, with small variation across the randomly-seeded runs, as shown in the training curves in Appendix~\ref{appendix_training_curves}, and across the  different evaluation laps: for \texttt{Monza} we outperform the fastest human player in 94.0\% of the laps, for \texttt{Tokyo} in 99.8\% of the laps and in \texttt{Spa} in all the laps. We note that the distribution in lap times is a result of the high-fidelity physics engine of the simulator, where small numerical differences can result in different behaviors, thus preventing the agent from repeating the same trajectory across multiple laps.

\begin{figure}[t]
\centering
\begin{minipage}{0.31\textwidth}
  \centering
  \includegraphics[width=0.99\linewidth]{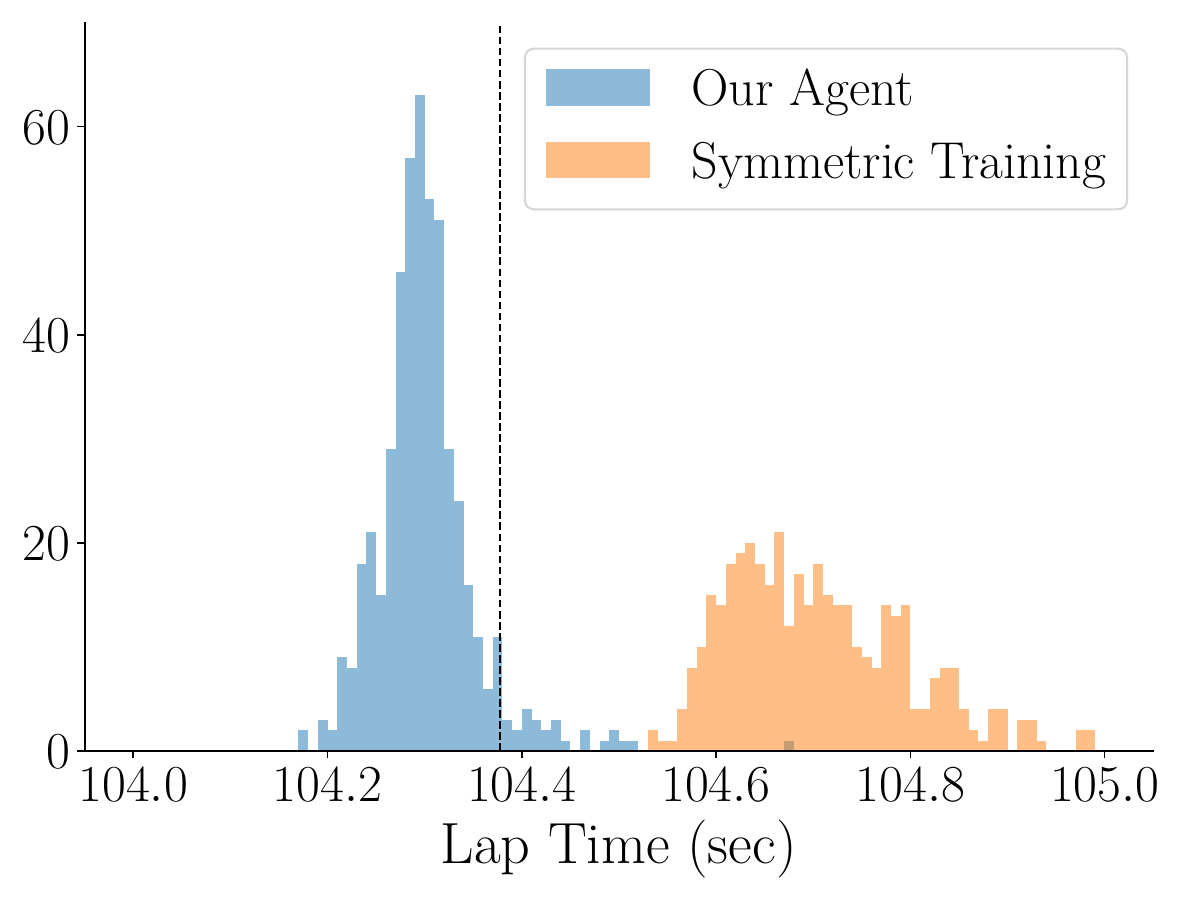}
\end{minipage}
\hfill
\begin{minipage}{0.31\textwidth}
  \centering
  \includegraphics[width=0.99\linewidth]{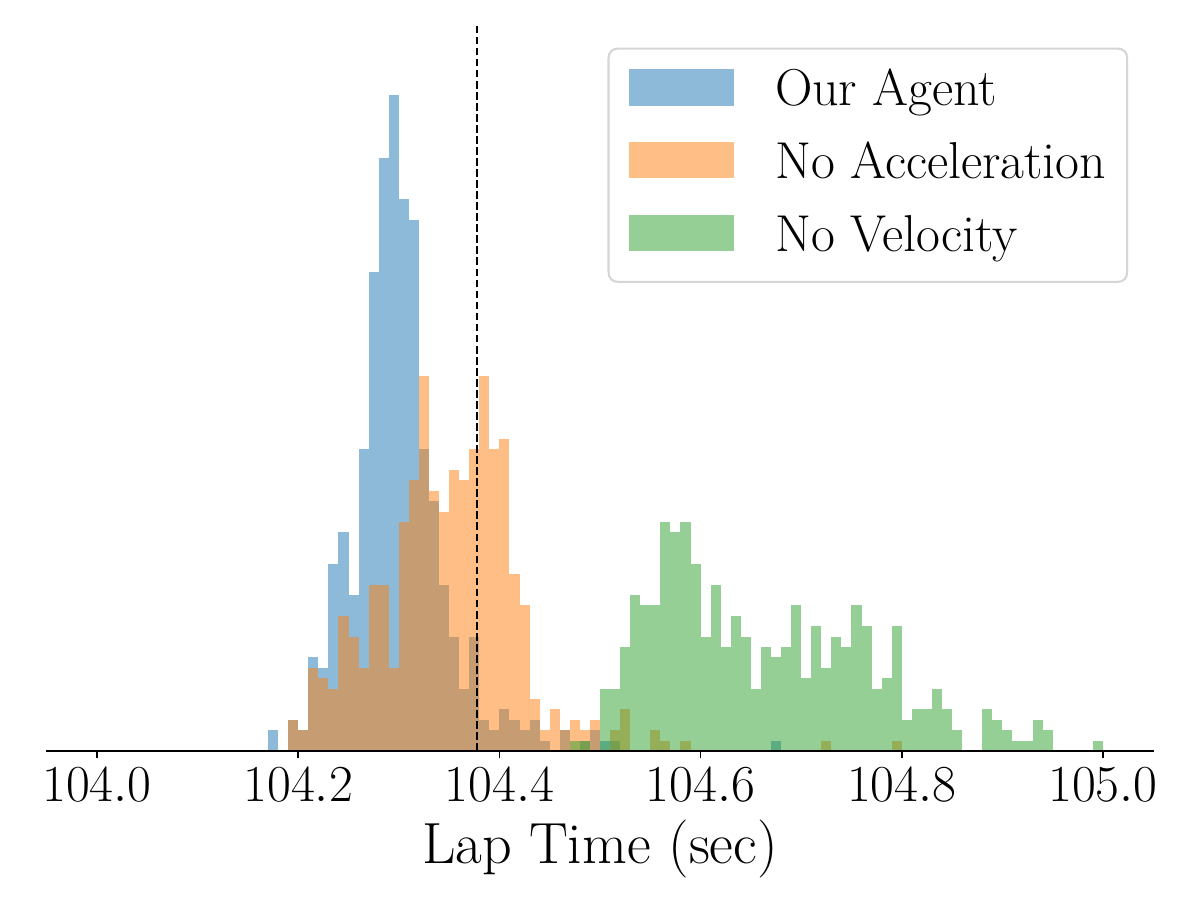}
\end{minipage}
\hfill
\begin{minipage}{0.31\textwidth}
  \centering
  \includegraphics[width=0.99\linewidth]{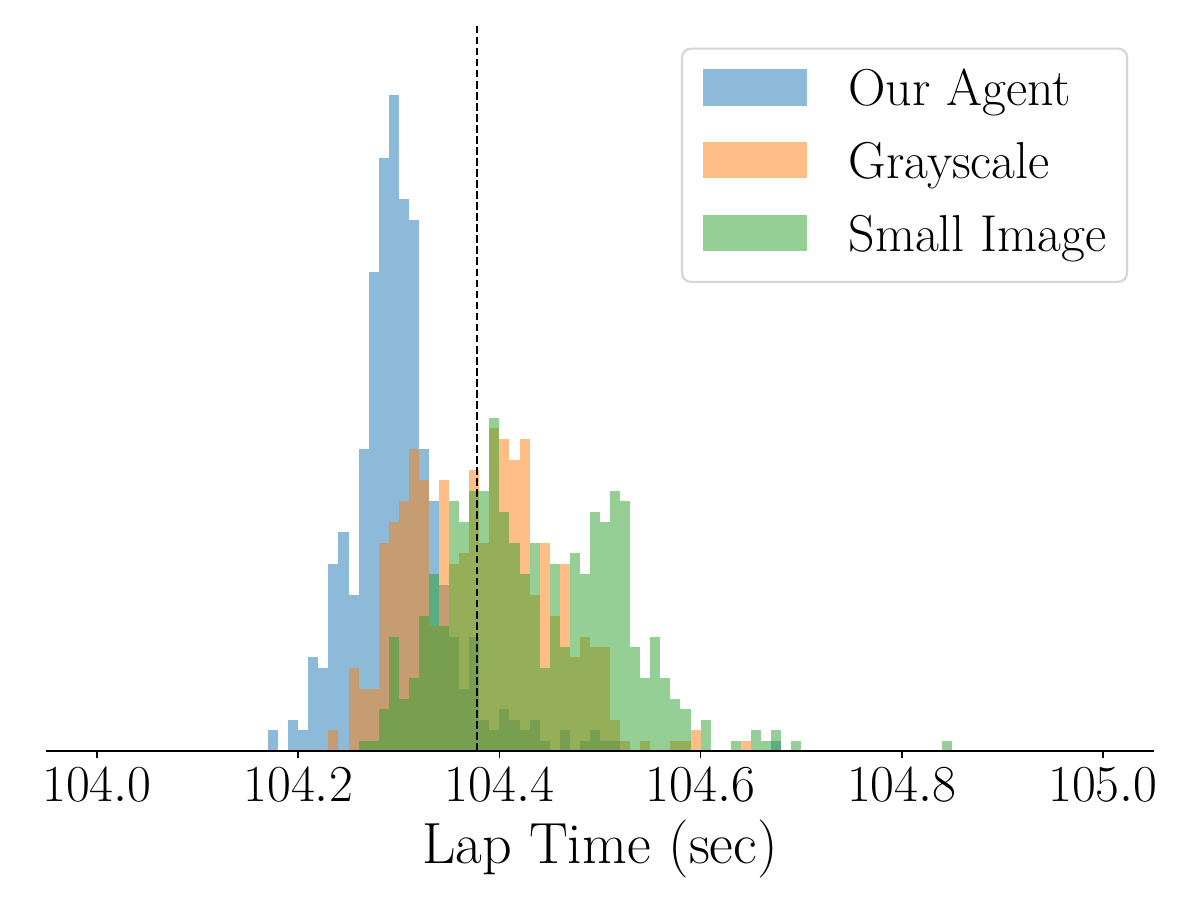}
\end{minipage}
\caption{Performance study of our racing agent in the \texttt{Monza} scenario in relation to the training architecture (left), local features (middle) and the image feature (right). We consider five randomly-seeded training runs and show the distribution of 500 evaluation laps, with 100 laps executed by the fastest policy in each training run. We highlight the lap time of the fastest human player (black line). One symmetric run failed to learn meaningful behavior and we exclude it from the analysis.}
\label{fig:ablation_study}
\end{figure}

\subsection{Ablation Study}
\label{sec:results:ablation}

We define ablated versions of our method to evaluate the contribution of different architectural and training choices to the performance of our method, in particular regarding (i) the training architecture, (ii) local features, and (iii) the image feature. For (i) we employ a symmetric training scheme, where we replace the course points in the critic's input with image observations (\emph{symmetric training}). For (ii) we remove acceleration features, (\emph{no acceleration}) and velocity features (\emph{no velocity}) from $\mathbf{o}^{p}$, and remove image features (\emph{no image}) from $\mathbf{o}^{l}$. For (iii) we remove color from the image observation (\emph{grayscale}) and reduce the size of the image\begin{wrapfigure}{r}{0.5\linewidth}
 \vspace{-2ex}
    \centering
\includegraphics[width=\linewidth]{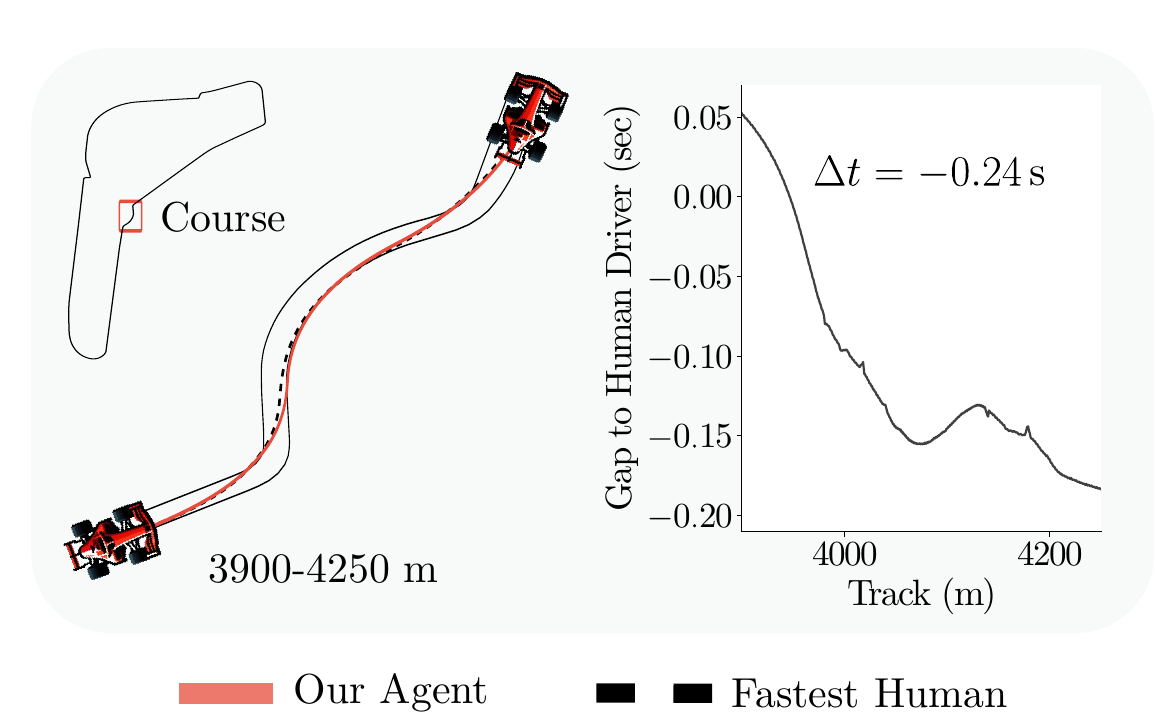}%
\caption{(Left) Trajectory comparison in the \texttt{Monza} track between our agent and the fastest human player in a chicane section. (Right) Gap of our agent to the human driver. Lower is better.}
\vspace{-5ex}
\label{fig:trajectories}
\end{wrapfigure}  observation to 32$\times$32 (\emph{small image}).

We present the results of our ablation study in Figure~\ref{fig:ablation_study}. Regarding (i) the results show that providing the critic with global features during training is fundamental for the performance of our agent as it mitigates the partial observability of the environment, allowing for a better estimate of the returns of the policy. Regarding (ii) we observe that removing velocity features results in a decrease in the performance of our agent as, naturally, velocity information is fundamental to racing at a consistently high level around the track. Regarding (iii), the results highlight that both color and a larger size of the image helps improve the performance of our method. Additionally, we found that the \emph{no image} agent is unable to drive. As such, we do not present this condition in Figure~\ref{fig:ablation_study}. This result further highlights the importance of visual information for our agent. In Appendix~\ref{appendix_additional_ablation_study} we report on additional studies regarding the range of course points for the critic's input, revealing that we can further improve the performance of our agent by fine-tuning this parameter; and on the synchronous communication of our training pipeline, highlighting that we can execute our trained policies in an asynchronous (i.e., real-time) version of our simulator without significant loss in performance.

\subsection{Qualitative Policy Study}
\label{sec:results:policy}
 We qualitatively evaluate the policy of our agent with regards to its trajectory against the best human drivers and the importance of visual features for the decision-making of our agent.

\noindent\textbf{Trajectory Analysis:} In Figure~\ref{fig:trajectories} we compare the trajectories of our agent and of the best human player in a chicane section. Our agent takes a driving line closer to the track limits, slowing down only 16.5\% of what the human driver slows down in the section, thus gaining 0.24 seconds. The novel racing behavior exhibited by our agent motivates its use as a training tool for human drivers. We note that while the agent is able to achieve super-human lap times, it is not faster than the best human player across all segments of the track. We present an extended version of this study, including comparisons to GT Sophy, across all scenarios in Appendix~\ref{appendix_trajectory_analysis}.

\noindent\textbf{Visual Analysis:} We employ Guided Gradient-weighted Class Activation Mapping (GGC)~\citep{selvaraju2017grad}, a visual analysis tool for image-based classification tasks~\citep{arrieta2020explainable,linardatos2020explainable}, to understand what high-level features in the input image are relevant to the decision-making of our agent. We modify the original algorithm for RL tasks, as described in Appendix~\ref{section:gradcam_theory}. In Figure~\ref{fig:ggc_analysis}, we present GGC visualizations for the steering action of our agent in the \texttt{Monza} track.

The results show a distinct pattern of behavior for different sections in the track. In long straights, far-away visual features, such as horizon of the track or the tree line, are more significant for the policy of our agent than close visual features, such as the curb of the track. Naturally, in these sections, the agent is travelling at high-speeds and mostly needs to focus on identifying where the straight ends. However, in chicanes and tight curves, our agent focuses on the closer curbs of the track which are fundamental to successfully change its direction without going off-track and incurring on a penalty. This \emph{gaze-like} behaviour echoes the one exhibited by human drivers~\citep{rito2020neurobehavioural, van2017differences}: during straight segments, the human eye gaze focuses straight ahead, with a stable distance in the horizon, and during curves the eye gaze is focused on the inner tangent (apex) of the curve. Additionally, our agent uses the uniqueness of the visual features in the track to localize itself: we see that it considers both forward features (track limits and horizon) and backward features (rear-view mirror) in its decision-making. We consider that the focus on the rear-view mirror indicates that the trained policy exploits the static track layout for localization. We provide additional visualizations for all scenarios in Appendix~\ref{sec:appendix_additional_ggc}.

\begin{figure*}[t]
    \begin{minipage}[b]{0.49\linewidth}
        \centering
        \includegraphics[width=0.95\linewidth]{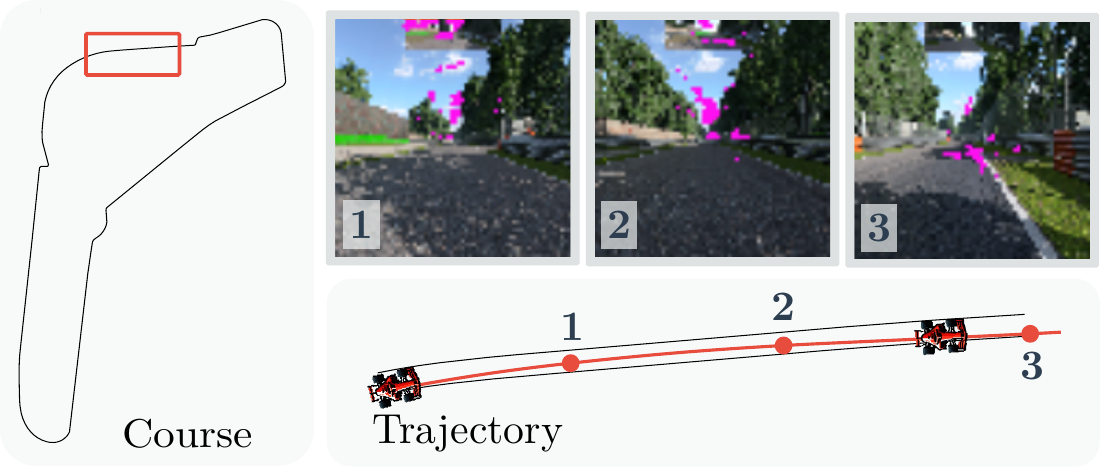}%
    \end{minipage}%
    \hfil
    \begin{minipage}[b]{0.49\linewidth}
        \centering
        \includegraphics[width=0.95\linewidth]{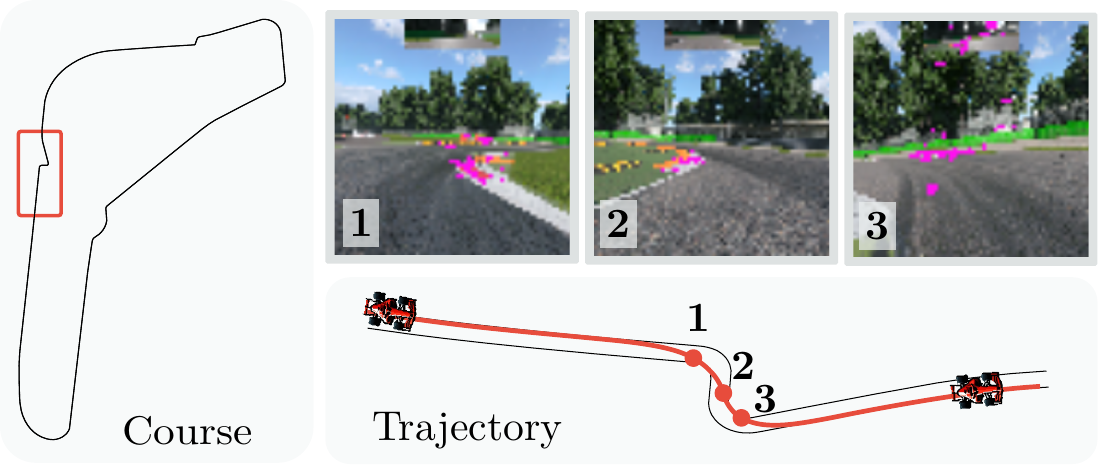}%
    \end{minipage}%

\caption{GGC visualization of our racing agent for two sections of the \texttt{Monza} track: (left) a straight section; (right) a chicane section. (Top) We show in pink the positive gradients for the delta steering angle action computed using the policy of our agent. We show the top 80\% of the gradients in the visualization, to reduce noise. We highlight two different behaviors: in straight sections, our agent focuses on far-away visual features, such as tree lines (Left: 1, 2) and distinct far-away shades (Left: 3); in chicane sections, our agent focuses on close elements that are fundamental to effectively change direction, changing its focus from the apex of the immediate curve (Right: 1, 2) to the curb on the opposite side (Right: 3). Best viewed with color and zoomed in.}
\label{fig:ggc_analysis}
\end{figure*}

\subsection{Perturbation Study}

We conduct an extensive evaluation of the robustness of our agent to input perturbations and different track/car conditions in Appendix~\ref{sec:results:generalization}. Amongst other results, the study highlights the importance of visual information for our agent: (i) changes in the lighting conditions of the environment (due to racing at a different time of the day) degrade significantly the performance of our agent; (ii) adversarial perturbations to the image observation, in particular to the image features computed using GGC, also degrade significantly the performance of our agent.

\section{Conclusions}
In this paper we presented the first super-human, vision-based reinforcement learning agent for autonomous car racing. To achieve this level of performance, we leverage an asymmetric actor-critic architecture that uses global features from the simulator to train accurate critic functions, while the policy function only uses local features to output actions at execution time. We demonstrated that our approach surpassed the fastest human lap time in three time trial scenarios and showed comparable performance to super-human methods that require global features for its policy. We hope our proposed approach helps to build the foundations for a novel research field on competitive autonomous racing agents with car-centric input features.

For future work, we consider three research threads to address the limitations of this paper.
First, we plan on extending our approach to racing scenarios with multiple vehicles in the track, in order to allow vision-based autonomous agents to race against human drivers in the same track.
Second, although we showed that our asymmetric architecture allows us to train super-human agents with a simple deep RL training setup, we still use propriocentric information as inputs of our agent.
To relax this necessity, we will explore incorporating recurrent neural networks, similarly to \citet{wadekar2021towards}.
Finally, we plan to add generalization capabilities to our agent, which deals with conditions unseen during training. We can extend our training setup to include various tracks and car models with additional image data augmentations~\citep{kostrikov2020image} to mitigate this issue, and eventually transfer the trained agent to real-world racing vehicles.

\subsubsection*{Broader Impact Statement}
\label{sec:broaderImpact}
We focused on evaluating our agent in a high-fidelity simulator in this paper. However, our research can also contribute to the development of real-world end-to-end autonomous race cars. Using car-centric inputs, agents can control vehicles without using external localization systems that usually require domain knowledge beforehand or expensive engineering costs to design. By extending our agent to real world setups, we could simplify the pipeline of autonomous vehicles.

\subsubsection*{Acknowledgments}
\label{sec:ack}
We are very grateful to Polyphony Digital Inc. and Sony Interactive Entertainment for enabling this research. Furthermore, we would like to thank our colleagues in Sony AI, Songyang Han, Hojoon Lee, Craig Sherstan, Florian Fuchs, and Patrick MacAlpine, for their feedback on this manuscript. The first author also acknowledges that this work has been partially supported by the European Research Council (ERC-BIRD), Swedish Research Council and Knut and Alice Wallenberg Foundation.


\bibliography{main}
\bibliographystyle{rlc}

\newpage
\appendix

\section{Extended Related Work}
\label{sec:appendix_extended_related_work}

Autonomous Racing (AR) is a subfield of autonomous driving research~\citep{yurtsever2020survey} that concerns autonomous vehicles that operate at their dynamical and power limits within racing environments (such as racing circuits)~\citep{betz2022autonomous}. 

\textbf{Classical Approaches for AR} Research in autonomous vehicle racing can traditionally be categorized into three different sub-areas: perception, planning and control. In \emph{perception}, the overarching goal is to enable high-frequency object detection, mapping and localization while the vehicle is driving around the track at high-speeds: \citet{massa2020lidar} propose a LIDAR-based localization system that exploits a previously built map of the track environment to provide localization, achieving an accuracy of two meters when the car is moving at 200 km/h; \citet{peng2021vehicle} contributes a multimodal odometry method (image, LIDAR, IMU) using factor-graph optimization to localize the vehicle in the track; \citet{strobel2020accurate} use YOLOV3~\citep{redmon2018yolov3} to detect light cones in the limits of the racing track for Formula Student competitions. In \emph{planning}, the overarching goal is to plan spatial and velocity trajectories (global and local) that minimize lap time across the track: \citet{herrmann2020minimum} formulate an optimization-based velocity planner as a multi-parametric sequential quadratic problem that can handle a spatial and time variable friction coefficient; \citet{vazquez2020optimization} propose a hierarchical controller for autonomous racing, where the high-level controller computes the optimal trajectory in the race track (raceline) and the low-level controller attempts to follow the precomputed optimal trajectory; other approaches attempt to plan high-level behavior (such as overtaking maneuvers, or energy management during a race) either by assigning plans to a specific cost function and selecting the plan with the lowest overall cost~\citep{liniger2015viability,sinha2020formulazero,o2020tunercar} or by combining the planner with game theoretical methods~\citep{notomista2020enhancing,schwarting2021stochastic,liniger2019noncooperative}. In \emph{control}, the overarching goal is to develop methods that are able to maintain the vehicle as close as possible to the planned spatial trajectory and speed profile. For this purpose, model predictive control (MPC) methods are widely employed: \citet{williams2018robust} propose a robust sampling-based MPC framework based on a combination of model predictive path integral control and nonlinear Tube-MPC~\citep{mayne2005robust}, highlighting the framework's robustness in a real-world autonomous racing task; \citet{gandhi2021robust} contribute a novel architecture for robust model predictive path integral control (RMPPI) and investigate its performance guarantees, highlighting its applicability in a real-world off-road navigation task; \citet{li2021autonomous} propose a nonlinear MPC model under a minimum time objective, which integrates
the opponent vehicle’s trajectory as a collision-avoidance constraint, to allow racing tasks with opponents. In our paper, contrary to all previous works, we explore end-to-end RL for autonomous racing vehicles that combines the pipeline of perception-planning-control into a single process.

\textbf{Deep Neural Networks for AR} Recent developments in deep neural networks (DNN) have allowed the development of end-to-end methods that are able to learn to race directly from observations. \citet{wadekar2021towards} explore different types of data collection techniques to train DNNs to output steering and throttle actions in a supervised learning manner. \citet{mahmoud2020optimizing} highlight that reducing the image size in CNN-based networks leads to an increase in the performance of a DNN-based racing method both in simulation and in the real-world. Contrary to these works, we focus in RL approaches that learn to perform racing tasks through trial-and-error interaction with the environment.

\textbf{Reinforcement Learning for AR} RL has also been shown to be a promising approach to learn suitable racing behavior, motivated in part by the development of realistic driving simulators that are able to model the dynamics of the car and of the track~\citep{o2019f1,herman2021learn,rong2020lgsvl,dosovitskiy2017carla}. Methods that employ RL to train racing agents often provide both image observations and additional features, that can be either propriocentric (e.g., velocity, acceleration) or related to the track (e.g., center line of the track). \citet{jaritz2018end} explore vision-based RL for driving agents in the context of a rally game. The propose an Asynchronous Advantage Actor-Critic (A3C)~\citep{mnih2016asynchronous} architecture that exploits both visual information (first-person camera images) and propriocentric features (velocity and acceleration) to learn to race in the game environment. The authors show that their approach is able to generalize to unseen tracks and highlight the importance of initializing the agents at random checkpoints during training for the performance of the method. However, the authors state that their method does not achieve optimal racing trajectories, ``lacking anticipation'', and is unable to complete the racing tracks without colliding several times with obstacles. Recently, \citet{cai2021vision} described an approach that merges imitation learning and model-based reinforcement learning to learn autonomous racing agents. Central to their contribution is the Reveries-Net architecture to learn a probabilistic world model~\citep{ha2018world}. The authors propose an iterative training procedure after pretraining the policy with expert demonstrations: (i) learn the world model with the current dataset of experiences, (ii) refine the policy using rollouts from the world model and (iii) collect new data using the refined policy in the environment. The authors evaluate their approach in simulation and real-world racing environments and demonstrate that their method outperforms previous imitation learning and RL methods in sample efficiency and performance. However, their method requires expert-level demonstrations to pretrain the policy and cannot be trained only from interaction with the environment. Despite their reported ability to consistently drive around the track, the racing performance of these methods is still sub-optimal: some works lack a performance comparison against humans~\citep{remonda2021formula,jaritz2018end} or, when such comparison is made, the methods still under perform significantly against median human users~\citep{cai2021vision,herman2021learn}: for example, \citet{cai2021vision} reports a 10 second gap to the lap time of a normal human user. \citet{gts_vision} also explores vision-based RL for racing agents. The authors propose to pretrain an image encoder on observations of the track environment collected by a random policy and, subsequently, use the frozen encoder during policy training. However, their method can only achieve expert-level performance in time trial tasks, still reporting a three second difference to the best human players. In this work, we contribute a novel vision-based RL agent that consistently outperforms the best human drivers in a racing task.

\begin{figure}[t]
\centering
\includegraphics[width=.8\linewidth]{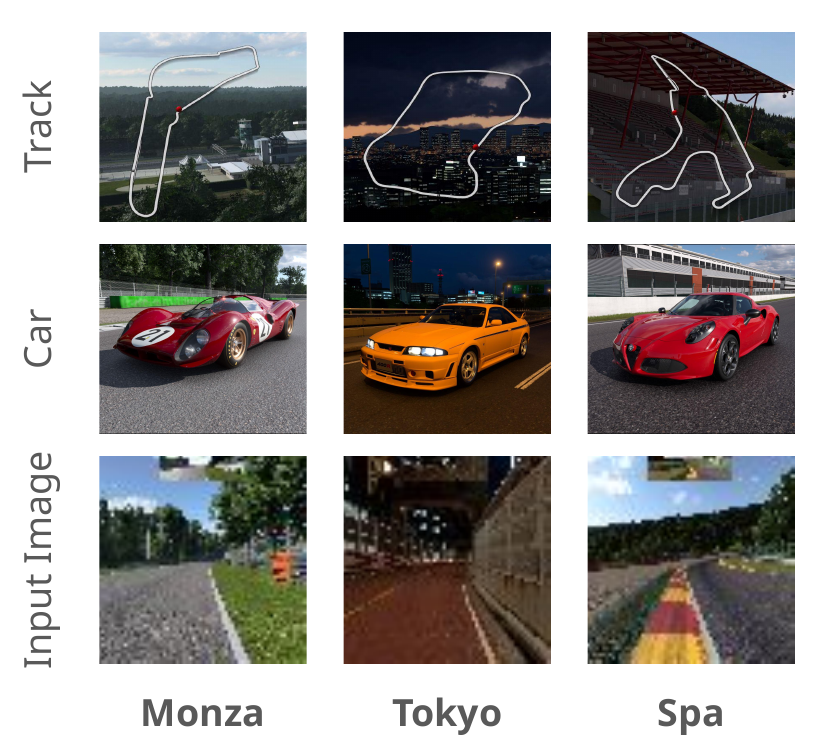}
\caption{Description of the tracks, cars and examples of image observations used to train and evaluate our racing agent. We rescale all images to 64$\times$64, without any modification. We turn off all HUD information. The top notch in images is a rear mirror view of the vehicle.}

\label{fig:scenarios}
\end{figure}

\textbf{Super-human Racing Performance in AR} Recently, super-human performance of autonomous racing RL agents have been reported by~\citet{gts} and~\citet{gt_sophy}. \citet{gts} introduced a model-free RL approach and designed a novel proxy reward that considers the progress of the agent in the course. The method is able to achieve super-human performance in time trial tracks in Gran Turismo Sport (GTS), a highly realistic racing simulation. Moreover, the authors show that their approach generates trajectories that are qualitatively similar to the ones recorded by the best human drivers, highlighting high-level racing behavior (such as in-out driving along curves). More recently, \citet{gt_sophy} introduced Gran Turismo Sophy (GT Sophy) agent, a RL agent that is able to achieve super-human performance both in time trial and racing tasks with multiple opponents in GTS. To achieve this level of performance, the authors contribute a novel asynchronous distributional actor-critic algorithm (QR-SAC) using multiple training scenarios (e.g., with different number of opponents, with randomized positions and speeds). Furthermore the authors designed a novel reward function that accounts for track-related behavior (e.g., progress in the course, off-course racing) and for event-related behavior (e.g., overtaking opponents or being overtaken, colliding with opponents). The authors show that GT Sophy is able to exhibit tactical skills that allow it to beat expert humans in head-to-head racing. However, to achieve super-human performance both approaches require \emph{global features}, such as forward looking course shape information. In this work, we contribute the first RL agent that is able to achieve super-human performance in a racing task using only car-centric local features during execution.

\textbf{Asymmetric Reinforcement Learning} Recent works have explored asymmetrical training architectures for reinforcement learning to mitigate partial observability during execution time~\citep{pinto2017asymmetric,salter2021attention,kamienny2020privileged, sinha2023asymmetric,baisero2022asymmetric}. \citet{pinto2017asymmetric} explored asymmetrical training in the context of learning policies in simulation for robotic systems that are transferable to real-world setups. To do so, the authors design an asymmetrical actor-critic training scheme in which the critic is provided with the state of the simulation environment and the policy is provided with RGBD information. Furthermore, the authors introduce domain randomization during training in the simulator, showing that it improves the robustness of the learned policy to distractor elements during execution in real-world experiments. However, domain randomization can often lead to an increase the complexity of the learning process, impacting the overall performance of the agent. To address this issue \citet{salter2021attention} propose to train two actor-critic agents that share experiences: one that is provided with state information and another that is provided with image information. Furthermore, the authors introduce an attention mechanism in each agent that is aligned throughout training. The authors show that the attention-based asymmetrical training scheme improves the efficiency and the robustness of the learning process. In contrast to prior work, \citet{kamienny2020privileged} explores providing privileged information (PI) to both the critic and the policy networks using dropout. In particular, the authors evaluate the use of PI-Dropout~\citep{lambert2018deep} in the context of RL and show how it outperforms other methods to exploit privileged information, such as  distillation or auxiliary losses, in scenarios with partial observability. In the previous works the asymmetrical training scheme is posited experimentally, without theoretical guarantees on the convergence of the algorithms. Recent works have studied the theoretical properties of asymmetrical reinforcement learning: \citet{baisero2022asymmetric} introduced an asymmetrical version of policy iteration and of the Q-learning algorithm with convergence guarantees; \citet{sinha2023asymmetric} proposed an asymmetrical version of the actor-critic algorithm and derive performance bounds on their algorithm. 
Asymmetric training has also been explored in the context of multi-agent reinforcement learning (MARL), in part to deal with the decentralized nature of executing policies for multiple agents~\citep{oliehoek2008optimal, rashid2020monotonic,sunehag2017value,lyu2023centralized}. Approaches in cooperative MARL often exploit the paradigm of Centralized Training with Decentralized Execution (CTDE): during training, agents have access to a centralized critic that exploits the joint observation of all agents; at execution time each agent can only exploit its own observation to run their policy~\citep{oliehoek2008optimal, sunehag2017value, rashid2020monotonic}.

In this work, we also explore an asymmetrical actor-critic training scheme, where we provide global features to the critic, allowing the execution of the policy only with local features. We show how our approach enables learning super-human racing agents.

\section{Additional Details of the Evaluation Scenarios}
\label{sec:appendix_evaluation_scenarios}
In this section we present additional details regarding our evaluation scenarios. In Figure~\ref{fig:scenarios} we present the track layouts, the cars employed and examples of image observations across the three scenarios. We highlight that our agent achieves super-human performance across a wide variety of car dynamics, track layouts and conditions.

\section{Additional Ablation Studies}
\label{appendix_additional_ablation_study}

\subsection{Course Point Range}
\label{appendix_additional_ablation_study_course_point}

\begin{figure*}[t]
\centering
\begin{minipage}{0.48\textwidth}
  \centering
  \includegraphics[width=0.99\linewidth]{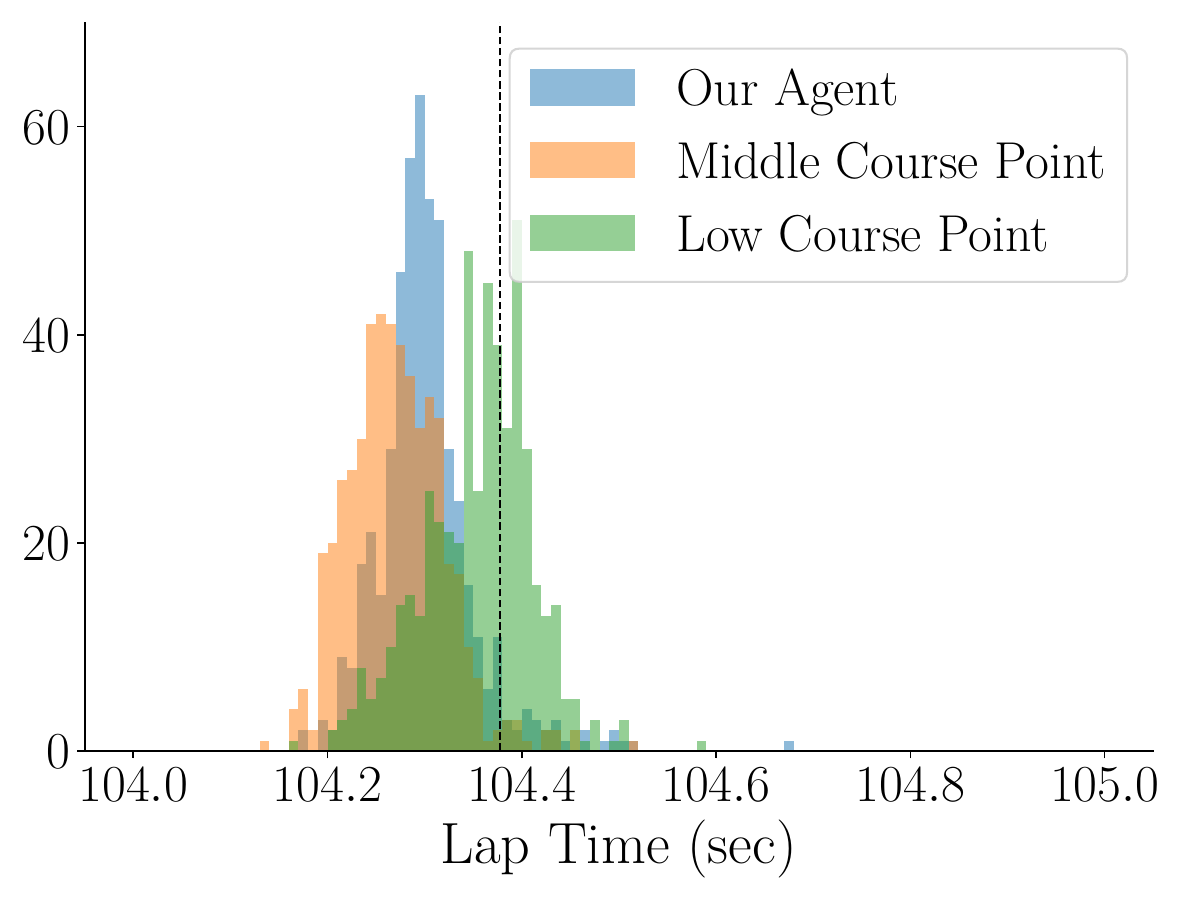}
\end{minipage}
\hfill
\begin{minipage}{0.481\textwidth}
  \centering
  \includegraphics[width=0.99\linewidth]{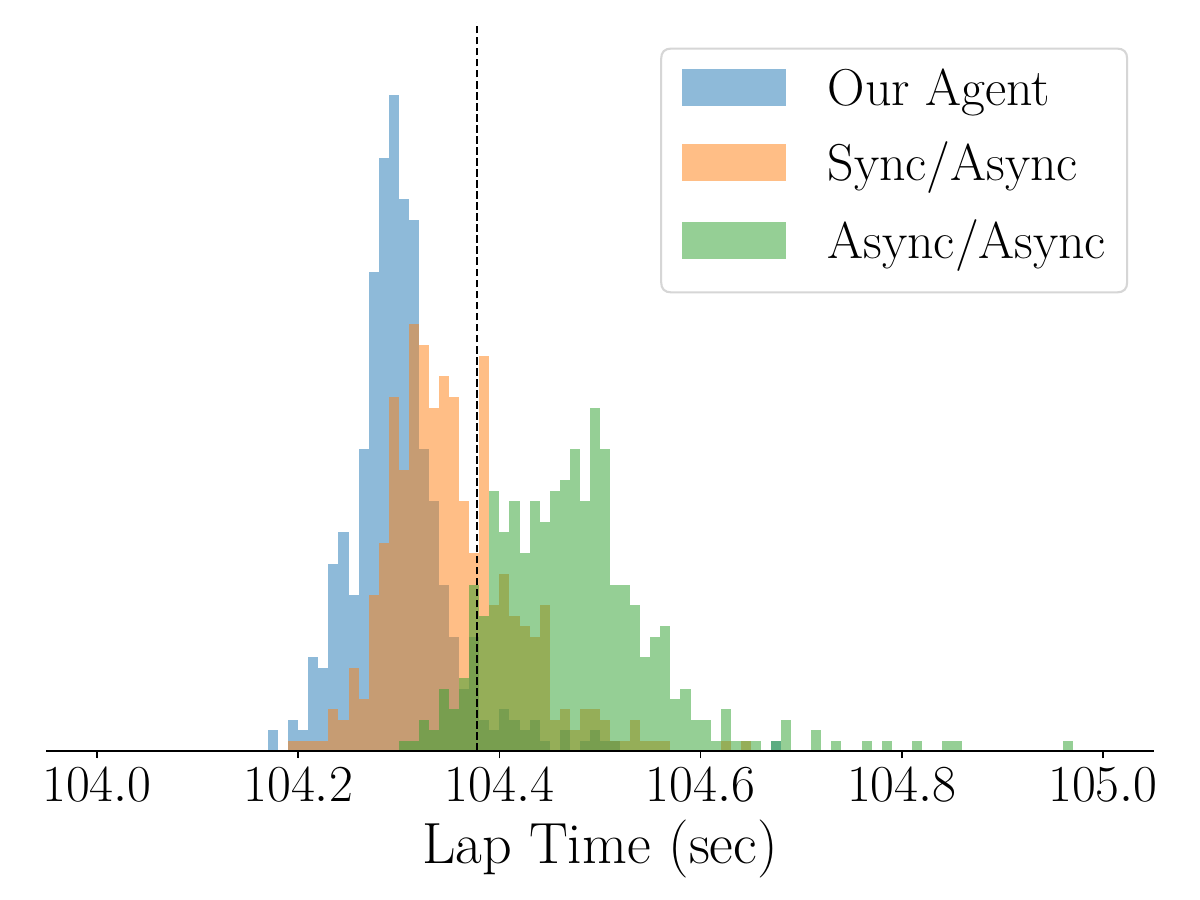}
\end{minipage}
\begin{minipage}{0.481\textwidth}
  \centering
  \includegraphics[width=0.99\linewidth]{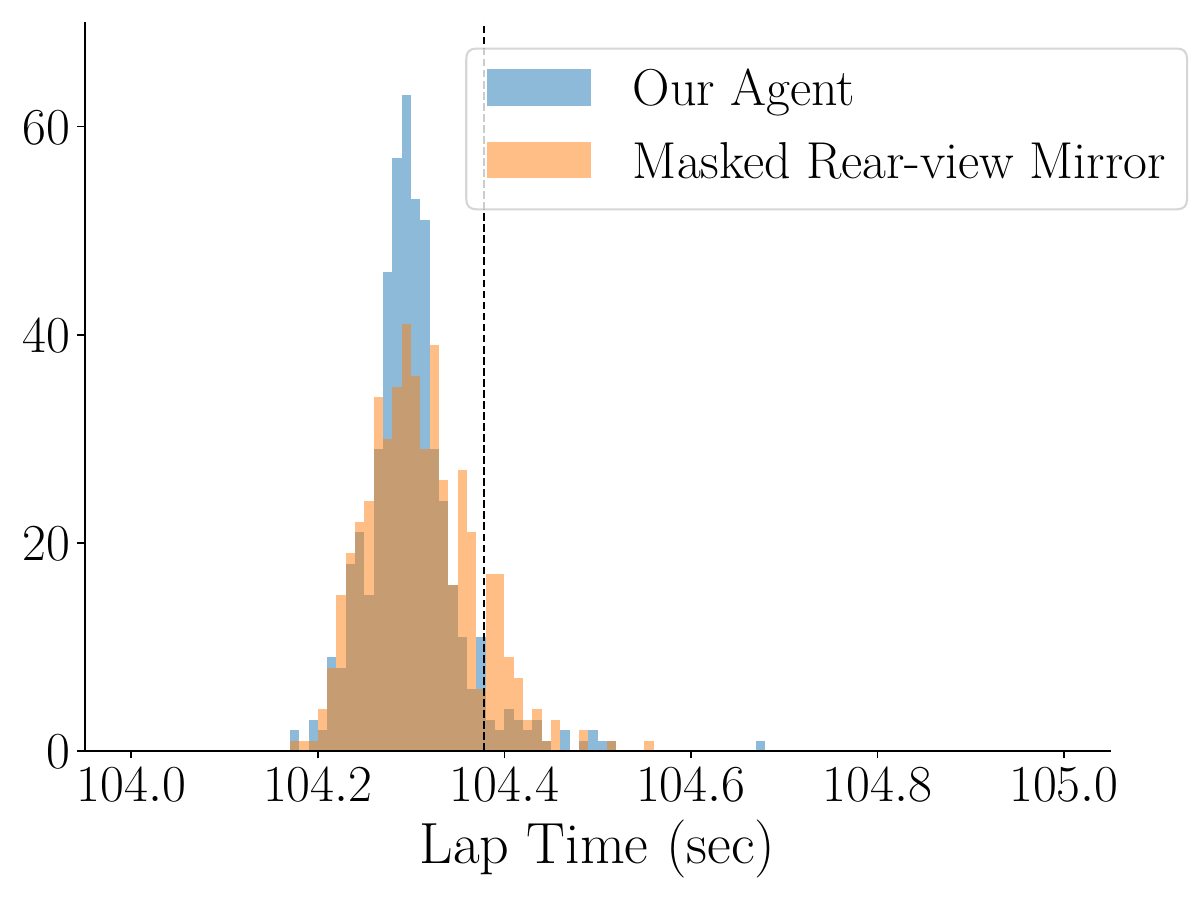}
\end{minipage}
\caption{Performance study of our racing agent in the \texttt{Monza} scenario in regards to (top left) different range of course points, (top right) different training/test synchronicity conditions and (bottom) different setups in the rear-view mirror. Additionally, we highlight the lap time of the fastest human player (black dashed line). We consider five randomly-seeded training runs and show the distribution of 500 evaluation laps, with 100 laps executed by the fastest policy in each training run. Lower is better.}
\label{fig:appendix_ablation_course_point}
\end{figure*}

We evaluate how the range of the course point feature affects the performance of our agent.
We reduce the range to 2 seconds and 4 seconds, denoted as \emph{low} and \emph{medium course points} respectively.
Figure~\ref{fig:appendix_ablation_course_point} shows the result of comparing our agent to the variations with different range of course points.
We observe that the shorter range of course points makes performance unreliable.
On the other hand, using medium course point range slightly improves the performance of our agent.
In this paper, we used the same range of course points as the one described in \citet{gt_sophy}.
However, this result indicates that tuning this hyperparameter could provide an additional performance improvement to our agent. We leave the tuning of this parameter for future work.

\subsection{Synchronous Training and Execution}
\label{appendix_additional_ablation_study_sync}

As discussed in Appendix~\ref{sec:appendix_model}, to train our agents we employ a synchronous training and testing scheme, similar to other simulators such as OpenAI Gym~\citep{openaigym2016}, where the simulator only executes simulation steps after receiving the next action commands sent by a rollout worker. However, by default, GT7 executes its simulation asynchronously in real time. We evaluate the feasibility of learning and executing policies in an asynchronous setting. Figure~\ref{fig:appendix_ablation_course_point} shows the result of comparing our agent, which trains and executes only on an synchronous mode, to the variations which: (i) train in synchronous mode and executes in asynchronous mode (\emph{Sync/Async}) and (ii) train and executes in asynchronous mode (\emph{Async/Async}). The results highlight the importance of the synchronous training, as training our agent with asynchronous mode results in a significant decrease in performance: this variation is only able to outperform the best human drivers only in $7.45\%$ of the laps. However, the performance of the policy of the agents that train in synchronous mode and execute asynchronously does not decrease significantly and is still able to outperform the best human drivers in $69.3\%$ of the laps. For future work, we plan on exploring parameter-efficient neural networks and optimizing hardware setups to mitigate the effect of latency on the control of our agents.

\subsection{Training with Masked Rear-view Mirror}
\label{appendix_additional_ablation_study_masked_rear_view_mirror}
In Section~\ref{sec:results:policy} we show that our agent considers information in the rear-view mirror section of the image observation to race across the task. To understand whether the rear-view mirror is fundamental to achieve the super-human performance, we additionally train our agents without rear-view mirror information, by masking this section with black pixels. Figure~\ref{fig:appendix_ablation_course_point} shows that our agent can still consistently achieve the super-human lap time even without the rear-view mirror. Based on this experimental result, we conclude that the attention to the rear-view mirror is an artifact of our end-to-end training scheme on a  track with a static layout.

\section{Training Curves}
\label{appendix_training_curves}

\begin{figure*}[!ht]
\centering
    \begin{minipage}[b]{0.8\linewidth}
        \centering
        \includegraphics[width=\linewidth]{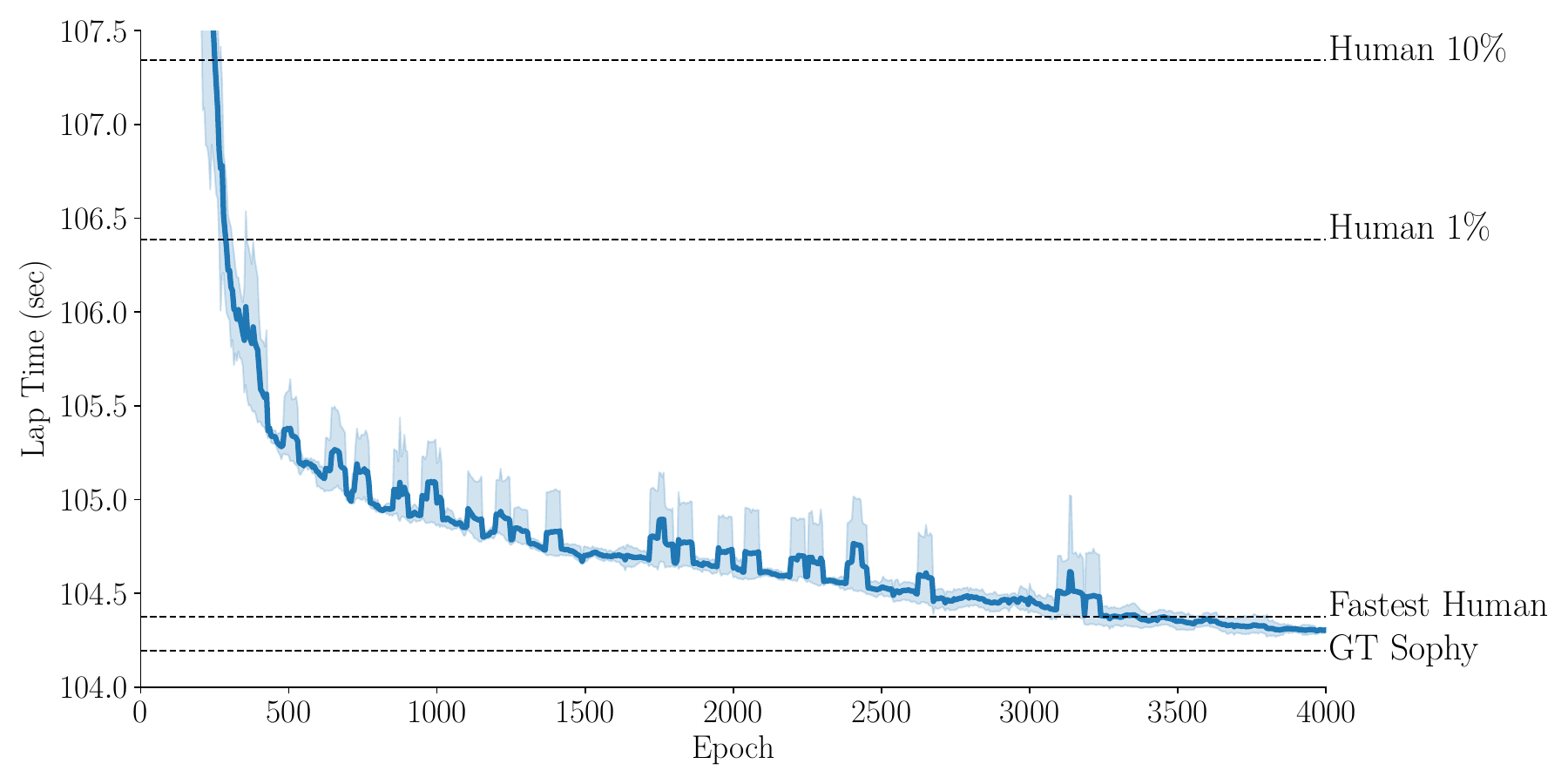}%
\subcaption{Monza}\label{fig:monza}%
    \end{minipage}%
    \hfil
    \begin{minipage}[b]{0.8\linewidth}
        \centering
        \includegraphics[width=\linewidth]{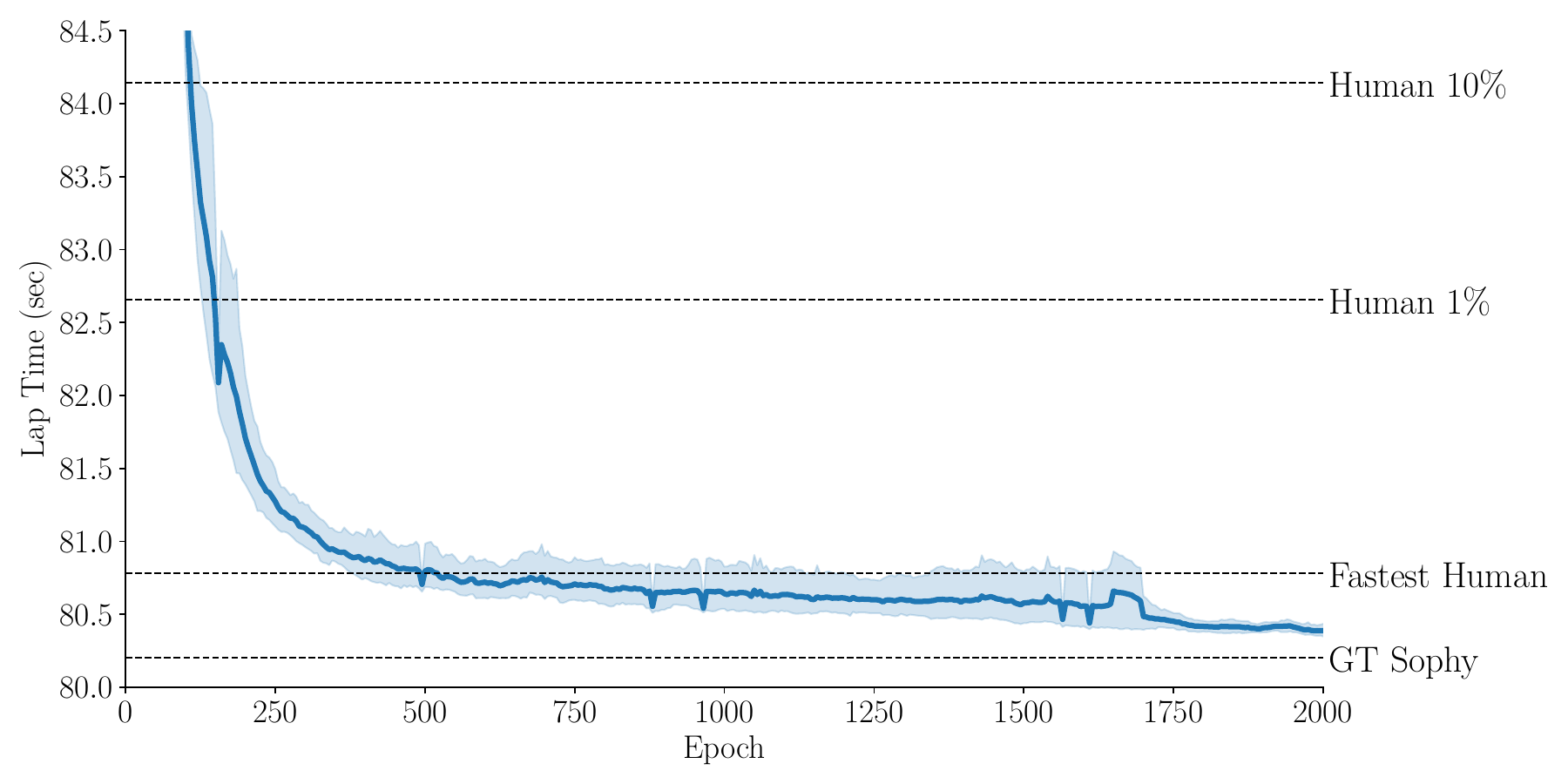}
        \subcaption{Tokyo}\label{fig:tokyo}%
    \end{minipage}%
    \hfil
    \begin{minipage}[b]{0.8\linewidth}
        \centering
        \includegraphics[width=\linewidth]{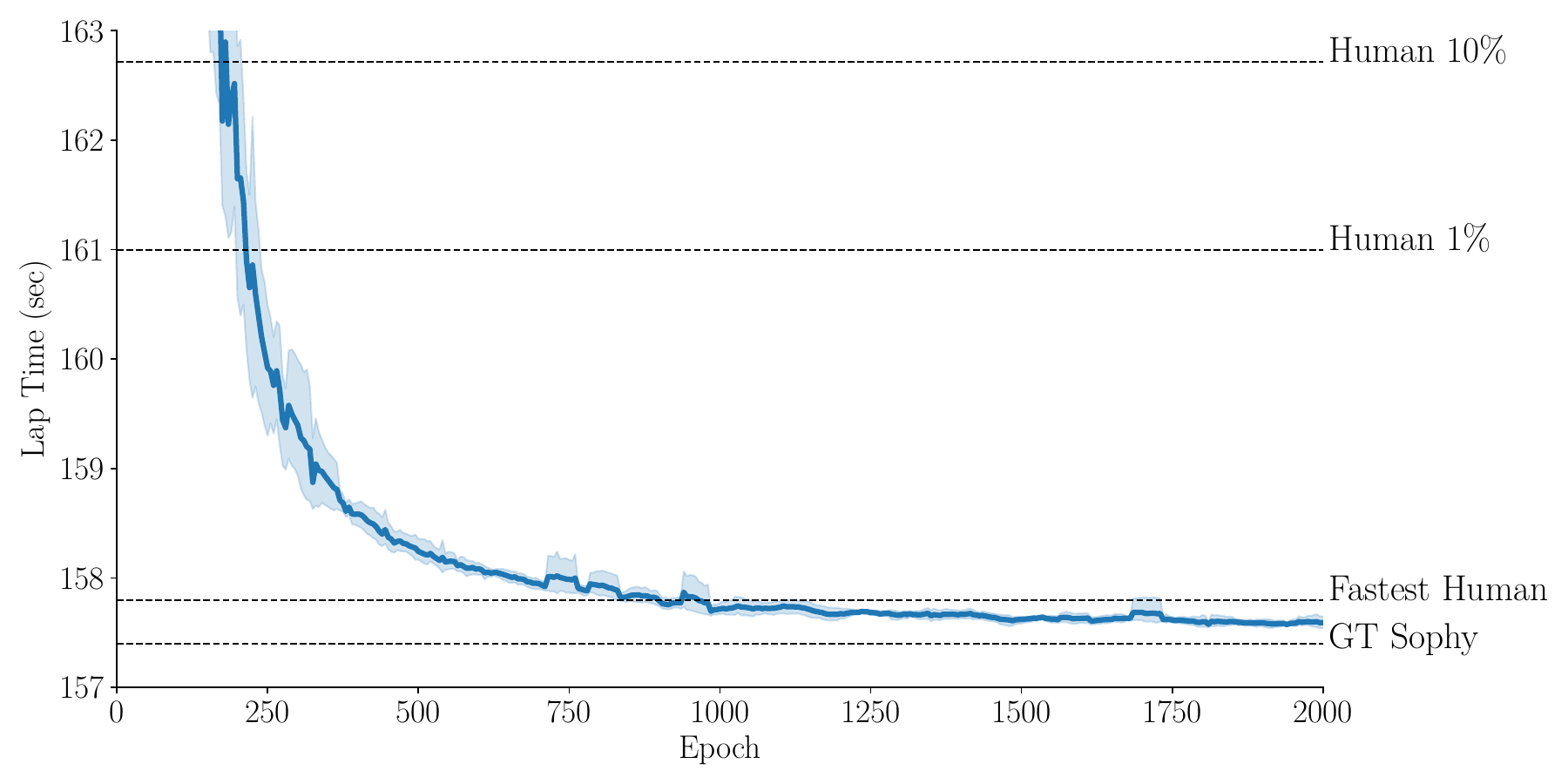}%
        \subcaption{Spa}\label{fig:spa}%
    \end{minipage}%
\caption{Training curves of our agent across all scenarios. We present the lap time per training epoch averaged over five randomly-seeded runs, with $95\%$ confidence interval. We compare our agent against human performance (10\%, 1\% and fastest) and GT Sophy. Training curves are smoothed for visual clarity.}
\label{fig:laptime}
\end{figure*}

\begin{figure*}[p]
\centering
    \begin{minipage}[b]{0.3\linewidth}
        \centering
        \includegraphics[width=\linewidth]{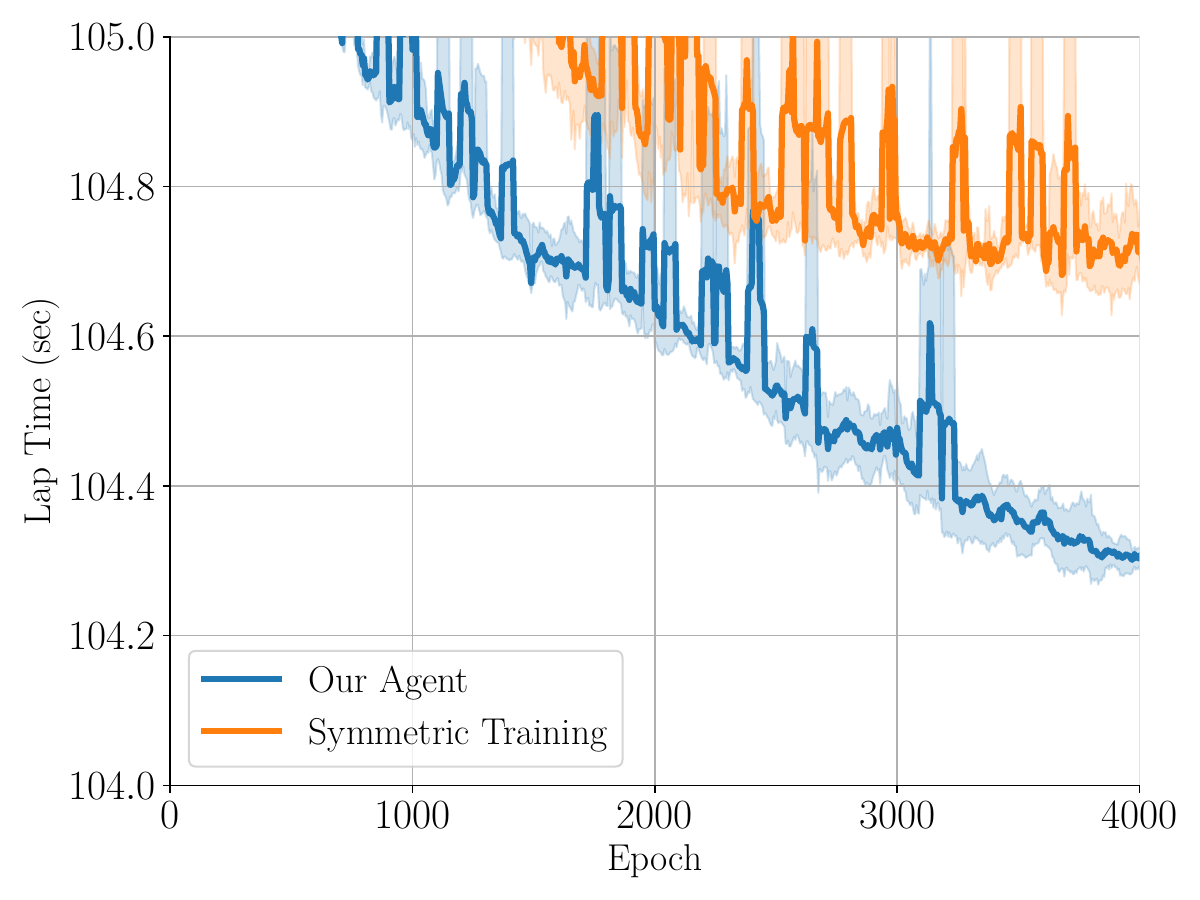}%
        \subcaption{Symmetric Training}\label{fig:symmetric}%
    \end{minipage}%
    \hfil
    \begin{minipage}[b]{0.3\linewidth}
        \centering
        \includegraphics[width=\linewidth]{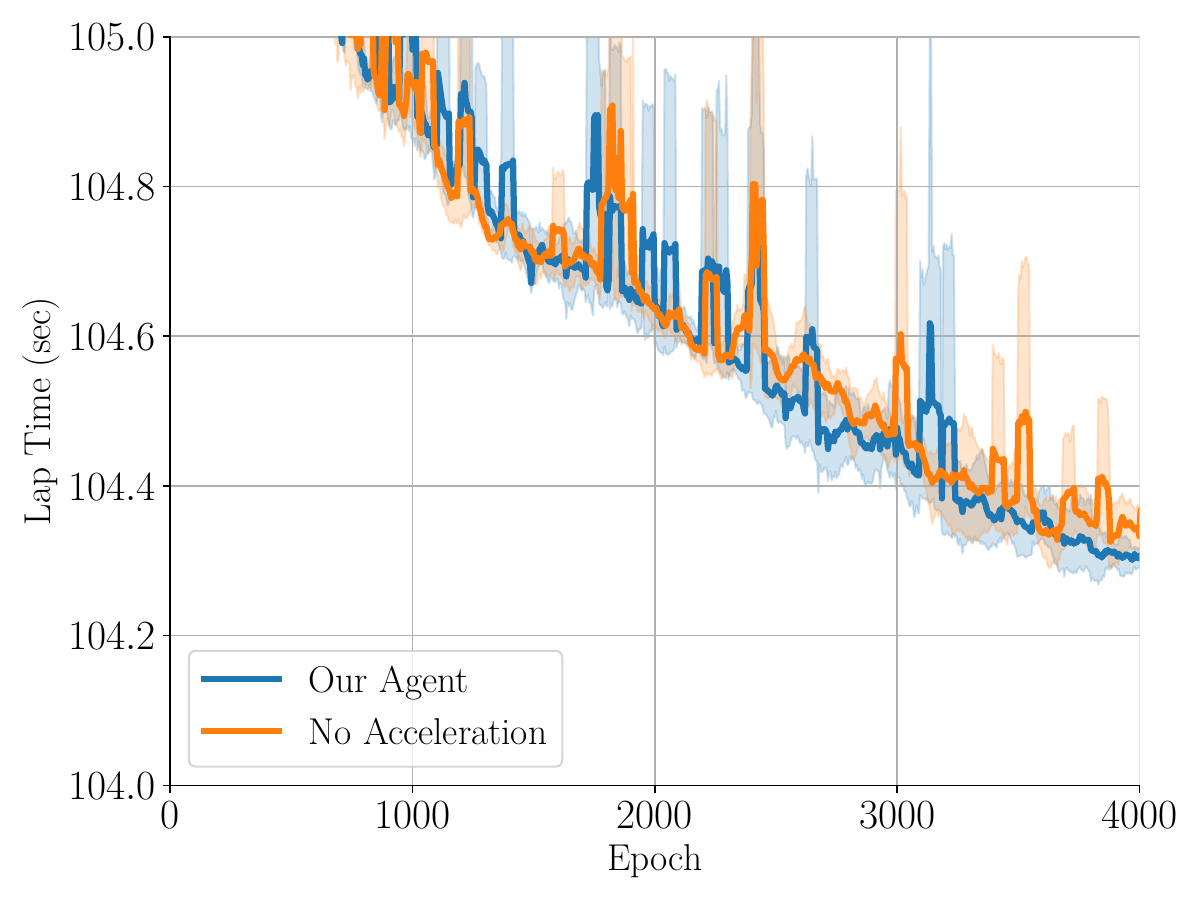}%
\subcaption{No Acceleration}\label{fig:no_accel}%
    \end{minipage}%
    \hfil
    \begin{minipage}[b]{0.3\linewidth}
        \centering
        \includegraphics[width=\linewidth]{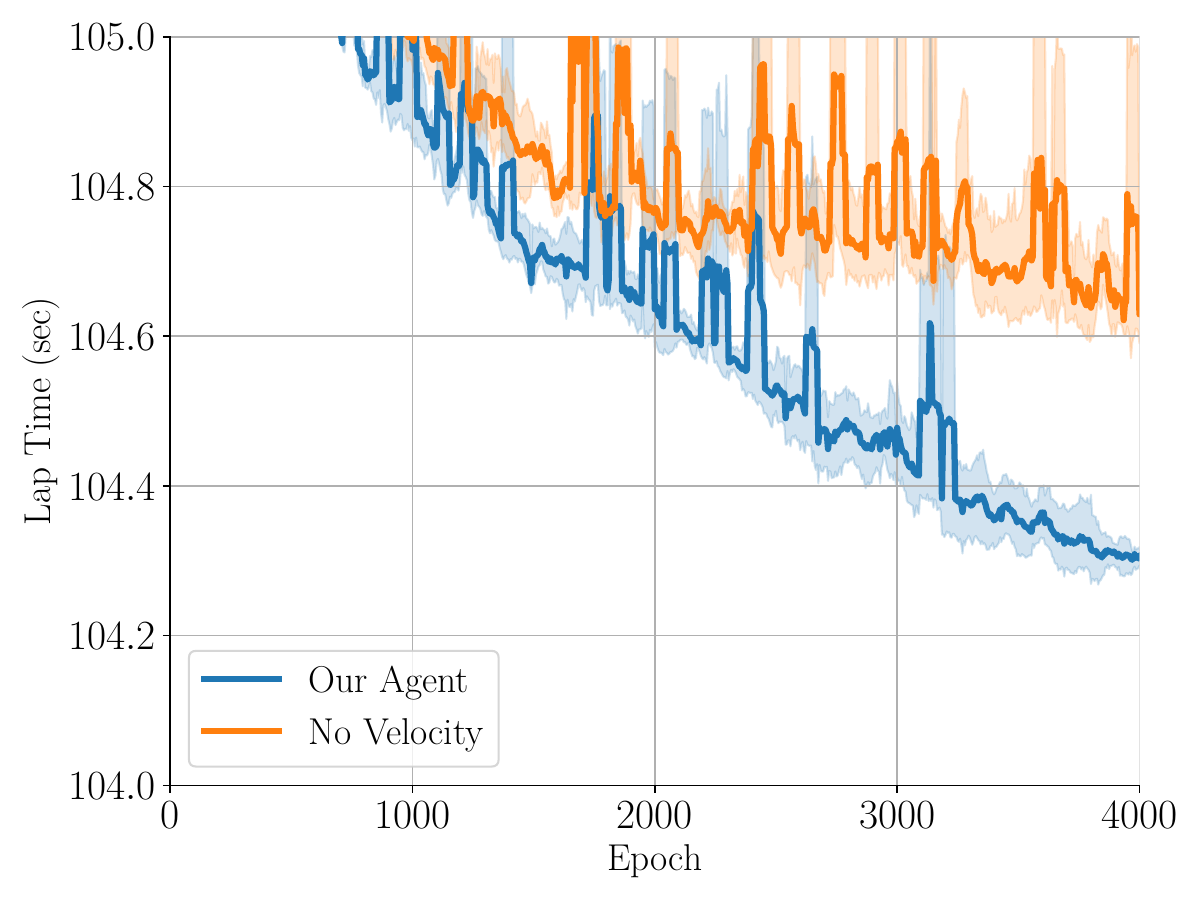}
        \subcaption{No Velocity}\label{fig:no_vel}%
    \end{minipage}%
    \hfil
    \begin{minipage}[b]{0.3\linewidth}
        \centering
        \includegraphics[width=\linewidth]{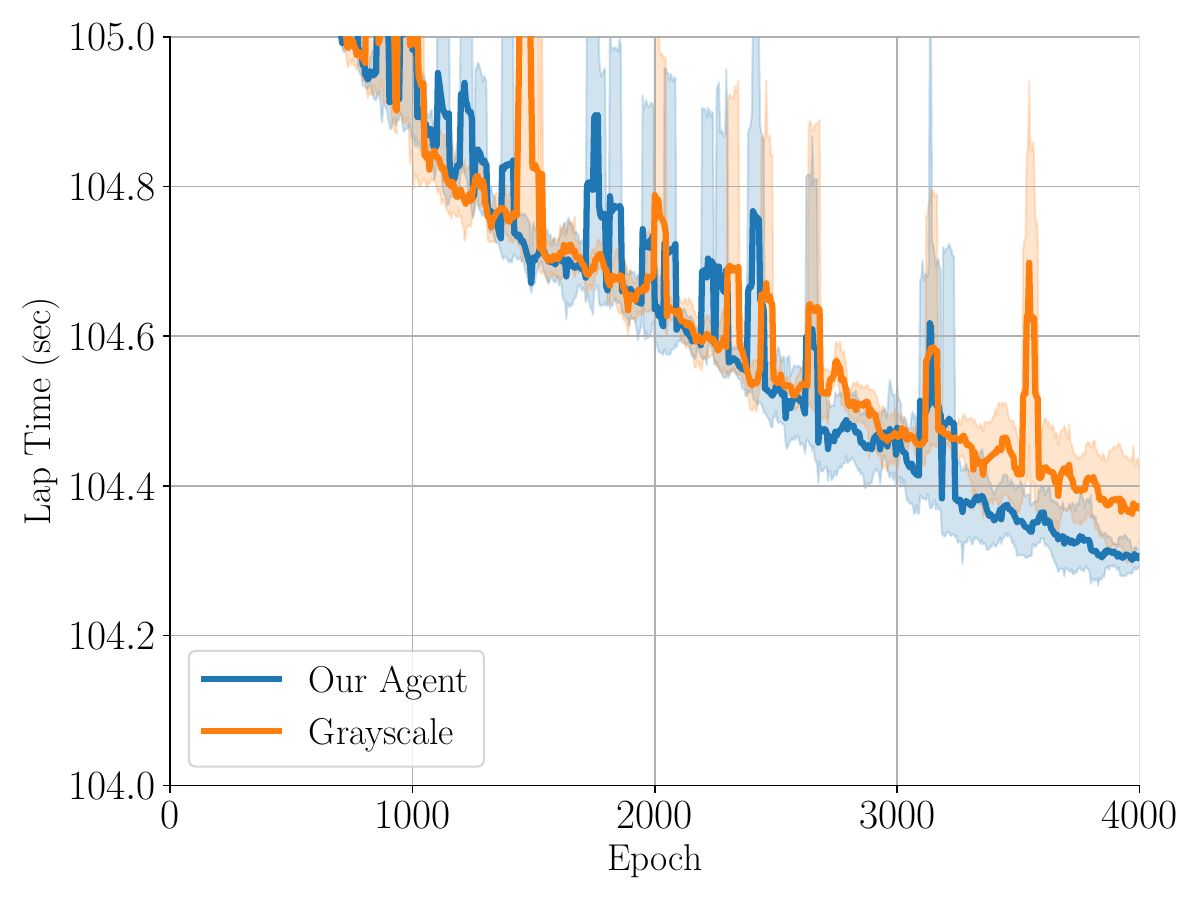}%
        \subcaption{Grayscale}\label{fig:gray}%
    \end{minipage}%
    \hfil
    \begin{minipage}[b]{0.3\linewidth}
        \centering
        \includegraphics[width=\linewidth]{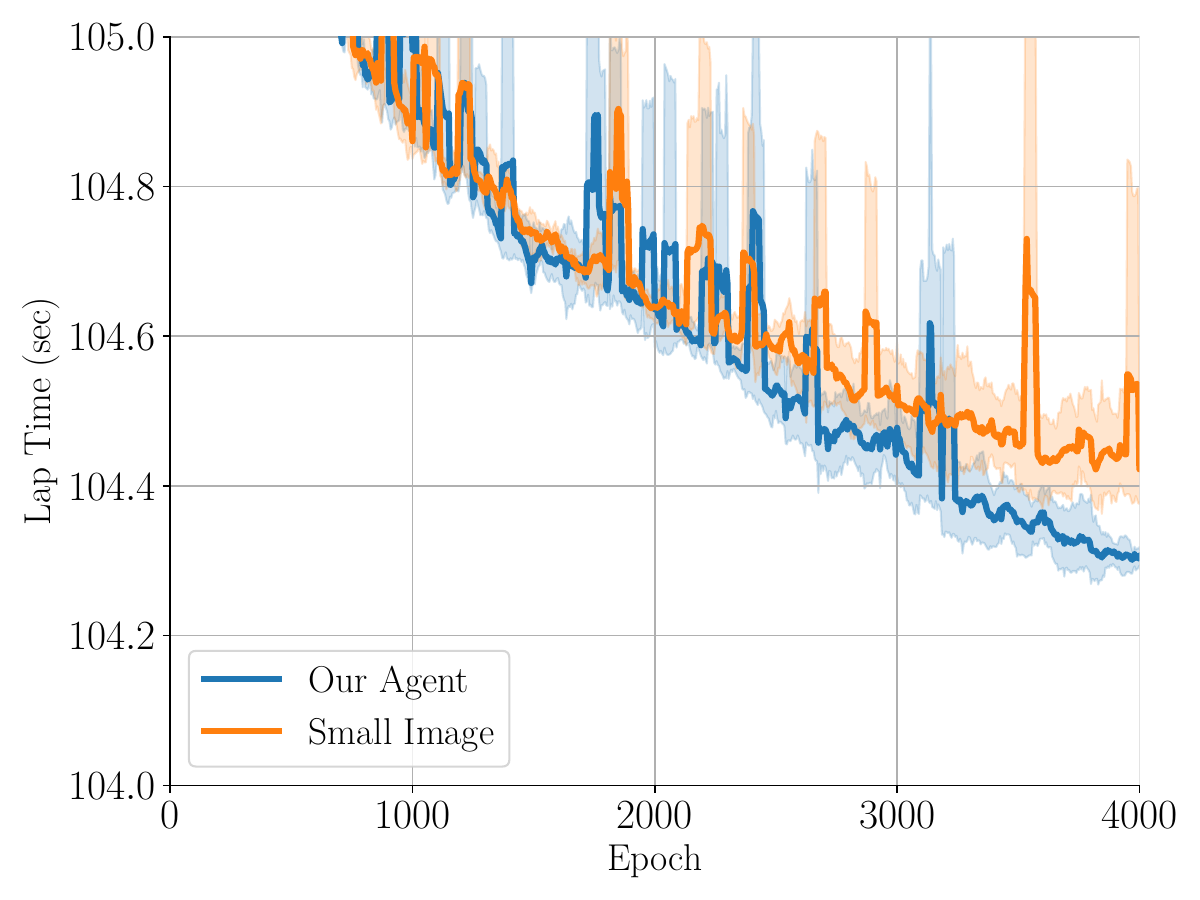}%
        \subcaption{Small Image}\label{fig:small_image}%
    \end{minipage}%
    \hfil
    \begin{minipage}[b]{0.3\linewidth}
        \centering
        \includegraphics[width=\linewidth]{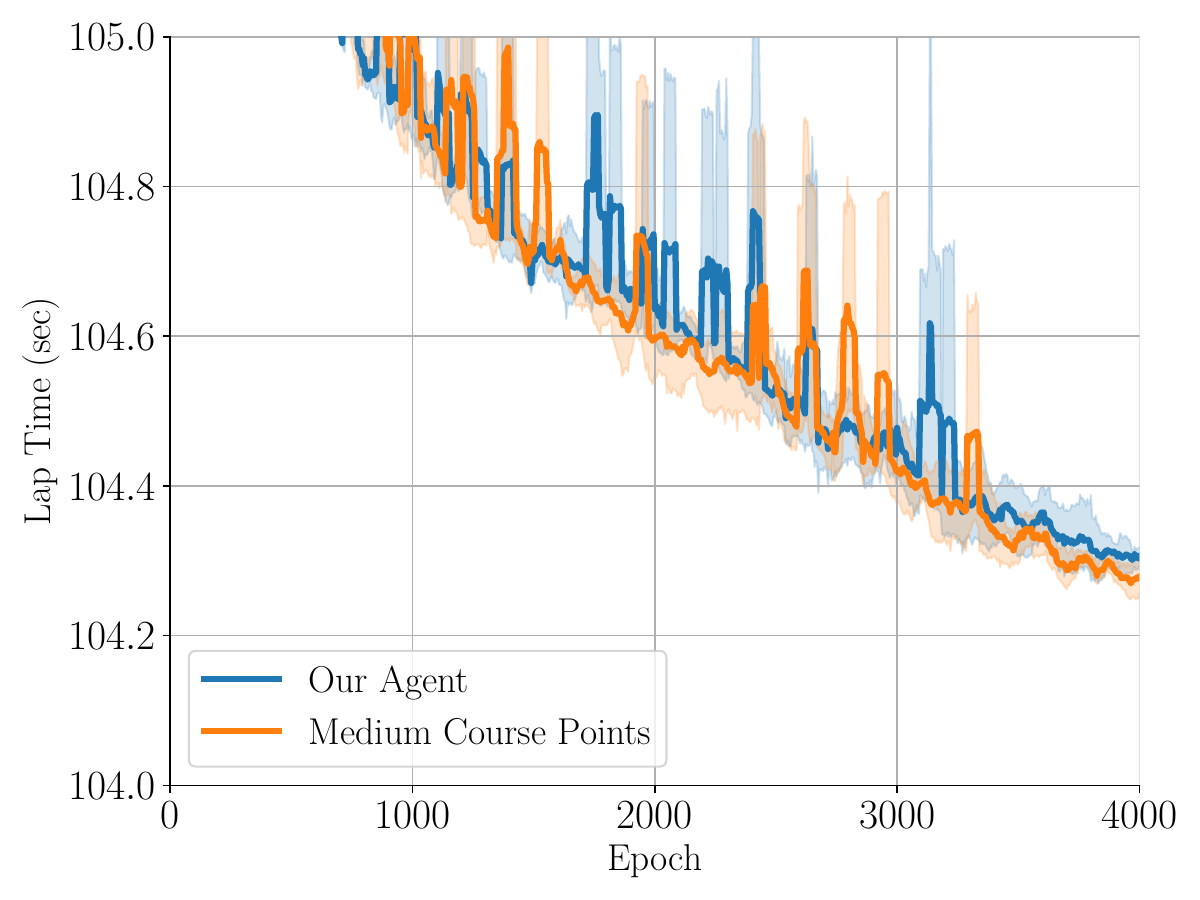}%
        \subcaption{Medium Course Points}\label{fig:medium_course_points}%
    \end{minipage}%
    \hfil
    \begin{minipage}[b]{0.3\linewidth}
        \centering
        \includegraphics[width=\linewidth]{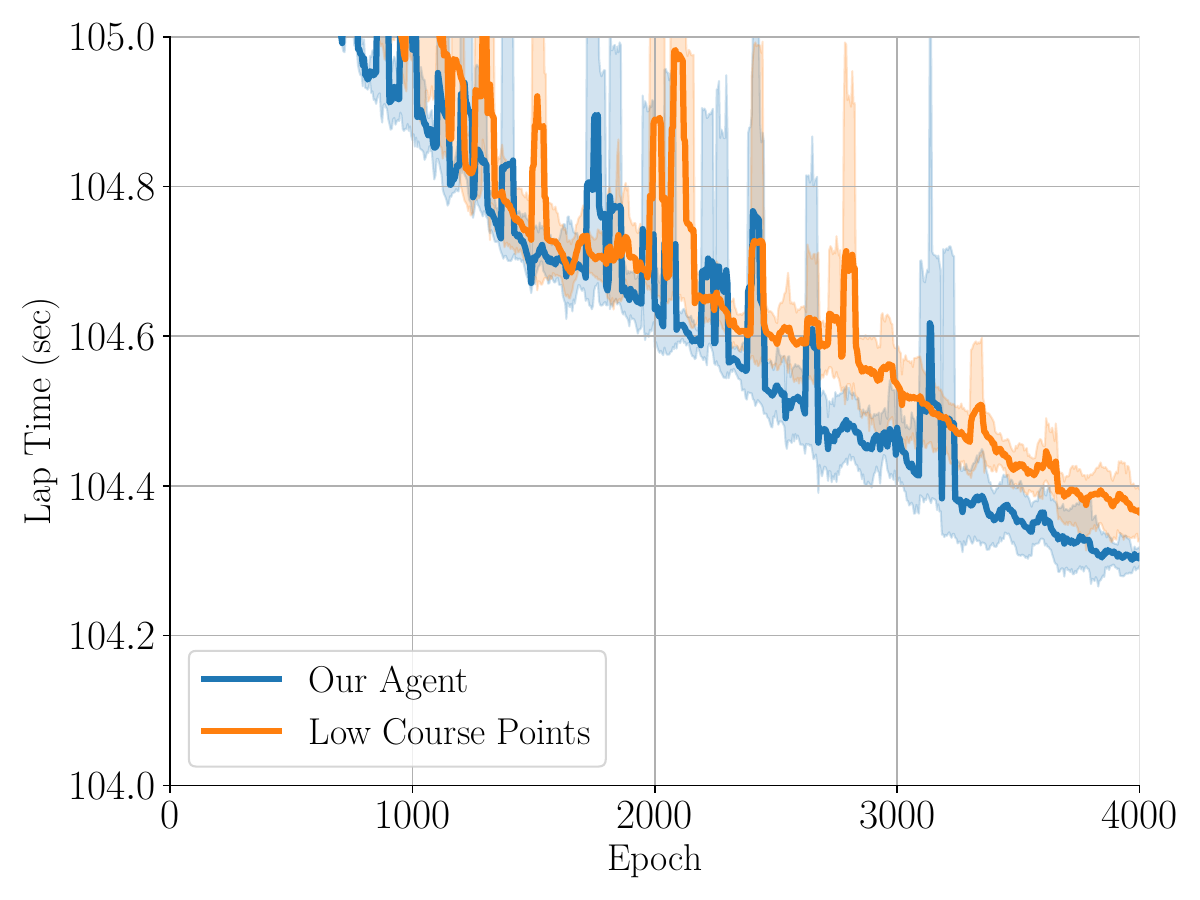}%
        \subcaption{Low Course Points}\label{fig:low_course_points}%
    \end{minipage}%
    \hfil
    \begin{minipage}[b]{0.3\linewidth}
        \centering
        \includegraphics[width=\linewidth]{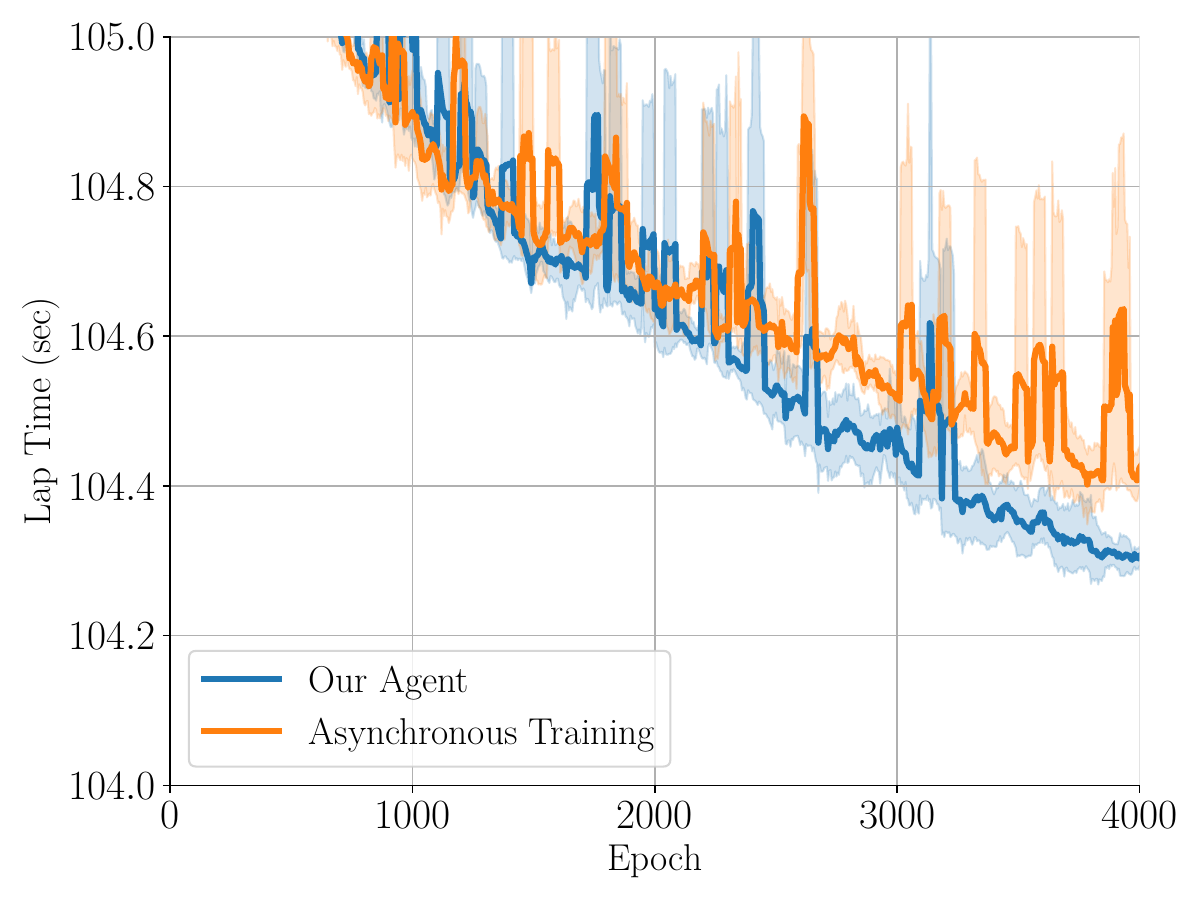}%
        \subcaption{Asynchronous Training}\label{fig:async_training}%
    \end{minipage}%
    \hfil
    \begin{minipage}[b]{0.3\linewidth}
        \centering
        \includegraphics[width=\linewidth]{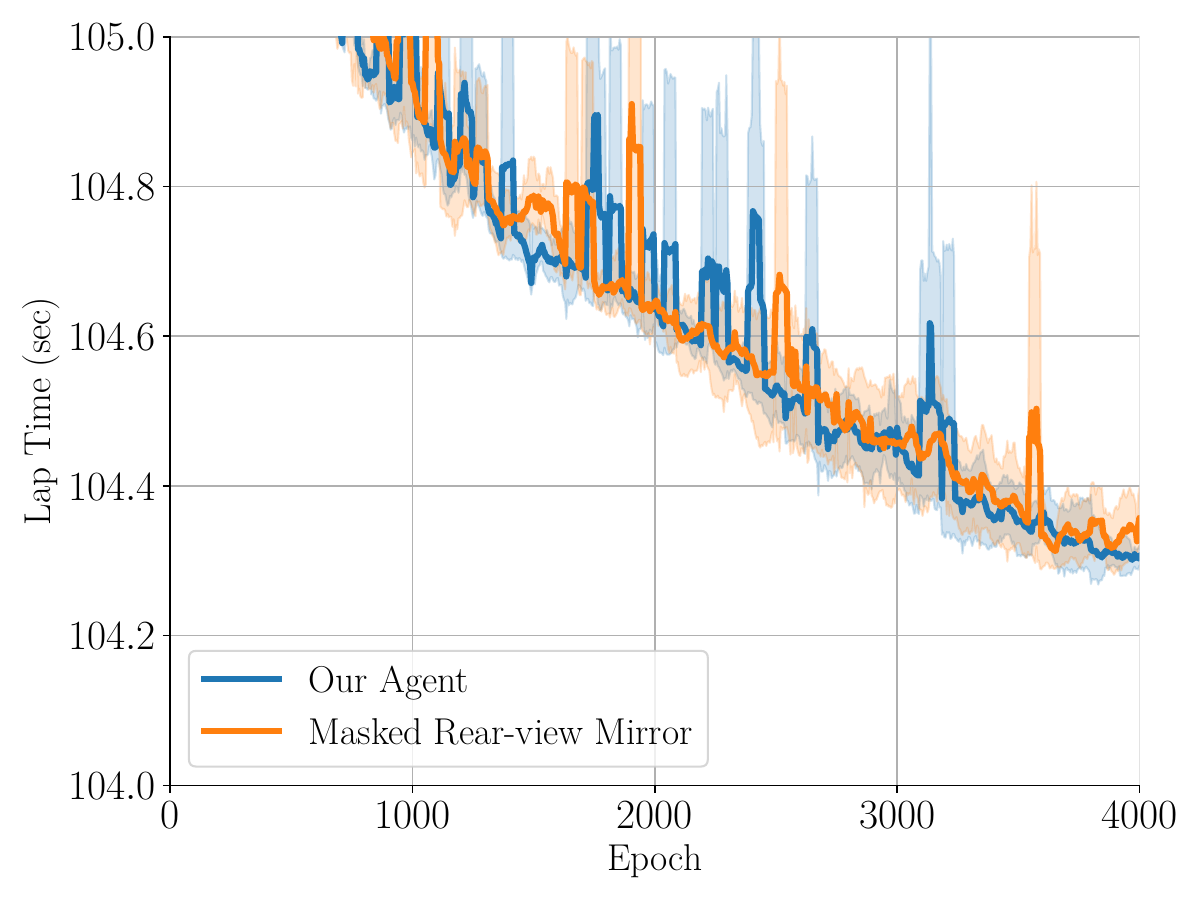}%
        \subcaption{Masked Rear-view Mirror}\label{fig:masked_rv_training}%
    \end{minipage}%
\caption{Training curves of our agent across all ablation conditions of Section~\ref{sec:results:ablation} and Appendix~\ref{appendix_additional_ablation_study}. We present the lap time per training epoch averaged over five randomly-seeded runs, with $95\%$ confidence interval. Training curves are smoothed for visual clarity.}
\label{fig:ablation_training_curves}
\end{figure*}

We present in Figure~\ref{fig:laptime} the overall training curves of our agent across all scenarios. The results show that our agent quickly exceeds top 1\% human performance and is able to consistently surpass the fastest human lap time. At the end of training, our agent also achieved comparable performance to GT Sophy.
We also present the overall training curves of our agent across all ablation conditions of Section~\ref{sec:results:ablation} and Appendix~\ref{appendix_additional_ablation_study} in Figure~\ref{fig:ablation_training_curves}.

\section{Additional Trajectory Analysis}
\label{appendix_trajectory_analysis}

\subsection{Comparison to Fastest Human Driver}
\begin{figure*}[p]
    \centering
        \includegraphics[width=0.9\textwidth]{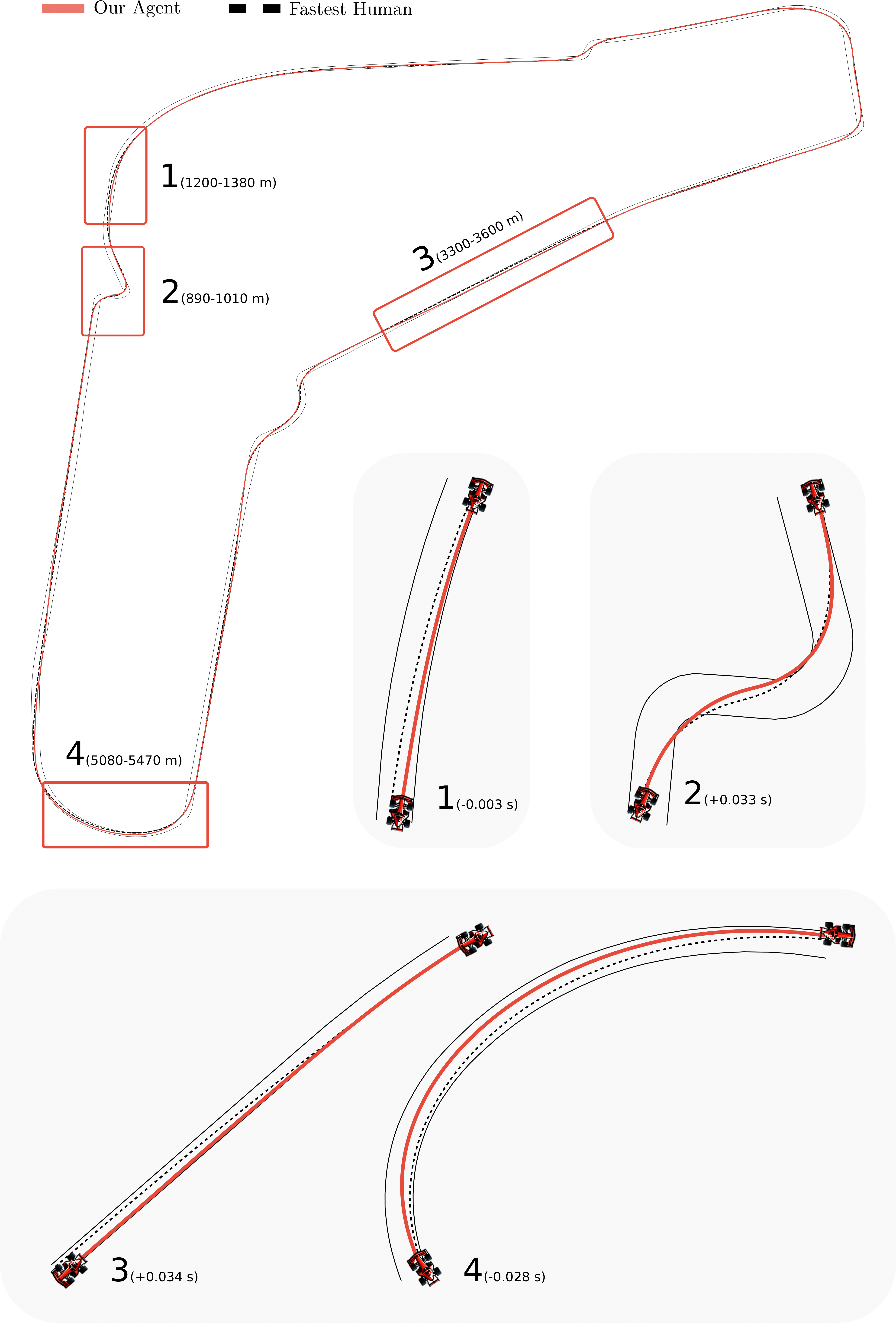}
    \caption{Trajectory comparison between our agent and the fastest human player in the \texttt{Monza} track. We highlight this comparison on (1, 3) straight sections approaching a curve, (2) a chicane section and (4) a curve section. We show in the figure the course progression of each segment as well as the time gained (negative values) or lost (positive values) to the best human driver. Best viewed zoomed in.}
    \label{fig:appendix_traj_monza}
\end{figure*}

\begin{figure*}[p]
    \centering
        \includegraphics[width=0.9\textwidth]{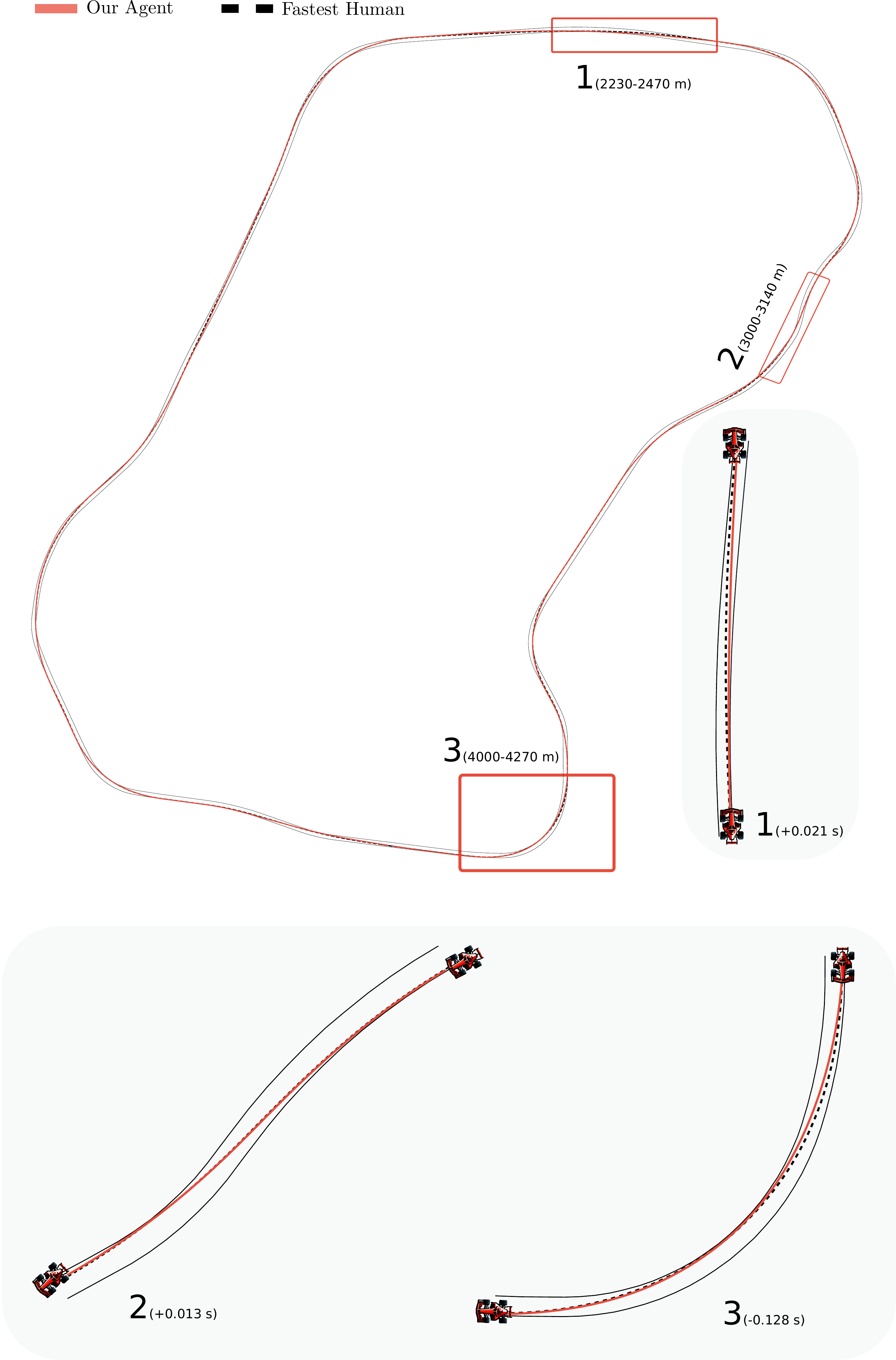}
    \caption{Trajectory comparison between our agent and the fastest human player in the \texttt{Tokyo} track. We highlight this comparison on (1, 2) straight sections and (3) a curve section. We show in the figure the course progression of each segment as well as the time gained (negative values) or lost (positive values) to the best human driver. Best viewed zoomed in.}
    \label{fig:appendix_traj_tokyo}
\end{figure*}

\begin{figure*}[p]
    \centering
        \includegraphics[width=0.9\textwidth]{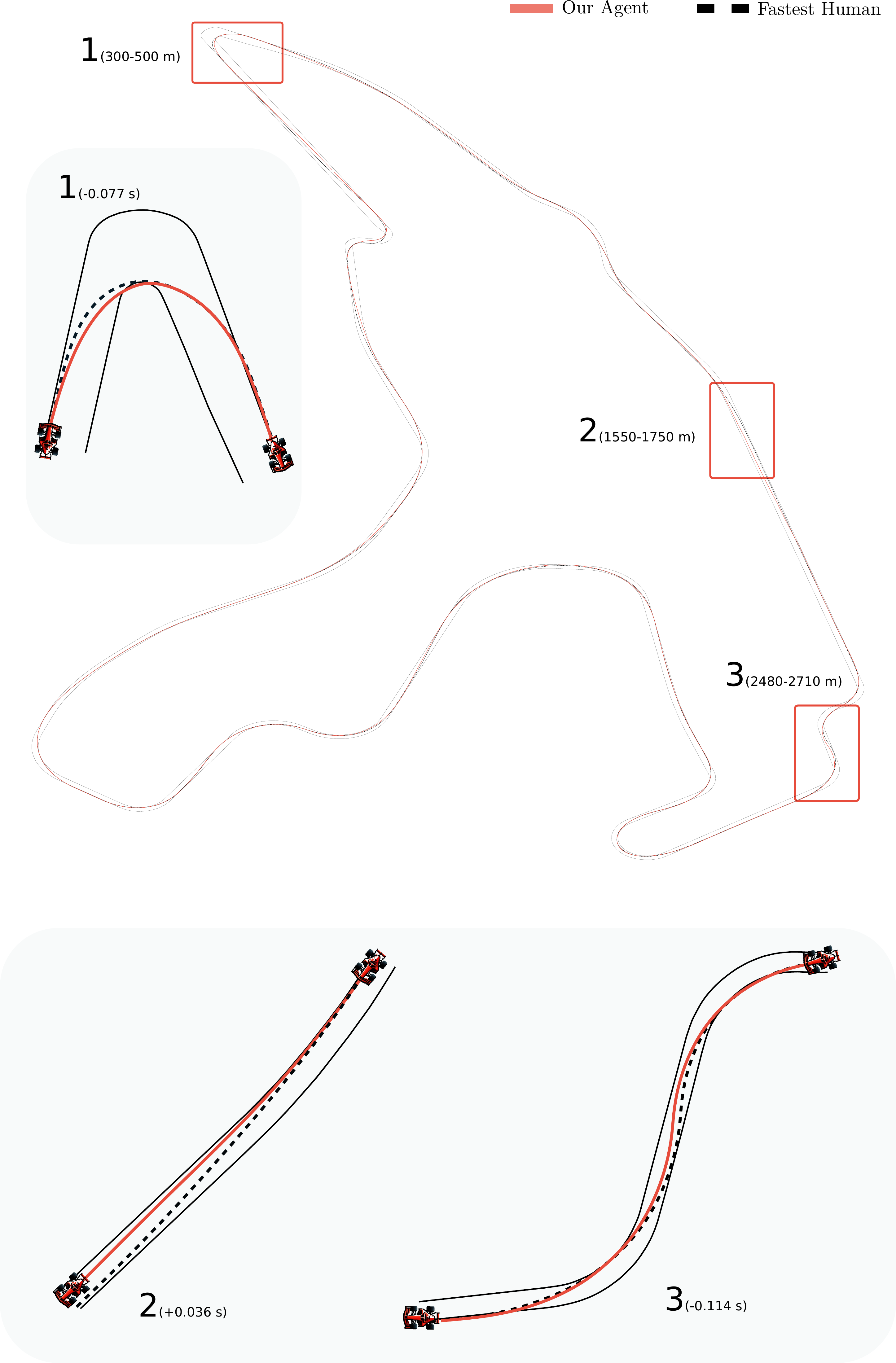}
    \caption{Trajectory comparison between our agent and the fastest human player in the \texttt{Spa} track. We highlight this comparison on (1) a U-turn section, (2) a straight section and (3) a chicane section. We show in the figure the course progression of each segment as well as the time gained (negative values) or lost (positive values) to the best human driver. Best viewed zoomed in.}
    \label{fig:appendix_traj_spa}
\end{figure*}

\begin{figure*}[p]
\centering
    \begin{minipage}[b]{0.9\linewidth}
        \centering
        \includegraphics[width=\linewidth]{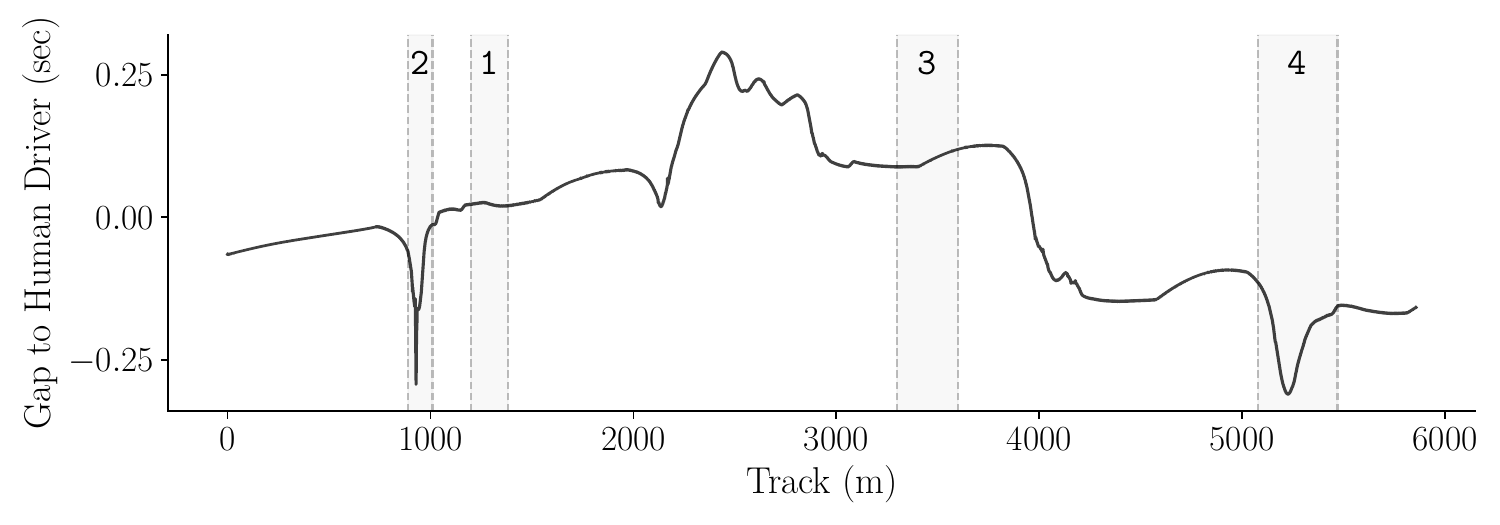}%
\subcaption{\texttt{Monza}}\label{fig:monza_gap}%
    \end{minipage}%
    \hfil
    \begin{minipage}[b]{0.9\linewidth}
        \centering
        \includegraphics[width=\linewidth]{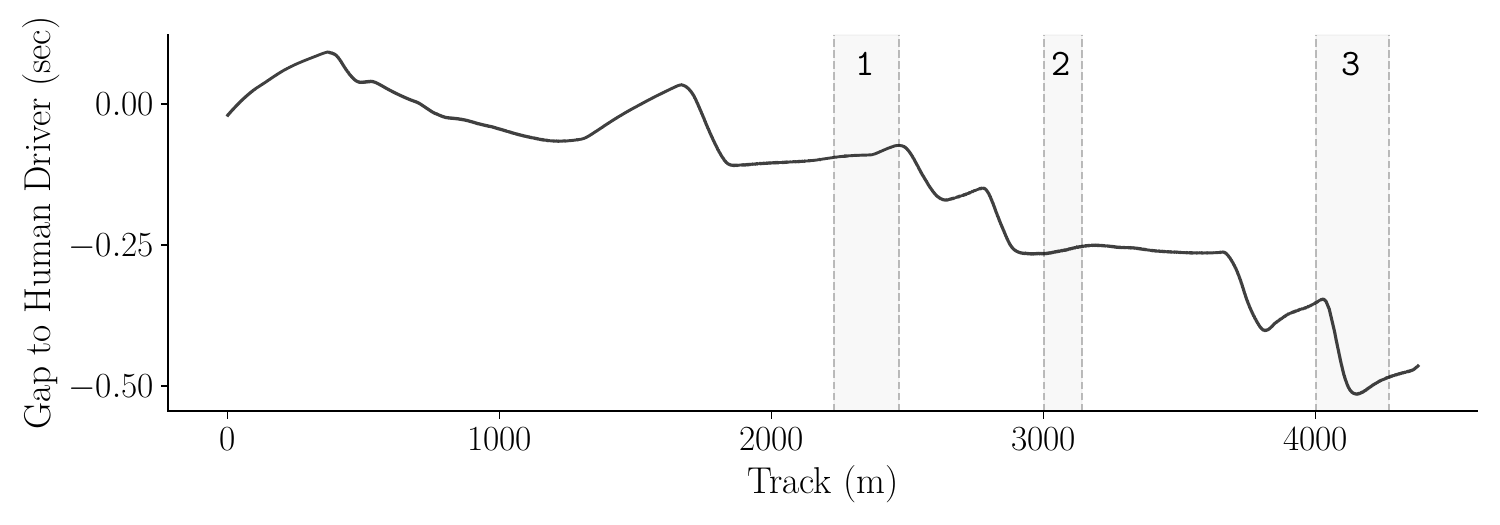}%
        \subcaption{\texttt{Tokyo}}\label{fig:tokyo_gap}%
    \end{minipage}%
    \hfil
    \begin{minipage}[b]{0.9\linewidth}
        \centering
        \includegraphics[width=\linewidth]{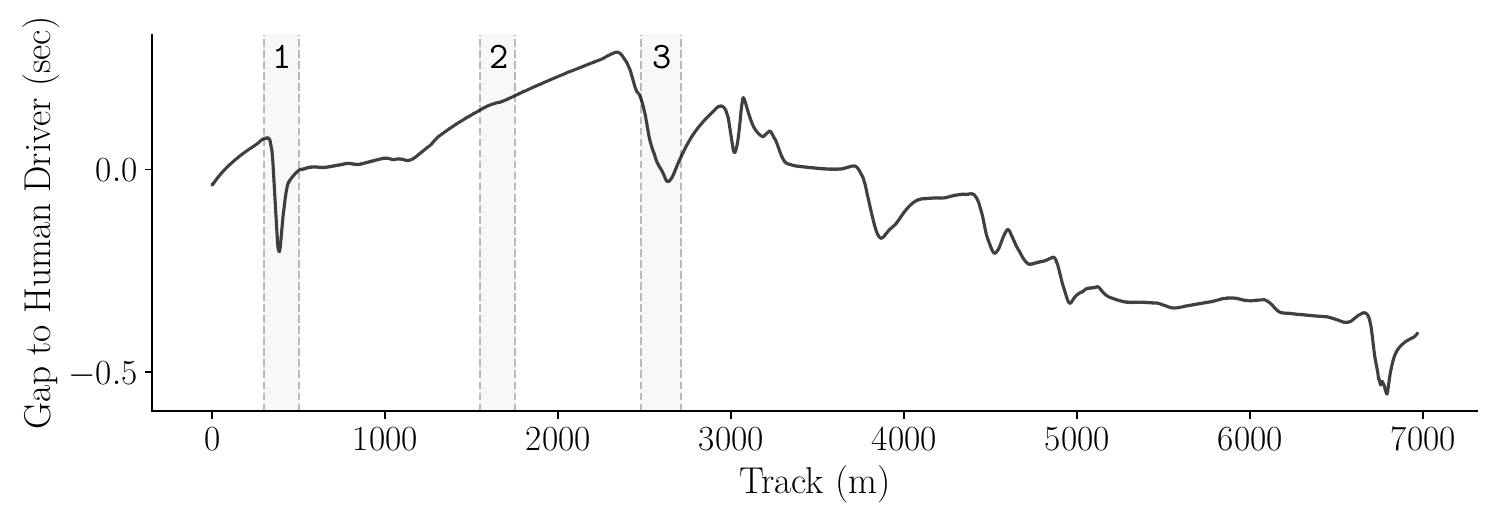}%
        \subcaption{\texttt{Spa}}\label{fig:spa_gap}%
    \end{minipage}%
\caption{Time difference between our agent and the best human reference driver as a function of the progression in the track. We identify the trajectory sections highlighted in the \texttt{Monza} (Figure~\ref{fig:appendix_traj_monza}), \texttt{Tokyo} (Figure~\ref{fig:appendix_traj_tokyo}) and \texttt{Spa} (Figure~\ref{fig:appendix_traj_spa}) tracks. Lower is better.}
\label{fig:ablation_gap}
\end{figure*}

\begin{figure*}[p]
    \centering
        \includegraphics[width=0.9\textwidth]{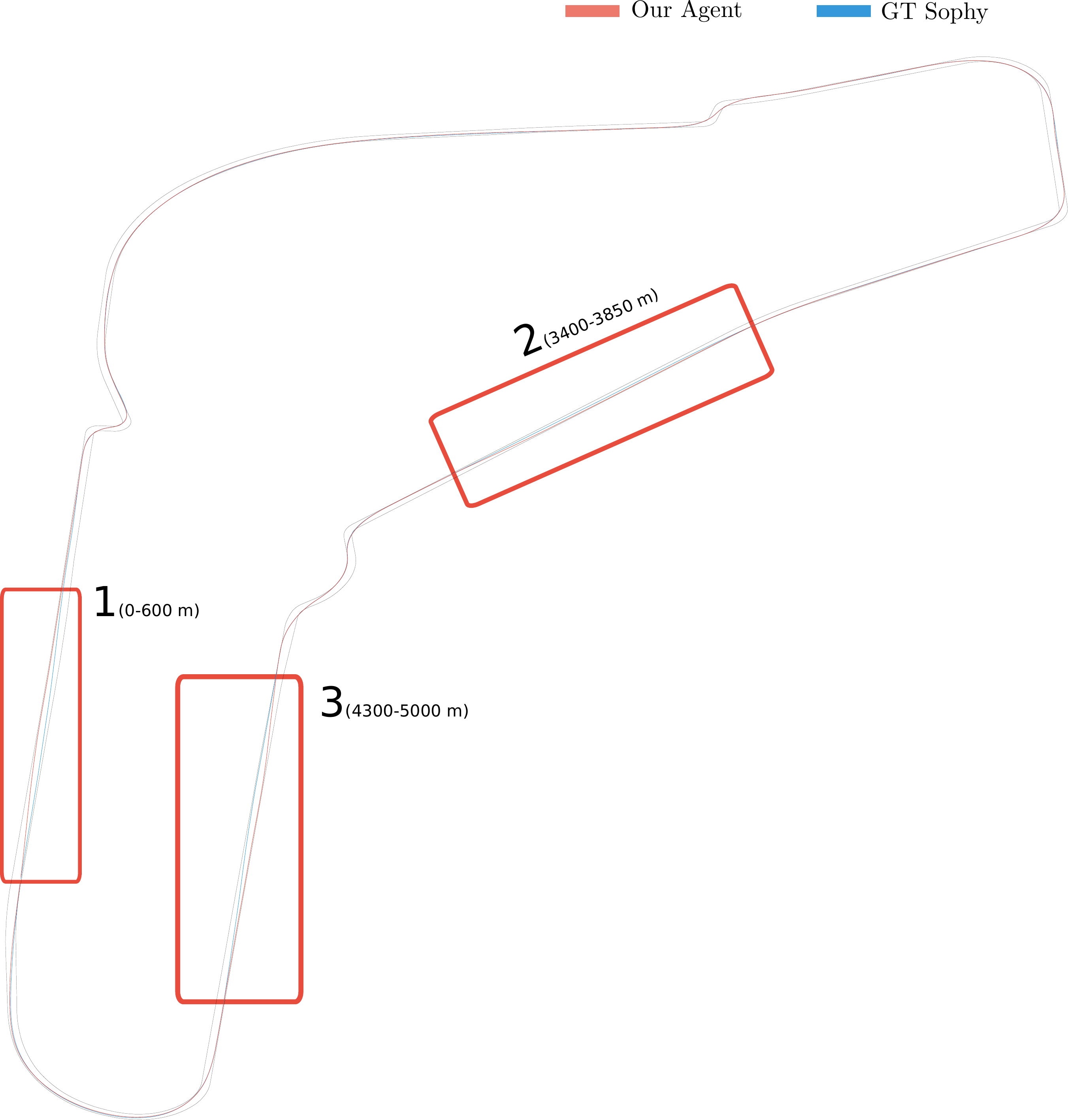}
    \caption{Trajectory comparison between our agent and GT Sophy~\citep{gt_sophy} in the \texttt{Monza} track. We highlight this comparison on sections that our trajectory is significantly different from Sophy's (1, 3) and sections where our agent loses time to Sophy (2). Best viewed in color and zoomed in.}
    \label{fig:appendix_traj_sophy_monza}
\end{figure*}

\begin{figure*}[p]
    \centering
        \includegraphics[width=0.9\textwidth]{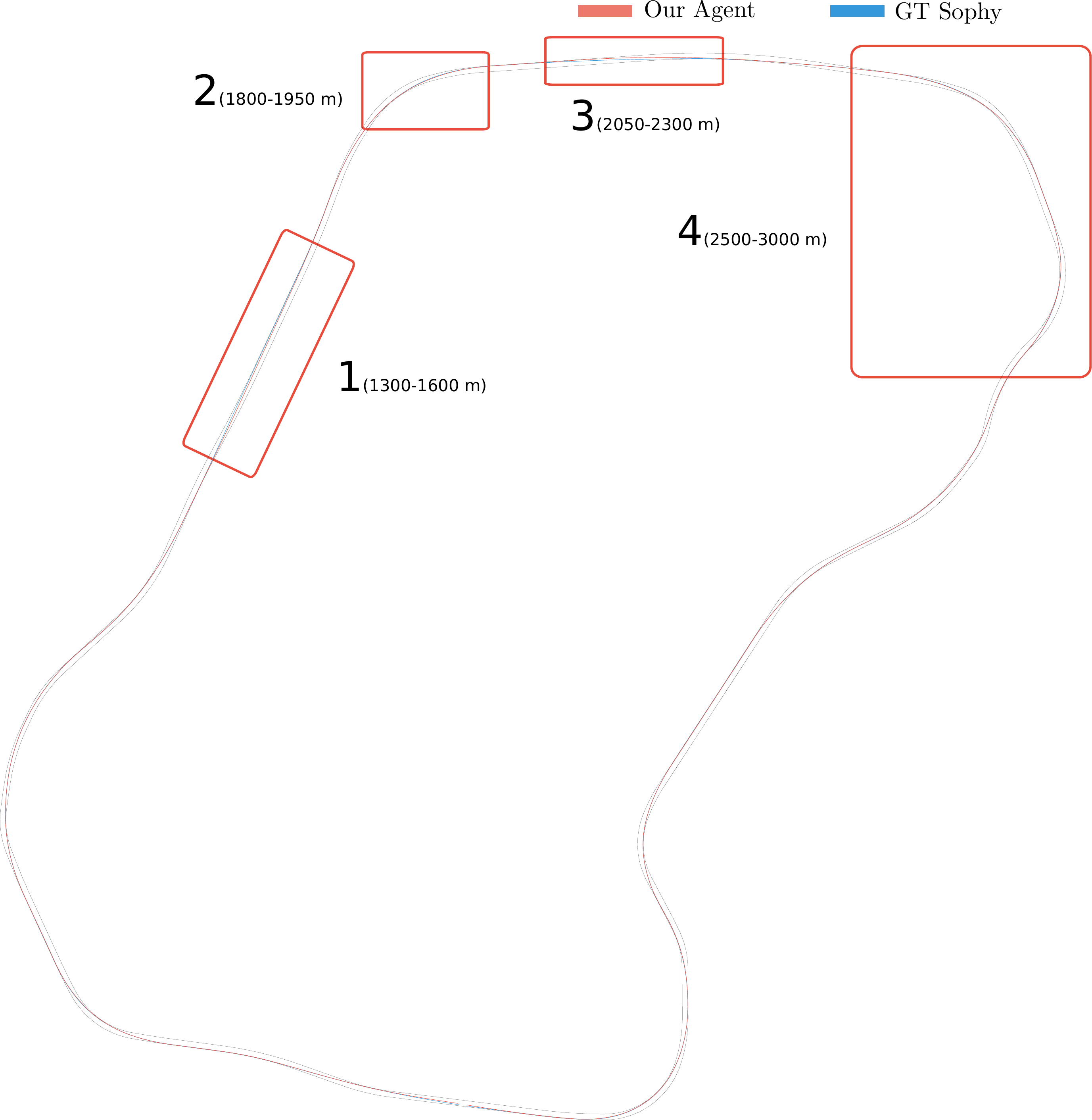}
    \caption{Trajectory comparison between our agent and GT Sophy~\citep{gt_sophy} in the \texttt{Tokyo} track. We highlight this comparison on sections where our trajectory is significantly different from GT Sophy's (1, 3) and sections where our agent loses time to Sophy (2, 4). Best viewed in color and zoomed in.}
    \label{fig:appendix_traj_sophy_tokyo}
\end{figure*}

\begin{figure*}[p]
    \centering
        \includegraphics[width=0.9\textwidth]{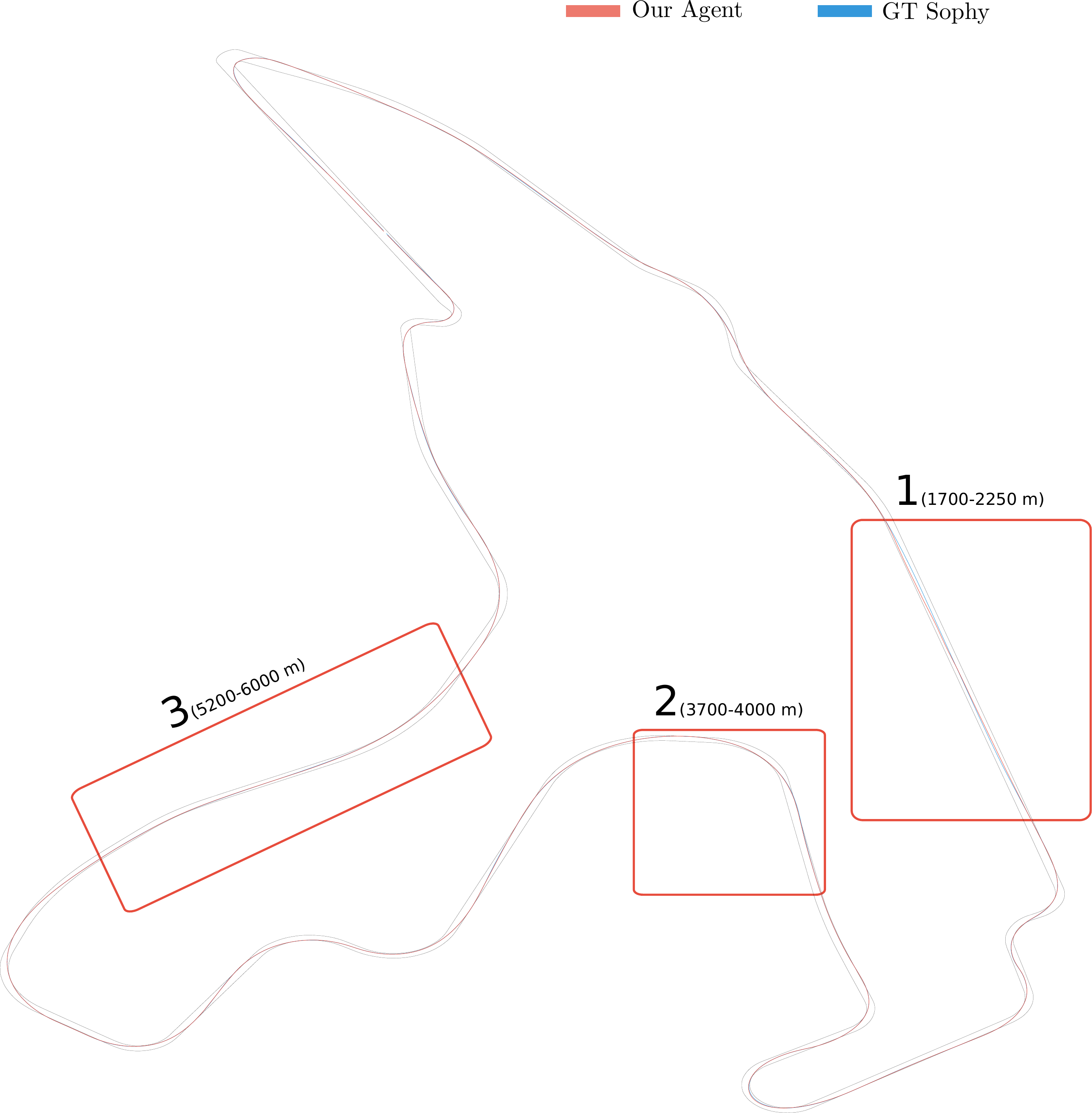}
    \caption{Trajectory comparison between our agent and GT Sophy~\citep{gt_sophy} in the \texttt{Spa} track. We highlight this comparison on sections where our trajectory is significantly different from GT Sophy's (1) and sections where our agent loses time to Sophy (2, 3). Best viewed in color and zoomed in.}
    \label{fig:appendix_traj_sophy_spa}
\end{figure*}

\begin{figure*}[p]
\centering
    \begin{minipage}[b]{0.9\linewidth}
        \centering
        \includegraphics[width=\linewidth]{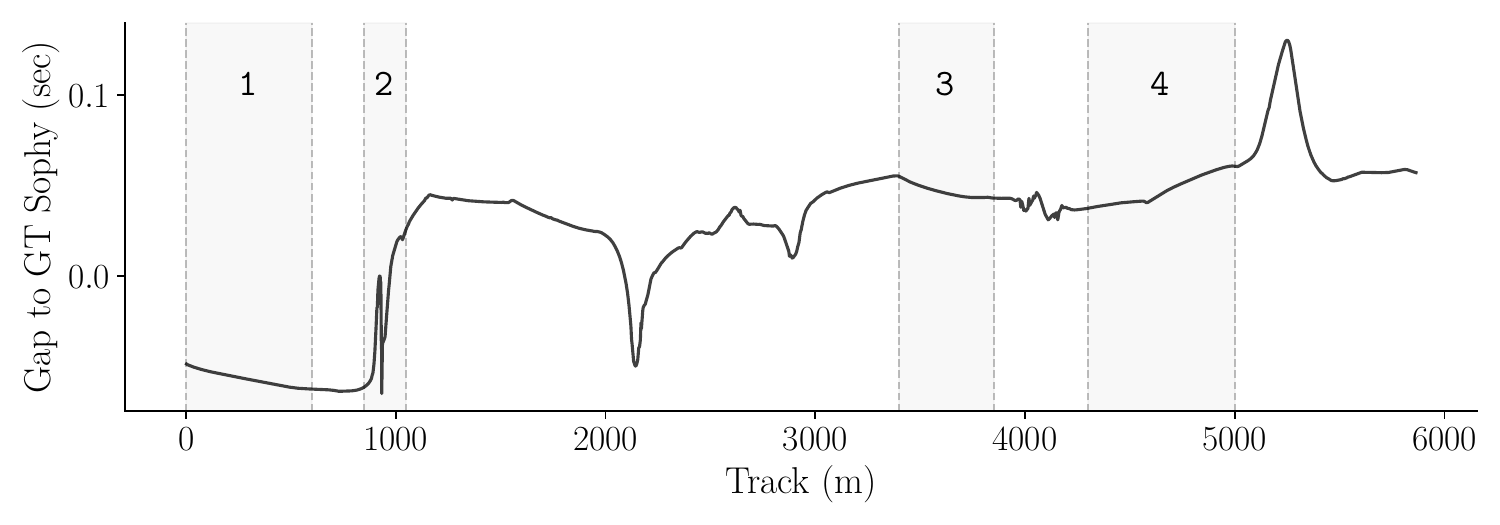}%
\subcaption{\texttt{Monza}}\label{fig:monza_sophy_gap}%
    \end{minipage}%
    \hfil
    \begin{minipage}[b]{0.9\linewidth}
        \centering
        \includegraphics[width=\linewidth]{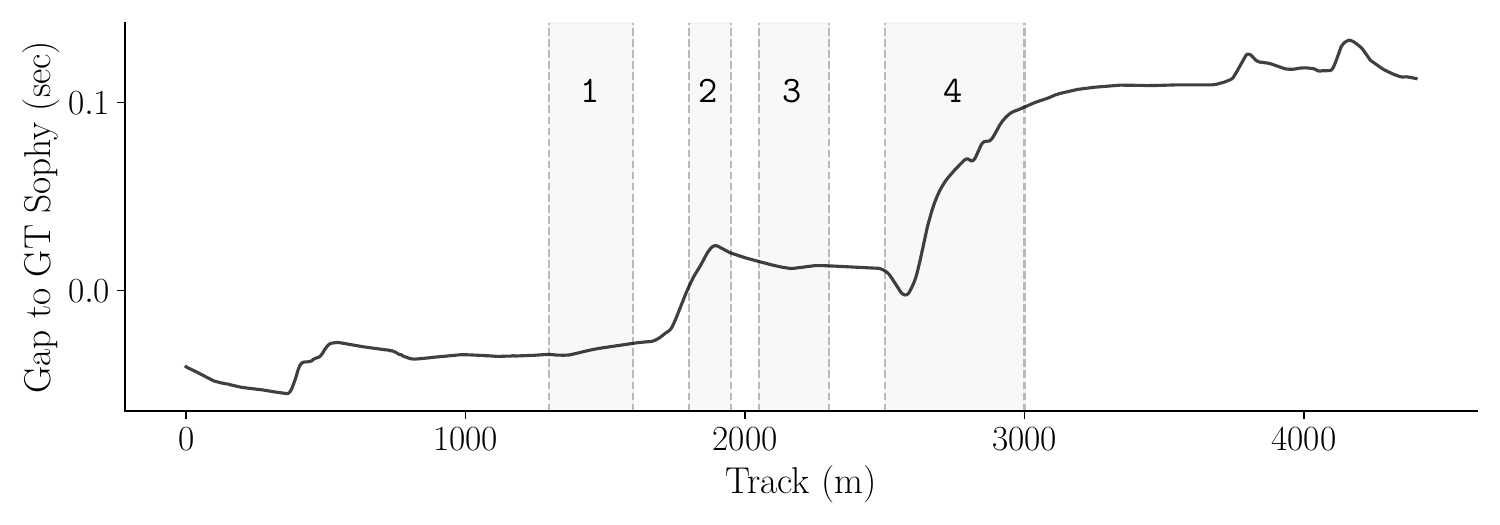}%
        \subcaption{\texttt{Tokyo}}\label{fig:tokyo_sophy_gap}%
    \end{minipage}%
    \hfil
    \begin{minipage}[b]{0.9\linewidth}
        \centering
        \includegraphics[width=\linewidth]{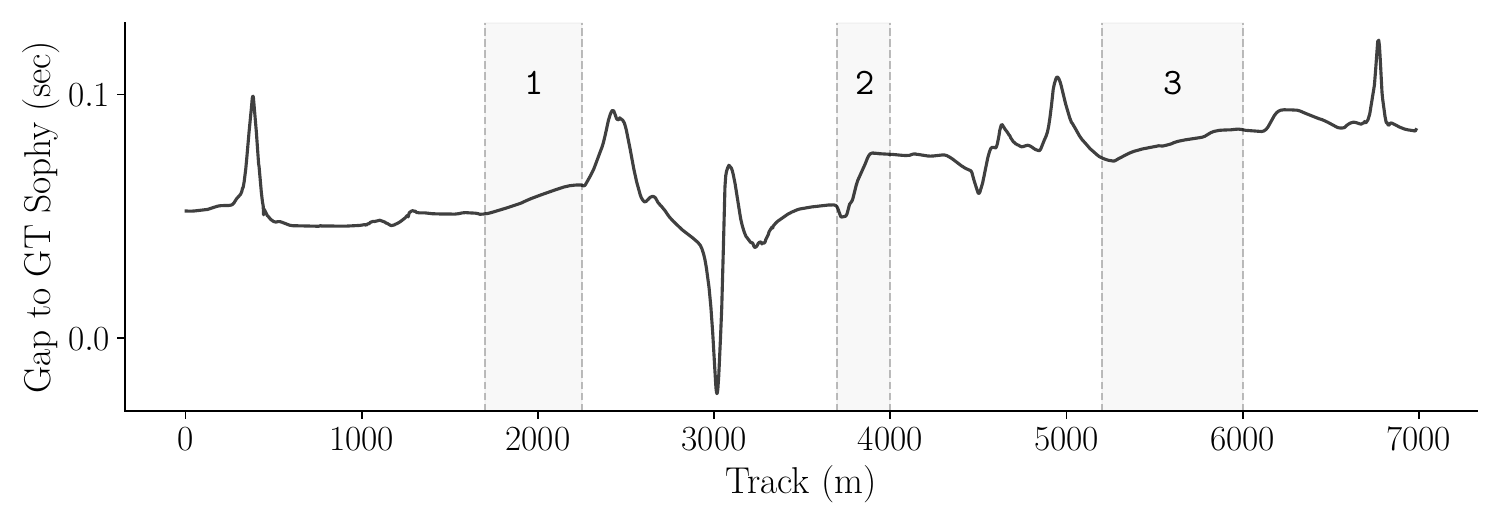}%
        \subcaption{\texttt{Spa}}\label{fig:spa_sophy_gap}%
    \end{minipage}%
\caption{Time difference between our agent and GT Sophy as a function of the progression in the track. We identify the trajectory sections highlighted in the \texttt{Monza} (Figure~\ref{fig:appendix_traj_sophy_monza}), \texttt{Tokyo} (Figure~\ref{fig:appendix_traj_sophy_tokyo}) and \texttt{Spa} (Figure~\ref{fig:appendix_traj_sophy_spa}) tracks. Lower is better.}
\label{fig:ablation_gap_sophy}
\end{figure*}

We qualitatively compare the trajectories of our agent and of the best human player across all scenarios: \texttt{Monza} in Figure~\ref{fig:appendix_traj_monza}, \texttt{Tokyo} in Figure~\ref{fig:appendix_traj_tokyo} and \texttt{Spa} in Figure~\ref{fig:appendix_traj_spa}.

The results show, across multiple sections of the tracks, that our agent does not simply follow the trajectory of the fastest human player but, in fact, exhibits novel racing behavior: in \texttt{Monza}, in straights our agent drives much closer to the curb, while the human player takes a more center line along the track, and in curves and chicanes it takes different driving lines; in \texttt{Tokyo} we once again see that our agent takes driving lines much closer to the curb than the fastest human; in \texttt{Spa} the results show that our agent takes different driving lines in both straights and curves. These significant differences in driving behavior raise the potential to use super-human racing agents as a training tool for human drivers, as previously identified in~\citep{gt_sophy}.

We also present the time difference between our agent and the fastest human reference driver along the progression of the track in Figure~\ref{fig:ablation_gap}. The results show the competitive nature and challenge of our task: our agent does not simply outperforms the human driver from the beginning of the lap, but gains and loses time against the human driver throughout the whole track. However, by the end of the track, our agent is able to outperform the fastest human driver.

\subsection{Comparison to GT Sophy}

We qualitatively compare the trajectories of our agent and of GT Sophy~\citep{gt_sophy} across all scenarios: \texttt{Monza} in Figure~\ref{fig:appendix_traj_sophy_monza}, \texttt{Tokyo} in Figure~\ref{fig:appendix_traj_sophy_tokyo} and \texttt{Spa} in Figure~\ref{fig:appendix_traj_sophy_spa}. The results show that overall the trajectory of our agent follows that of Sophy, despite our agent not having access to global features. However, the results show that often in long straights, for example Figure~\ref{fig:appendix_traj_sophy_monza} (1, 3), our agent takes a different racing line to Sophy, due to the absence of long-range information about the track in our agent's input.

Additionally, in Figure~\ref{fig:ablation_gap_sophy} we present the time difference between our agent and GT Sophy along the progression of the track, to understand where our agent actively loses time. The results show that,  despite the similarity of the trajectories in these sections, our agent mostly loses time in curve sections, for example Figure~\ref{fig:appendix_traj_sophy_spa} (2, 4). This result hints that GT Sophy, having access to long-range, precise information about the forward track limits, can approach the curve with a better velocity profile, leading to the sudden increase in the time difference between our agent and GT Sophy.

\section{Perturbation Study}
\label{sec:results:generalization}

\begin{figure}[p]
\centering
    \begin{minipage}[b]{0.55\linewidth}
        \centering
        \includegraphics[width=0.95\linewidth]{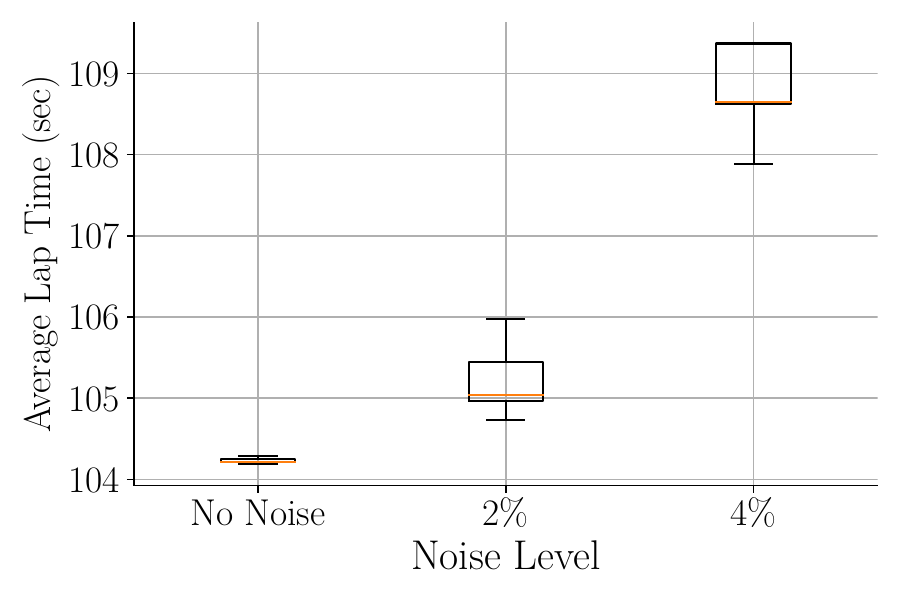}%
        \subcaption{Noise in propriocentric observations}
        \label{fig:sensor_noise}
    \end{minipage}%
    \hfil
    \begin{minipage}[b]{0.55\linewidth}
        \centering
        \includegraphics[width=0.95\linewidth]{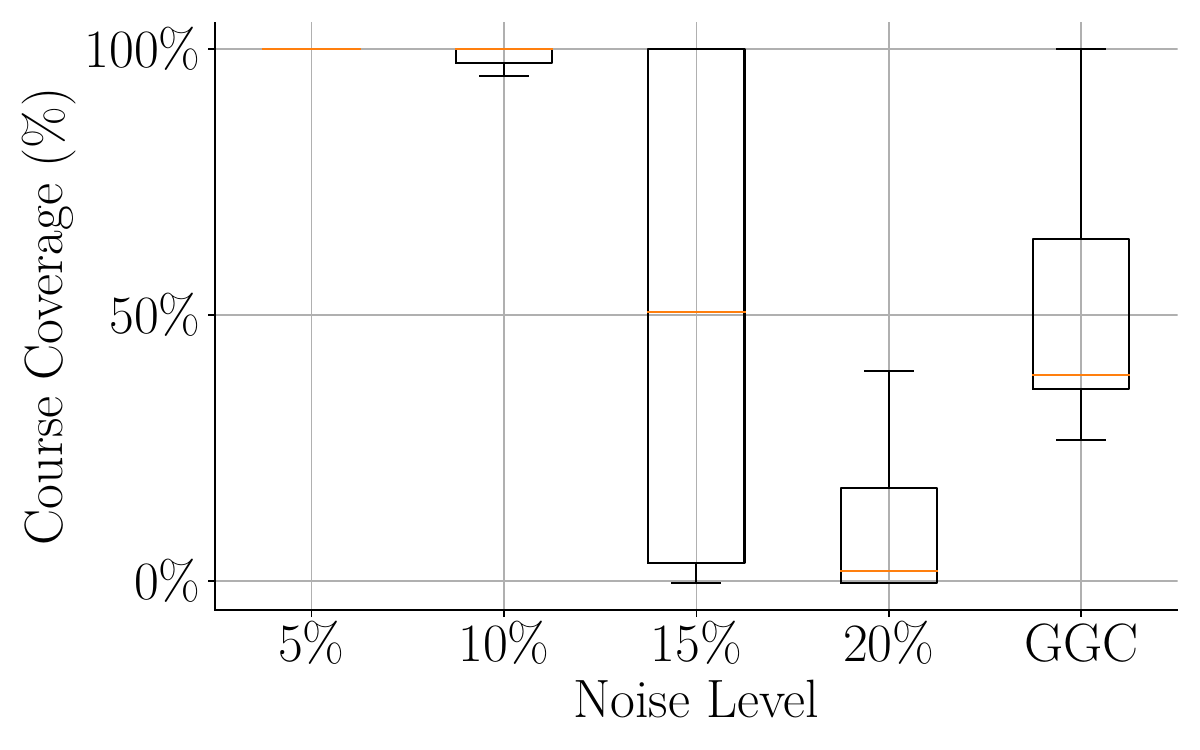}%
        \subcaption{Noise in image observations}
        \label{fig:sensor_image}
    \end{minipage}%
    \hfil
    \begin{minipage}[b]{0.55\linewidth}
        \centering
        \includegraphics[width=0.95\linewidth]{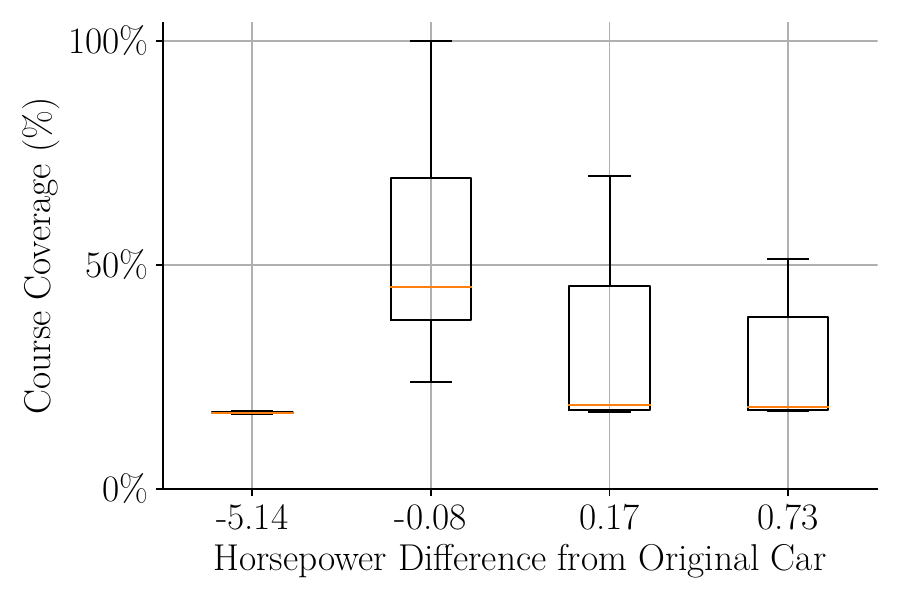}%
        \subcaption{Generalization to different cars}
        \label{fig:different_car}
    \end{minipage}%
\caption{Results of the perturbation study of our agent in the \texttt{Monza} scenario: (a) Completed lap times of agents with noise in the propriocentric features added during evaluation; (b) Course coverage of agents with noise in the image features added during evaluation; (c) Course coverage of agents with different car models during evaluation.}
\label{fig:robustness}
\end{figure}

To identify the limitations of our agent, we conducted an additional evaluation under various perturbation conditions. In this section, we select the top five agents from each \texttt{Monza} trial, for all evaluation. Note that training an agent able to generalize to unseen conditions during training is outside the scope of this work. For completeness, we present here a robustness evaluation and leave the further improvement of the generalization of our agent to future work.

\noindent \textbf{Noise in propriocentric observations:} We evaluated our agent under different levels of noise in the propriocentric observations by following the same procedure of \citet{gts}.
All agents completed laps with 2\% noise and 33\% of the agents completed laps with 4\% noise. However, the agents are no longer able to drive with more than 6\% noise.
Figure \ref{fig:sensor_noise} shows the average lap times of the completed laps, which suggests that the agent loses racing performance with increased noise levels in the propriocentric observation.

\noindent \textbf{Noise in image observations:} We evaluated our agent under different levels of noise in the image observations.
The noise level is defined as the percentage of randomly selected pixels from the complete image that are replaced by black pixels at each time step.
We additionally evaluated the same type of noise applied to the pixels highlighted by GGC analysis, as described in Section~\ref{sec:results:policy}, which we denote as GGC.
Note that GGC, on average, highlights up to 5\% of the total number of pixels of the image in our setting.
Figure \ref{fig:sensor_image} shows the course progression achieved by the agents with each noise level.
The agents consistently completed laps with up to 10\% noise and didn't complete any laps at 20\% noise.
Interestingly, the agents dropped performance significantly with GGC-based noise even though GGC only highlights 5\% of pixels at most.
This result further indicates that the pixels highlighted by GGC are vital for our agent to control the vehicle.

\noindent \textbf{Generalization to different time-of-the day conditions:} GT7 allows us to change the time of the race, resulting in a change in environmental elements, such as position and number of clouds, but also on the lighting conditions of the track. We then use the time of the race as a feature to evaluate the robustness of our agent against image perturbations. We observe that evaluating our agents at different times than the one of training results in their inability to complete laps around the track.

\noindent \textbf{Generalization to different cars:} To test the robustness of our agent to unseen dynamics, we evaluate it with different car models from the one used during training. We sorted all car models available in GT7 by horsepower and selected four \emph{neighbor} car models (two slower cars and two faster cars)\footnote{The selected car models are Porsche 959 '87, Peugeot 205 Turbo 16 Evolution 2 '86, Dodge Viper GTS '02, and Lamborghini Countach 25th Anniversary '88, in ascending order of horsepower.}.
Figure \ref{fig:different_car} shows the course progression achieved by each car model.
The result suggests that executing our policy in cars with increasingly higher horsepower leads to the worse generalization capability: due to their higher performance (e.g. in terms of velocity), the agent may experience dynamical states not experienced during training (e.g., high velocities in straight sections). On the other hand, the results show that our agent is comparably better in driving lower-performance cars.

\section{Guided Grad-CAM for Visual Analysis of Policies}
\label{section:gradcam_theory}

We adapt the basic Guided Gradient-weighted Class Activation Mapping (GGC)~\citep{selvaraju2017grad} algorithm to RL scenarios, following:

\begin{enumerate}
    \item During execution, we perform a forward pass given image and propriocentric observations $(\mathbf{o}^i, \mathbf{o}^p)$. The policy network outputs an action-specific average value $a$, from a truncated Gaussian distribution $\mathcal{N}(a, a_\sigma)$.
    
    \item We compute the gradient of the mean $a$ with respect to the feature maps $A^k$ in the last convolutional layer of the image encoder (Conv4 in Table~\ref{table:architecture}):
    \begin{equation}
    \frac{\partial a}{\partial A^k}.
    \end{equation}

    \item We apply a global average pooling to the Grad-CAM gradient to obtain neuron importance weights $w^k_{a}$ for each feature map:
    \begin{equation}
    w^k_{a} = \frac{1}{Z} \sum_{i} \sum_{j} \frac{\partial a}{\partial A^k_{ij}}, \quad 
    \end{equation}
    where $Z = W \times H$ is the normalization factor for the spatial dimensions of the feature map, $W$ is the weight of the feature map and $H$ is the height of the feature map.

    \item We obtain the Grad-CAM activation map, $L^{\text{Grad-CAM}}$ by performing a weighted combination of the feature maps followed by a ReLU function:
    \begin{equation}
    L^{\text{Grad-CAM}}_a = \text{ReLU}\left(\sum_k w^k_a A^k\right).
    \end{equation}
    
    \item We perform a separate backpropagation step to compute the gradient of the action value $a$ with respect to the input image observation $\mathbf{o}^i$:
    \begin{equation}
    \frac{\partial a}{\partial \mathbf{o}^i}.
    \end{equation}
    Furthermore, we filter out negative values from this gradient, in order to consider only features that have a positive influence on the action.
    
    \item We multiply the Grad-CAM activation map with the guided backpropagation result to obtain the Guided Grad-CAM visualization:
    \begin{equation}
    L^{\text{Guided Grad-CAM}}_{a} = L^{\text{Grad-CAM}}_{a} \times \frac{\partial a}{\partial \mathbf{o}^i}.
    \end{equation}
    \item Finally, we normalize the GGC visualization and, possibly, clip the result by a predefined noise threshold value.
\end{enumerate}

\section{Additional Guided Grad-CAM visualizations}
\label{sec:appendix_additional_ggc}

\begin{figure}[p]
    \centering
    \begin{subfigure}{1.\textwidth}
        \centering
        \includegraphics[width=\textwidth]{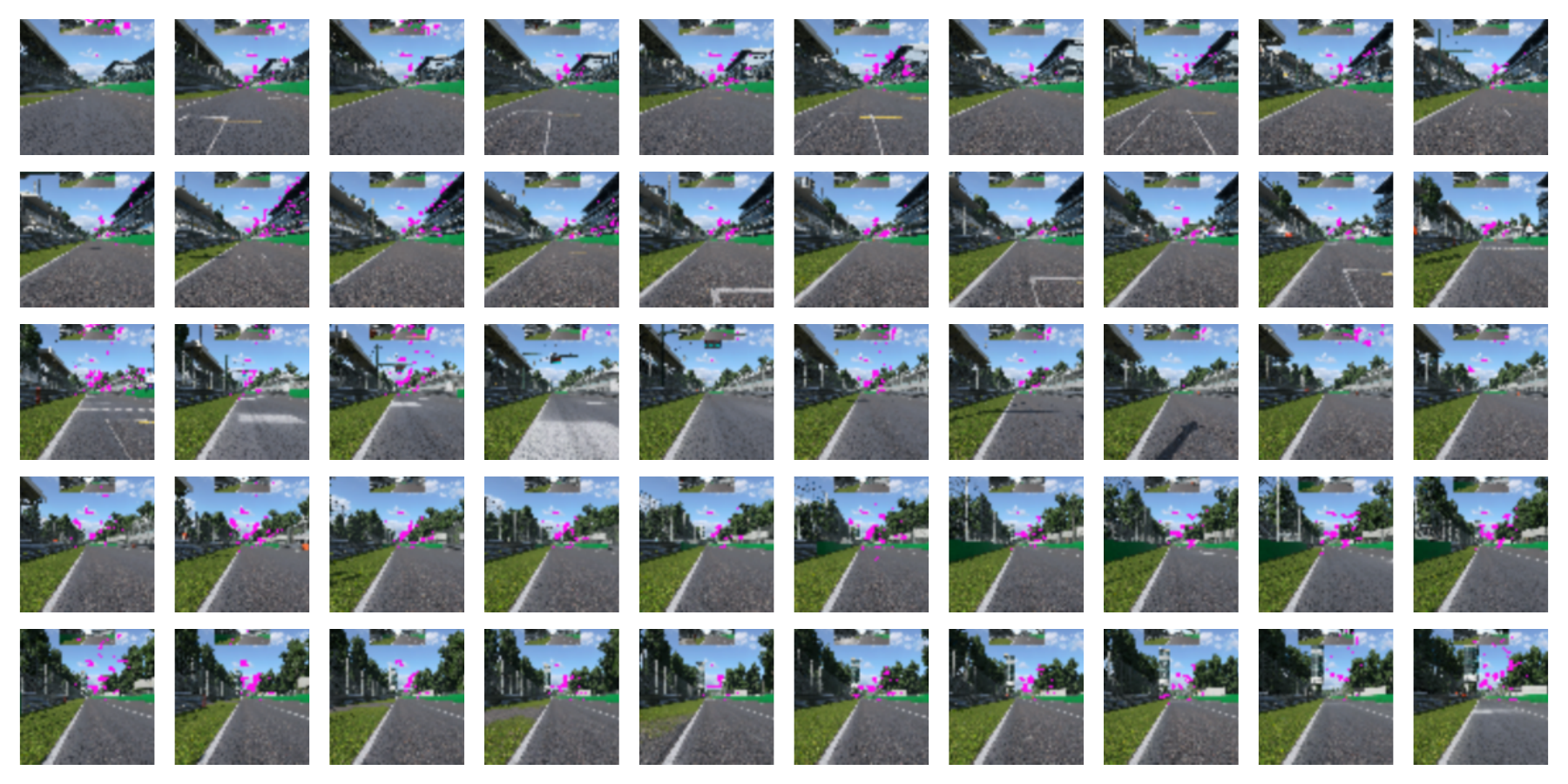}
        \caption{Straight section (150-539 m)}
    \end{subfigure}
    \hfill
    \begin{subfigure}{1.\textwidth}
        \centering
        \includegraphics[width=\textwidth]{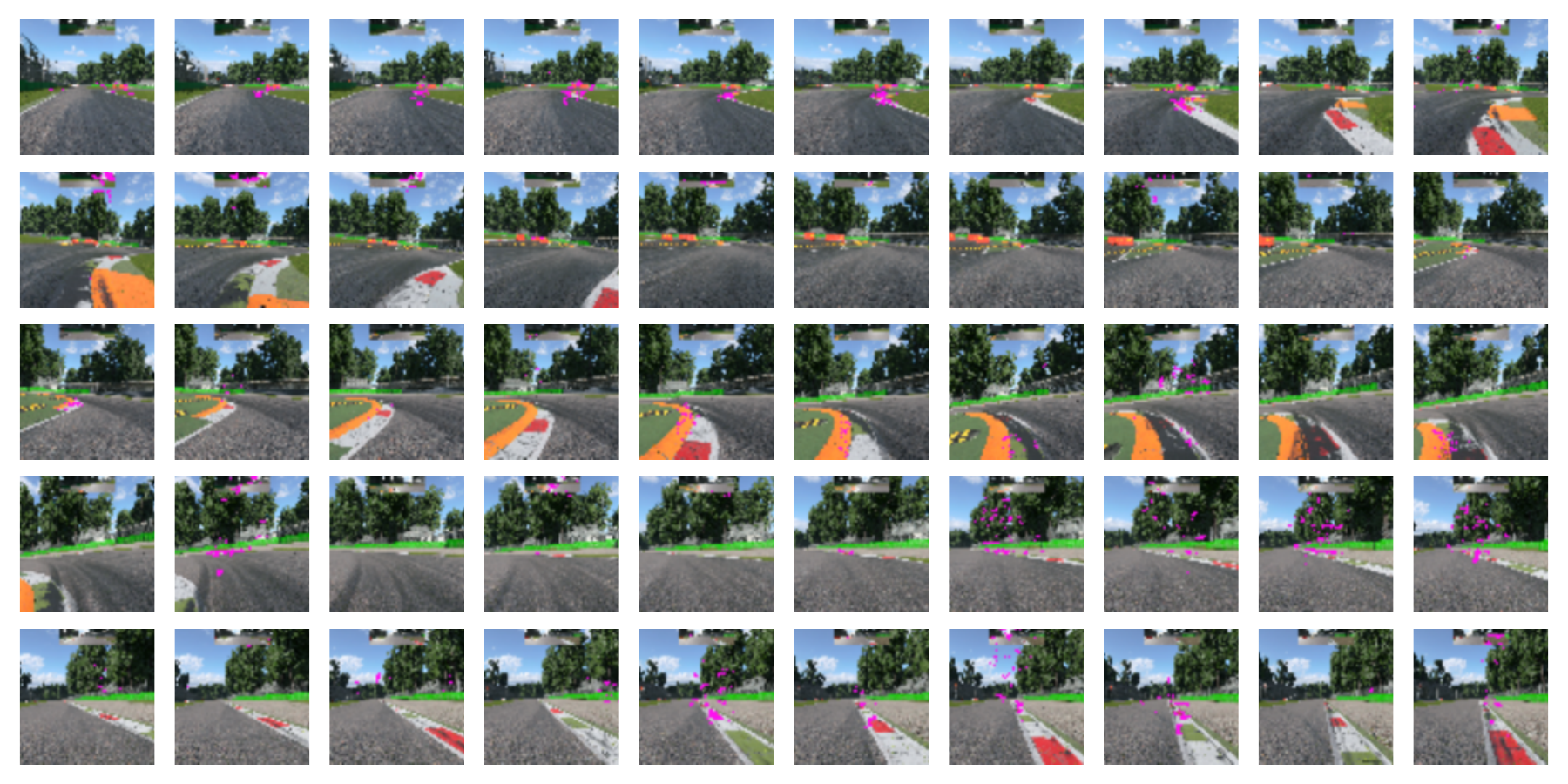}
        \caption{Chicane section (899-993 m)}
    \end{subfigure}
    \caption{Guided Grad-CAM (GGC) visualization of our racing agent for two sections of the \texttt{Monza} track: (Top) a straight section; (Bottom) a chicane section. We show in pink the positive gradients for the delta steering angle action computed using the policy of our agent. We show the top 80\% of the gradients in the visualization, to reduce noise. Best viewed with color and zoomed in.}
    \label{fig:appendix_gradcam_monza}
\end{figure}

\begin{figure}[p]
    \centering
    \begin{subfigure}{1.\textwidth}
        \centering
        \includegraphics[width=\textwidth]{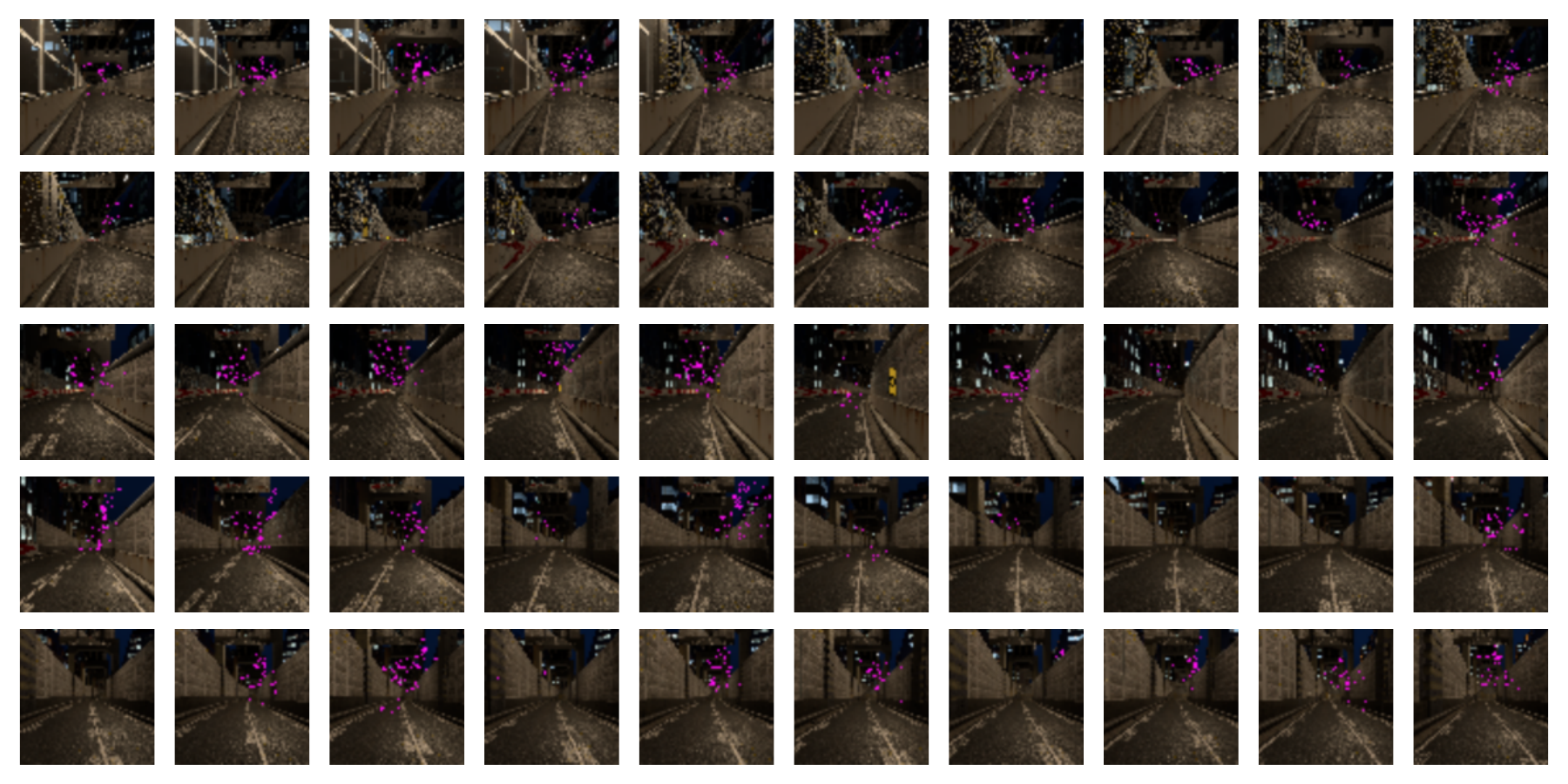}
        \caption{Straight outdoor section (625-898 m)}
    \end{subfigure}
    \hfill
    \begin{subfigure}{1.\textwidth}
        \centering
        \includegraphics[width=\textwidth]{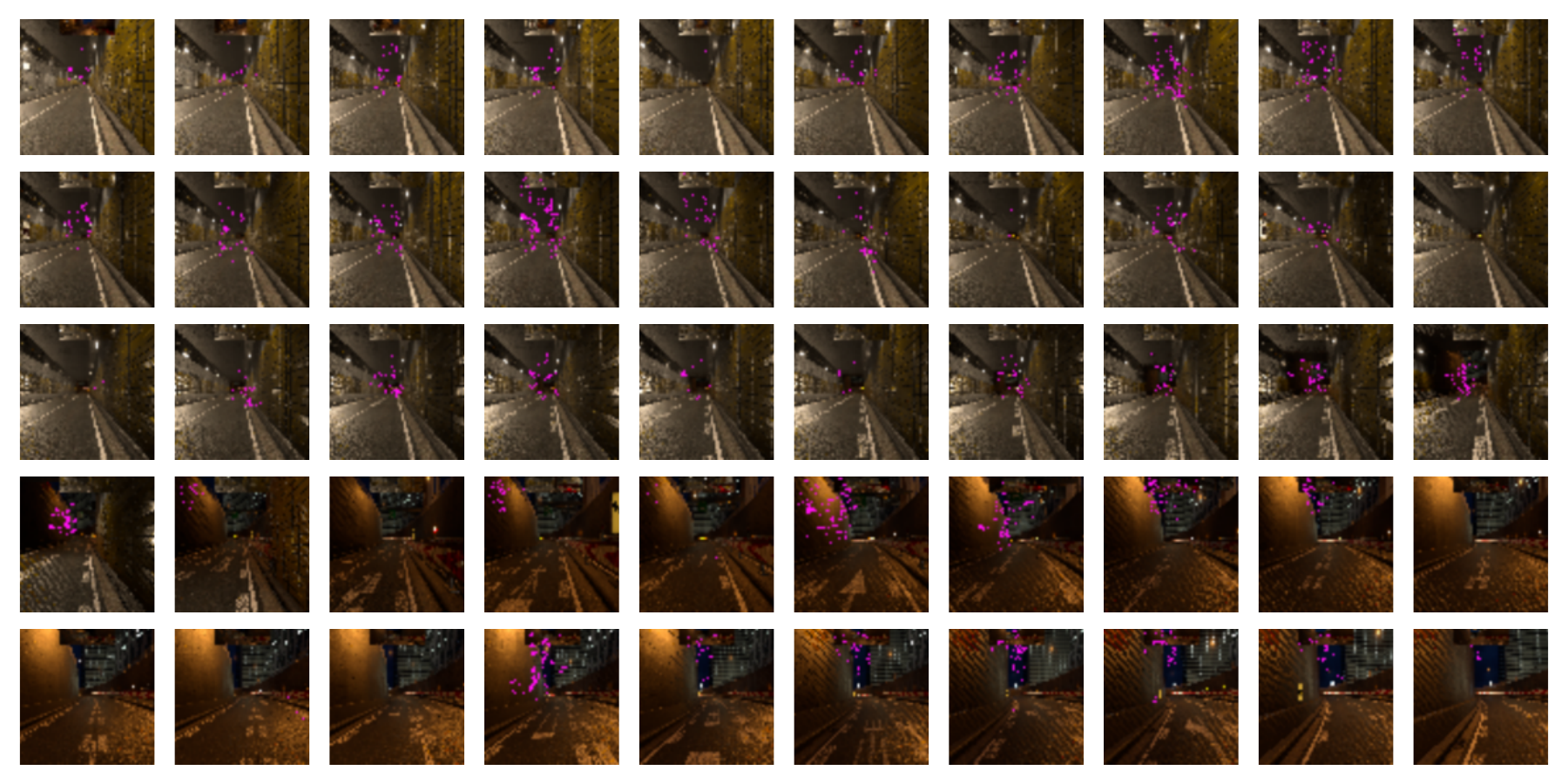}
        \caption{Straight tunnel section (3499-3787 m)}
    \end{subfigure}
    \caption{Guided Grad-CAM (GGC) visualization of our racing agent for two sections of the \texttt{Tokyo} track: (Top) a straight outdoor section; (Bottom) a straight indoor section. We show in pink the positive gradients for the delta steering angle action computed using the policy of our agent. We show the top 80\% of the gradients in the visualization, to reduce noise. Best viewed with color and zoomed in.}
    \label{fig:appendix_gradcam_tokyo}
\end{figure}

\begin{figure}[p]
    \centering
    \begin{subfigure}{1.\textwidth}
        \centering
        \includegraphics[width=\textwidth]{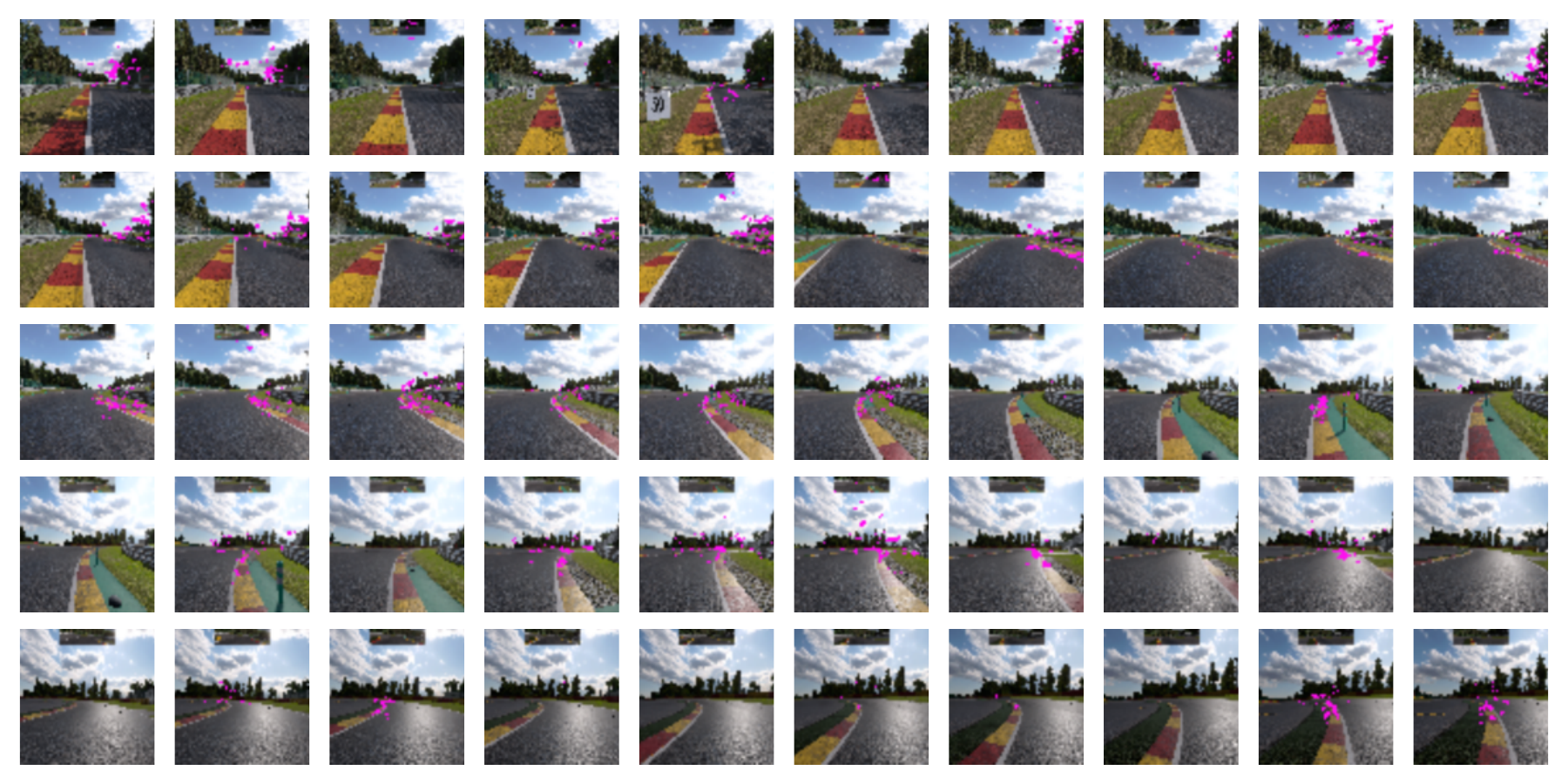}
        \caption{Chicane section (2296-2484 m)}
    \end{subfigure}
    \hfill
    \begin{subfigure}{1.\textwidth}
        \centering
        \includegraphics[width=\textwidth]{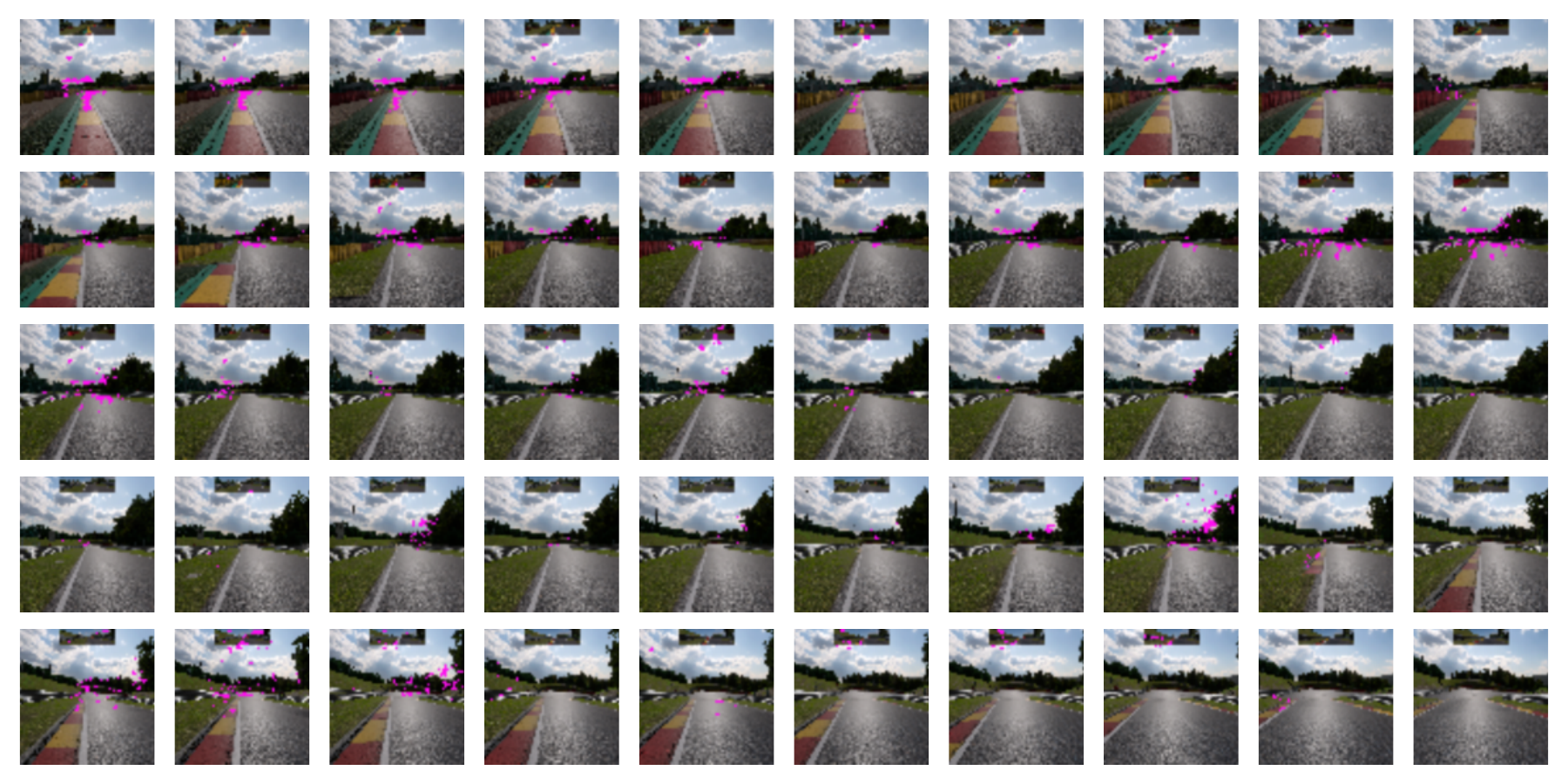}
        \caption{Straight section (2741-2963 m)}
    \end{subfigure}
    \caption{Guided Grad-CAM (GGC) visualization of our racing agent for two sections of the \texttt{Spa} track: (Top) a chicane section; (Bottom) a straight section. We show in pink the positive gradients for the delta steering angle action computed using the policy of our agent. We show the top 80\% of the gradients in the visualization, to reduce noise. Best viewed with color and zoomed in.}
    \label{fig:appendix_gradcam_spa}
\end{figure}

We present additional GGC visualizations for the action \emph{delta steering angle} for our agent across all tracks: \texttt{Monza} (Figure~\ref{fig:appendix_gradcam_monza}), \texttt{Tokyo} (Figure~\ref{fig:appendix_gradcam_tokyo}) and \texttt{Spa} (Figure~\ref{fig:appendix_gradcam_spa}). Once again, the results shows that our agent focuses on different features depending on the track section: in long straights, our agent focuses on far-away visual features, such as horizon of the track or tree lines in the distance, while in chicanes and tight curves, our agent focuses on the curbs of the track, which are fundamental to successfully change its direction without going off-track. 

\section{Implementation Details}
\label{sec:appendix_model}

\noindent\textbf{Model Architecture:} We provide the detailed architecture of our agent in Table~\ref{table:architecture}.

\noindent\textbf{Data collection:} At the beginning of every episode, the initial position of the agent is uniformly sampled from in-course areas as well as left and right off-course areas within 5\% of track width. The agent faces towards a center line 30m ahead and the launch speed is uniformly sampled from 0 to 104.607km/h. We reset the episode every 150 seconds.

\noindent\textbf{Distributed training:} Unlike widely used RL simulators that can execute faster with powerful computing resources (e.g., MuJoCo~\citep{mujoco}), GT7 executes its simulation in real time. To compensate for the simulation speed, we use an asynchronous distributed training scheme by following \citet{gt_sophy}. In this work, we consider 20 rollout workers for data collection and policy evaluation, each assigned to a different PlayStation$^\text{\textregistered}$ 4 system connecting to rollout workers via ethernet.
However, the game screen retrieval via ethernet induces additional latency, which makes it difficult to train agents in real time. Therefore, we configured the simulator to block simulation steps until the simulator receives the next action commands sent by the rollout worker. We applied this setting to all experiments including the baseline GT Sophy. We evaluate the impact of the synchronous communication scheme on the performance of our agent in Appendix~\ref{appendix_additional_ablation_study_sync}.

\section{Reward and Training Hyperparameters}
\label{sec:appendix_training}

We present in Table~\ref{table:reward_parameters} the reward function parameters used in our work, selected empirically. Moreover, we present in Table~\ref{table:hyperparameters} the list of training hyperparameters used by our approach. We keep the same hyperparameters across all scenarios, except for the number of training epochs: in \texttt{Monza} we use 4000 training epochs and in both \texttt{Tokyo} and \texttt{Spa} we use 2000 epochs, where an epoch consists of 6000 gradient steps. We use a higher number of training epochs in \texttt{Monza} because in this scenario we employ a faster racing car, which requires a more precise maneuver to achieve a higher level of performance.

\begin{table}[t]
  \centering
  \renewcommand{\arraystretch}{1.2} 
  \setlength\arrayrulewidth{0.1pt} 
  \caption{Model architecture of our racing agent.}
  \label{table:architecture}
  \begin{tabular}{@{}lcc@{}}
    \toprule
    Layer Description & Input Dimensions & Output Dimensions \\ \midrule
    \rowcolor{lightblue} \multicolumn{3}{l}{Actor Network} \\ 
    Conv1: 64 filters, 4x4, stride 2, ReLU & 64x64x3 & 32x32x64 \\
    \arrayrulecolor{gray!20}\hdashline\arrayrulecolor{black}
    Conv2: 128 filters, 4x4, stride 2, ReLU & 32x32x64 & 16x16x128 \\
    \arrayrulecolor{gray!20}\hdashline\arrayrulecolor{black}
    Conv3: 256 filters, 4x4, stride 2, ReLU & 16x16x128 & 8x8x256 \\
    \arrayrulecolor{gray!20}\hdashline\arrayrulecolor{black}
    Conv4: 512 filters, 4x4, stride 2, ReLU & 8x8x256 & 4x4x512 \\
    \arrayrulecolor{gray!20}\hdashline\arrayrulecolor{black}
    FC, 128 units, ReLU & 4x4x512 & 128 \\ \midrule
    MLP FC1: 2048 units, ReLU & 145 (128 + 17) & 2048 \\
    \arrayrulecolor{gray!20}\hdashline\arrayrulecolor{black}
    MLP FC2: 2048 units, ReLU & 2048 & 2048 \\
    \arrayrulecolor{gray!20}\hdashline\arrayrulecolor{black}
    MLP FC3: 2048 units, ReLU & 2048 & 2048 \\
    \arrayrulecolor{gray!20}\hdashline\arrayrulecolor{black}
    MLP FC4: 2048 units, ReLU & 2048 & 2048 \\
    \arrayrulecolor{gray!20}\hdashline\arrayrulecolor{black}
    MLP FC Output: 4 units, \texttt{Tanh} & 2048 & 4 \\
    \midrule
    \rowcolor{lightblue} \multicolumn{3}{l}{Critic Network} \\ 
    MLP FC1: 2048 units, ReLU & 531 & 2048 \\
    \arrayrulecolor{gray!20}\hdashline\arrayrulecolor{black}
    MLP FC2: 2048 units, ReLU & 2048 & 2048 \\
    \arrayrulecolor{gray!20}\hdashline\arrayrulecolor{black}
    MLP FC3: 2048 units, ReLU & 2048 & 2048 \\
    \arrayrulecolor{gray!20}\hdashline\arrayrulecolor{black}
    MLP FC4: 2048 units, ReLU & 2048 & 2048 \\
    \arrayrulecolor{gray!20}\hdashline\arrayrulecolor{black}
    MLP FC Output: 32 units, linear & 2048 & 32 \\
  \end{tabular}
\end{table}

\begin{table}[t]
\centering
\renewcommand{\arraystretch}{1.2} 
\setlength\arrayrulewidth{0.1pt} 
\caption{Reward parameters of our racing agent.}
\label{table:reward_parameters}
\begin{tabular}{@{}lc@{}}
\toprule
Parameter & Value \\
\midrule
$\lambda^o$ & 10 \\
\arrayrulecolor{gray!20}\hdashline\arrayrulecolor{black}
$\lambda^w$ & 10 \\
\arrayrulecolor{gray!20}\hdashline\arrayrulecolor{black}
$\lambda^s$ & 3 \\
\arrayrulecolor{gray!20}\hdashline\arrayrulecolor{black}
$\lambda^h$ & $5$ \\
\arrayrulecolor{gray!20}\hdashline\arrayrulecolor{black}
$c^d$ & 0.014 \\ 
\arrayrulecolor{gray!20}\hdashline\arrayrulecolor{black}
$c^s$ & 182.883569 \\
\arrayrulecolor{gray!20}\hdashline\arrayrulecolor{black}
$c^o$ & 0.034 \\
\end{tabular}
\end{table}

\begin{table}[t]
\centering
\renewcommand{\arraystretch}{1.2} 
\setlength\arrayrulewidth{0.1pt} 
\caption{Training hyperparameters of our racing agent.}
\label{table:hyperparameters}
\begin{tabular}{@{}lc@{}}
\toprule
Hyperparameter & Value \\
\midrule
Activation function & ReLU \\
\arrayrulecolor{gray!20}\hdashline\arrayrulecolor{black}
Optimizer & Adam~\citep{kingma2014adam} \\
\arrayrulecolor{gray!20}\hdashline\arrayrulecolor{black}
Batch size & 512 \\
\arrayrulecolor{gray!20}\hdashline\arrayrulecolor{black}
Policy learning rate & $2.5 \times 10^{-5}$ \\
\arrayrulecolor{gray!20}\hdashline\arrayrulecolor{black}
Critic learning rate & $2.5 \times 10^{-5}$ \\ 
\arrayrulecolor{gray!20}\hdashline\arrayrulecolor{black}
Global norm of critic gradient clipping & 10 \\
\arrayrulecolor{gray!20}\hdashline\arrayrulecolor{black}
Discount factor & 0.9896 \\
\arrayrulecolor{gray!20}\hdashline\arrayrulecolor{black}
SAC entropy temperature~\citep{sac} & 0.01 \\
\arrayrulecolor{gray!20}\hdashline\arrayrulecolor{black}
Number of quantiles & 32 \\
\arrayrulecolor{gray!20}\hdashline\arrayrulecolor{black}
Multi-step & 7 \\
\arrayrulecolor{gray!20}\hdashline\arrayrulecolor{black}
Replay buffer size & 2.5M \\
\end{tabular}
\end{table}

\begin{figure}[]
    \centering
    \begin{subfigure}{0.46\textwidth}
        \centering
        \begin{subfigure}{0.48\textwidth}\includegraphics[width=\textwidth]{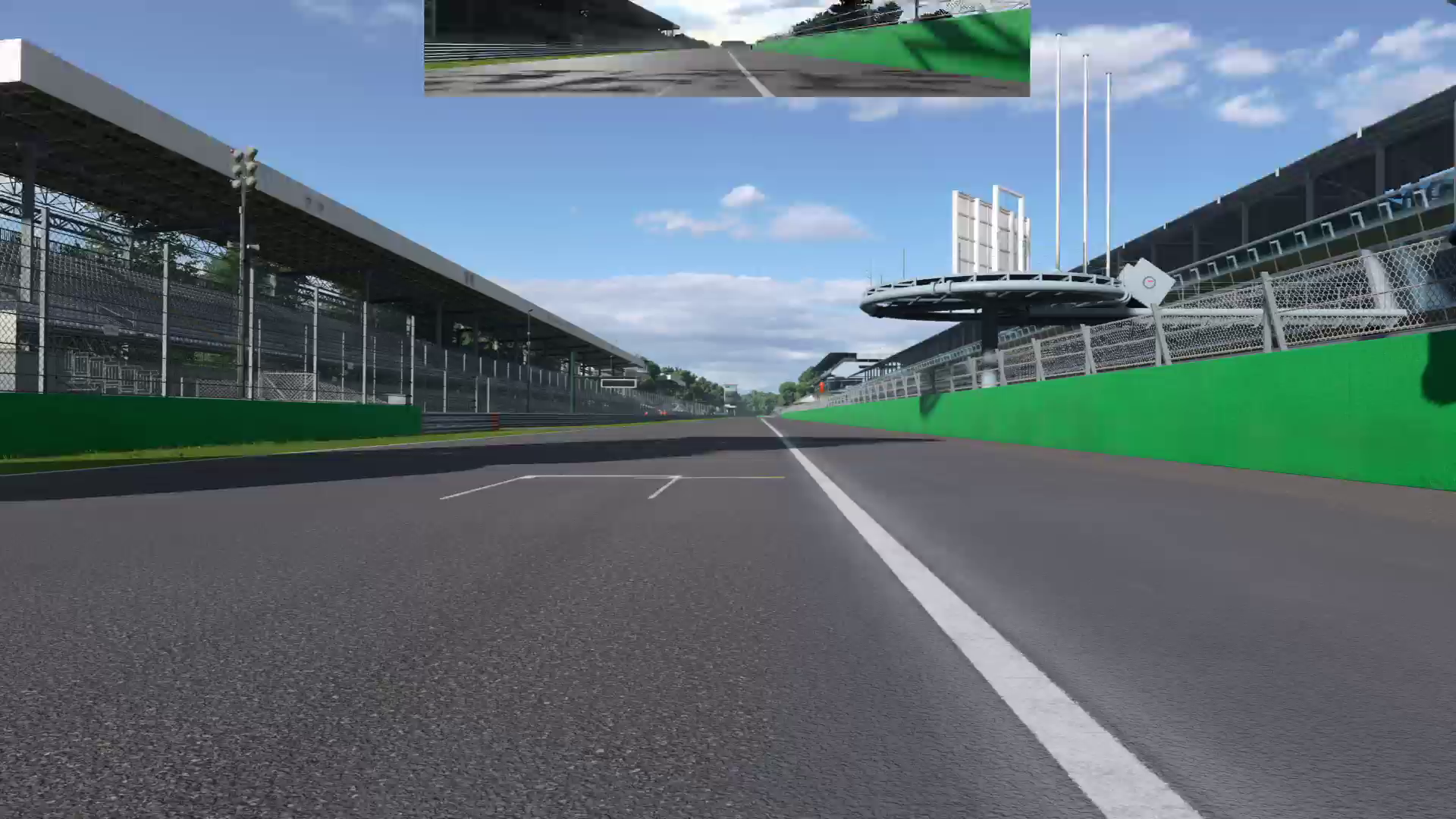}\end{subfigure}
        \begin{subfigure}{0.48\textwidth}\includegraphics[width=\textwidth]{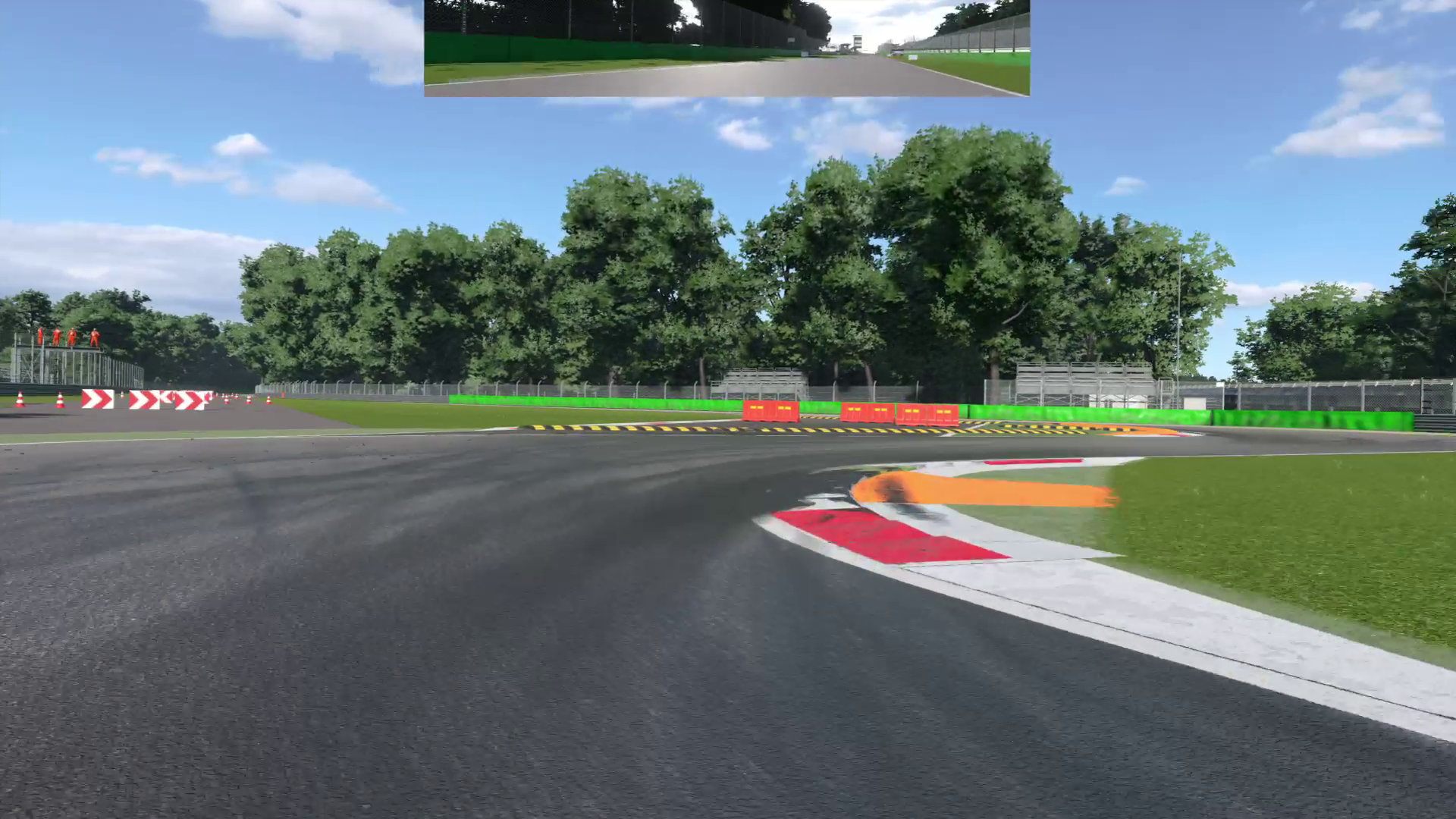}\end{subfigure}
        \caption{$1920\times1080$ (Native screen resolution)}
    \end{subfigure}
    \begin{subfigure}{0.26\textwidth}
        \centering
        \begin{subfigure}{0.48\textwidth}\includegraphics[width=\textwidth]{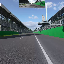}\end{subfigure}
        \begin{subfigure}{0.48\textwidth}\includegraphics[width=\textwidth]{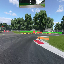}\end{subfigure}
        \caption{$64\times 64$ (Our Agent)}
    \end{subfigure}
    \begin{subfigure}{0.26\textwidth}
        \centering
        \begin{subfigure}{0.48\textwidth}\includegraphics[width=\textwidth]{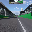}\end{subfigure}
        \begin{subfigure}{0.48\textwidth}\includegraphics[width=\textwidth]{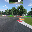}\end{subfigure}
        \caption{$32\times32$ (\emph{Small Image})}
    \end{subfigure}
    \caption{Example observations with three different resolutions: in images with $64\times 64$ resolution we can still observe some level of detail in most objects; However, in images with $32\times 32$ we are able to recognize only basic elements of the environment (e.g., track limits, sky, background).}
    \label{fig:appendix_small_image}
\end{figure}

\section{Additional Details on Image Resolution}
\label{sec:appendix_small_image}
To highlight how image compression affects the elements in the observation of our agent, we provide examples of game images with different resolutions in Figure~\ref{fig:appendix_small_image}: the original $1920\times1080$ image, a $64\times64$ image used by our agent and a reduced $32\times32$ image used in our ablation study in Section~\ref{sec:results:ablation}. The figure shows that across all resolutions we can identify critical elements of the environment to perform the task (such as the track limits), yet with decreasing level of precision.

The choice of resolution also affects the total parameter count of our model: since we keep the same architecture for both our agent and the \emph{small image} ablated version, the size of feature maps for the latter version in each convolutional layers is half the size of values of the former, described in Table~\ref{table:architecture}. It may be possible to improve the performance of the \emph{small image} ablation by optimizing the convolutional encoder architecture for input images of $32\times32$, but we leave this exploration to future work.

\end{document}